# Can Turing machine be curious about its Turing test results?

*Three informal lectures on physics of intelligence*

Alex Ushveridze [1]
Capella University, Minneapolis, USA

**_Abstract_**

*What is the nature of curiosity? Is there any scientific way to understand the origin of this mysterious force that drives the behavior of even the "stupidest" naturally intelligent systems and is completely absent in their "smartest" artificial analogs? Can we build AI systems that could be curious about something, systems that would have an intrinsic motivation to learn? Is such a motivation quantifiable? Is it implementable? Will we ever see artificially built systems having their views, values or goals? Or maybe the only mission of AI is to imitate intelligence, fool Turing test judges and build next-generation gadgets? These are the main questions I will try to address here, in these lectures.*

*I will discuss this problem from the standpoint of physics. Treating intelligence as a physical phenomenon will not only allow us to understand what its driving force is, but will also give us a powerful formalism capable of studying it mathematically in a systematic and unified way. The relationship between physics and intelligence is a consequence of the fact that "correctly predicted information" is nothing but an energy resource, and the process of thinking can be viewed as a process of accumulating and spending this resource through the acts of perception and, respectively, decision making. The natural motivation of any autonomous system to keep this accumulation/spending balance as high as possible allows one to treat the problem of describing the dynamics of thinking processes as a resource optimization problem.*

*Here I will propose and discuss a simple theoretical model of such an autonomous system which I call the Autonomous Turing Machine (ATM). The potential attractiveness of ATM lies in the fact that it is the model of a self-propelled AI for which the only available energy resource is the information itself. For ATM, the problem of optimal thinking, learning, and decision-making becomes conceptually simple and mathematically well tractable. This circumstance makes the ATM an ideal playground for studying the dynamics of intelligent behavior and allows one to quantify many seemingly unquantifiable features of genuine intelligence.*

*A closer look at this subject reveals its cross-disciplinary nature: it turns out that there are many striking parallels between diverse branches of artificial intelligence on the one hand and theoretical physics and business economics on the other hand. For this reason, I wanted to target this text to a maximally broad audience, including physicists, computer scientists, business analysts and philosophers of science. I hope that this explains its somewhat informal and not quite academic style and also a relatively high-level character of references.*

[1] Email: alex.ushveridze@capella.edu

# Contents

# Foreword: The physics of intelligence

## Is motivation quantifiable?

Many of today's AI systems[2] look astonishingly smart. They can perform complex tasks, learn fast, outperform humans in many areas and even pass Turing test[3] . Tomorrow's AI systems will probably look smarter[4]. Huge progress in this direction is stimulated by impressive advances in the area of "deep learning" with all its sub-branches such as sparse auto-encoders, restricted Boltzmann machines and others[5].

The concept of an auto-encoder – a learning machine which first encodes (compresses) external information and then decodes (decompresses) it trying to reproduce its original version – turned out to be extremely fruitful. There are at least three reasons for that. First of all, from the architectural standpoint, the auto-encoder is universal and can be applied to any data. Secondly, it does need a supervisor and thus may lie in the background of truly autonomous AI systems. Finally, it makes the learning process conceptually simple. The last circumstance is an indirect but strong indicator of the fundamental nature of the idea of an auto-encoder (because fundamental principles, by definition, cannot be complex). Another very fruitful idea was the concept of restricted Boltzmann machines – in some sense -- statistical versions of auto-encoders based on the use of probability distributions typical for complex physical systems in thermodynamic equilibrium.

So we can clearly see at least two trends here: one – towards the simplification of mathematical structures used for formalizing the concept of intelligent behavior and the other – towards incorporating physical theories in attempts to explain it. We are not yet in the position to present a universal algorithm capable of learning in any situation and from any data, but there is no doubt that sooner or later such an algorithm will be created (Domingos, 2015). However, when created, will it represent any truly intelligent system?

Since there is no accepted definition of "intelligence" (Hawkins & Blakeslee, 2005), (Legg & Hutter, 2007), the audience may split in answering this question. Probably the overwhelming majority will say yes. I do not belong to this group because I do not believe that intelligence can be reduced to the simple *ability* of a system to learn. If we define intelligence this way, we will miss the main point – the motivation to evolve towards intelligence. What should motivate, for example, the auto-encoders or restricted Boltzmann machines to have the learning-supporting architectures they have? After all, what motivates them to learn? These are not naïve questions, not at all. If the *motivation* to learn is absent – the system hardly can be qualified as intelligent and autonomous. Indeed, to be intelligent, it is not sufficient to be *able* to calculate, be *able* to solve problems and be *able* to answer questions – one should *need* to calculate, *need* to solve problems and *need* to answer questions. Even more: one should need to *ask* questions (Zoldi, 2015), (Lipson, 2007). All these particular needs should originate from an intrinsic need of a system to do something useful not only for us, its creators (and this is what the existing AI systems already do) but for *themselves*. This "artificial egocentrism" or "machine curiosity" – whatever we call it – is something we need to take very seriously because otherwise we will be forced to always deal with "universal answering machines". Those could be very helpful, no doubt, but not "naturally" intelligent. They never will have an internal drive to increase the level of their intelligence and will never be independent on us, humans. What I would like to have instead is, metaphorically said, a certain "universal asking machine" which, as I sincerely hope, will be kind enough and not too busy with its problems to be willing to answer some of my questions too.

[2] There is no strong agreement on what the term "Artificial Intelligence" (AI) -- should actually mean. According to the standard text-book classification, like for example given in (Russell, & Norvig, 2003), there are four candidates: thinking humanly, acting humanly, thinking rationally and acting rationally. We hope to demonstrate in this article that, at the fundamental level there is actually no distinction between these four.

[3] The statement that some existing AI systems are already able "to pass Turing test" may seem highly controversial. Probably, today it would be safer to say "to win Loebner Prize" instead. See for example (Oppy & Dowe, 2016). For the philosophical aspects of Turing Test see also (Saygin, et al., 2000).

[4] A good analysis of expert surveys on the future of AI is given in (Müller & Bostrom, 2014).

[5] For the detailed exposition of deep learning concepts and methods see for example (Bengio, 2009), (Deng & Yu, 2014), (Ng, 2016a). For technical tutorials with working demos and well documented code I would recommend (Ng, 2016b) and (Masters, 2016).

But do we need such machines? Are we ready for this switch from "machine learning" to "machine asking"? Is it not too dangerous for the humankind? Is "AI slavery", so to speak, a safer solution for us than "AI partnership"? These are all rhetorical and futurological questions. The problem I want to discuss here is much more practical from today's standpoint: is it possible, at least in principle, to build such a machine?

I think the global answer to this question should be yes, which follows from the very fact of our existence: nature has already demonstrated the feasibility of this program. I do not think it is necessary to mimic how nature has done it – the only thing we need is to understand the basic idea and once it is understood start looking for the shortcuts – like we have already done many times in the past. And this is where physics can help us.

## The magic formula

The secret word linking intelligence to physics is "energy". Each time the physicists succeed in relating energy to something else results in huge breakthroughs in both science and industry. This is quite understandable because energy is the main *resource* supporting our life and any clue shedding light on the ways of obtaining and controlling it is of primary importance for us.

One of the simplest and best known examples of such relationships is the famous Einstein's formula $E = mc^2$ which establishes the equivalence between mass and energy. Another example is Planck's formula $E = \hbar\nu$ establishing the relation between the frequency of a light wave and the energy of its quant (the photon). The role played by these two formulas on our lives is hard to overestimate. It is huge: one can safely say that most of our today's technologies are directly or indirectly based on them.

Here I am going to discuss another formula for energy which is not as widely known as the previous two but whose impact on our life may be even higher. This formula was derived by Ralph Landauer in 1961 (Landauer, 1961) and has a very simple-looking form:

$$E = kTI\ln 2.$$

Here, $k$ is the so-called Boltzmann's constant, $T$ is temperature, and $I$ is information. Why is this formula so important? Because it establishes the equivalence between information and energy, which, in turn, creates the link between intelligence and physics. This formula is just a quantitative manifestation of the amazing fact that information is a resource, exactly in the same sense as gas is a resource for our cars and food is a resource for all living beings. It is simply a certain low entropy stuff that can be consumed and then converted into useful work as any other fuel can. Consumption of information is what we call perception and the term "useful work" may stand for any of the useful actions the intelligent system can perform.

Of course, the energy scales characterizing these consumption-action processes are negligibly small in comparison with the scales we deal on an everyday base and at which our today's computers operate. Indeed, Landauer formula shows that 1 bit of information at room temperature can be converted into about $3 \times 10^{-21}$Joules of energy. Even in terms of the maximal capacities of personal computers it looks a rather small amount. For example, 1 TB of information may give us the energy equal to only 24$\times$ $10^{-9}$ Joules, which, roughly speaking, is the energy of a grain of rice moving with the velocity of 1 cm/sec. However, this tiny amount could be quite noticeable at the microscopic scales (or more precisely, at the molecular levels) at which our future computers may work. And this is the key point because it allows us to say that future intelligent systems working on molecular scales may have the *motivation* of behaving in such a way that maximizes the balance between their perceptions and actions!

## Thinking as a resource optimization process

So we see that one of the benefits of treating information as a resource is that the notion of motivation arises in this case in a very natural way. And this also leads us to the conclusion that internally motivated intelligent behavior could be mathematically describable as the process of optimizing the usage of a resource. Since the latter is a physical quantity, this opens the possibility of studying the behavior of intelligent systems by using a purely physical language. And this is an intriguing possibility.

Technically every resource optimization problem is reduced to the two closely related questions: how to maximize consumption and how to minimize spending.

To answer the first question, we can imagine a hypothetical and somewhat idealized situation where the system we are interested in is microscopic and the energy scales $kT$ are quite comparable with its energy needs. In this case the system would be interested in extracting the energy directly from the information that surrounds it. There is, however, a little problem with such an extraction because not any information can be used as fuel. To be a fuel (or, in other words, to be able to perform some useful work), the information must be *a priori* known. To be more precise, the energy value contained in one bit of information depends on the extent to which we know it in advance. In other words, the only way for a system to increase the energy value of the information it consumes is to predict it better. This fact creates a strong internal motivation for a system to learn – i.e. to discover the relationships between different spatial and temporal parts of the external world.

What about the second question? At first glance the answer to it seems quite straightforward: the system should maximally reduce the number of its actions because each action leads to the loss of energy. However, this simple recipe hides a rather serious problem. The point is that the system cannot simply skip all of the actions it performs regardless of their kind. There are some of them that should be performed, no matter what, again and again. The most typical examples of such non-skippable actions are those that are crucial for the system's survival, as, for example, the search for new possible locations of fuel. There are some other non-skippable actions like learning new patterns, recovering from errors, etc. This situation creates another strong motivation for a system to be maximally disciplined in spending its resource and carefully decide which actions to perform and which not.

Combining these two answers together we arrive at the picture in which the thinking process appears as a chain of observations and decisions or, in the resource-based language, as a chain of consumptions and spendings. The maximization of the consumption/spending balance along this chain is a very non-trivial optimization problem. The better the system can solve it, the smaller the amount of external energy supply needed for normal functioning will be.

It is very tempting to stop here and jump into a discussion of this energy saving problem because of its huge theoretical importance and many practical applications. However, we will do that later. Now we want to go a little bit further and ask the following question: If the process of thinking is the result of optimizing the cumulative consumption and spending balance, is it possible to optimize it in such a way that it would remain positive all the time? Or, in other words, can we close the perception-decision chain and convert it into a perception-decision loop? Answering "yes" to this question would mean that the intelligence could be self-propelled – i.e., not requiring any extra sources of the traditional fuel at all. Anything that such a system would need for normal functioning it could find in the information that surrounds it.

The last statement probably needs some clarification, simply because of the equivalence between information and fuel we just stated. Indeed, how can an AI system distinguish between what to look for, for fuel or for information, if they are the same? Of course, it cannot, but the main idea behind this approach is just to avoid treating information and fuel as two different types of resources. We want to use instead only one type because this is methodologically much easier. It is convenient to choose for this role the information – and the formula $E = kTI\ln 2$ allows to do so.

The goal of this paper is just to explore the very possibility of a bit-level self-propulsion – which is the road to autonomous AI systems. True motivation and thus true intelligence is not achievable without true autonomy. It is the real driver of evolution because the motivation for increasing the efficiency and thus complexity comes from the need to survive in the situations when the external supply of resources is not guaranteed.

## The autonomous Turing machines

To start realizing this program in a consistent way, we will need some toy model simple enough from the purely technical perspective and sufficiently rich at the same time from the standpoint of its practical usefulness and further scalability. This model should lie somewhere on the border between intelligent, living and non-living systems and, to some extent, have features of all of them. Ideally, the role of such a model should be similar to the role played by the hydrogen atom model in physics or by the “Hello World!” program in programming. If one understands physics or programming at this level, then one probably has all the chances to understand the rest.

But how shall we find an adequate toy model suitable for our needs? Fortunately, things are not as complicated as they may seem at first glance. The model which may satisfy us at least at the initial stage does exist, is describable in a very simple way and can be thought of as a very natural extension of the good old Turing machine. Because of this similarity we will call our model the Autonomous Turing Machine.

Firstly, remember some basic facts about the standard Turing machine (TM) (Barker-Plummer, 2016), (Penrose, 1994). The TM is a hypothetical computing device living in a hypothetic memory space divided into an infinite number of cells. This device is capable of successively visiting diverse memory cells, reading information stored in them, processing it and writing the results back to the same memory space. The machine itself can be in a certain finite number of internal states. Its central processing unit – the so-called finite state automaton – operates as follows: it takes as an input the current state of a machine together with the symbol it has just read from the current cell and produces the output. The latter includes a) the new symbol which overwrites the old one in the same cell, b) the new state of the machine and c) the new move which is the instruction to go in a given direction or just to stop.

The initial content of the memory space is our assignment – this is what we want a machine to do for us, which includes both the data and the program – i.e., the instruction how to process the data. After completing all the computation, the machine stops. The content of memory space at that time is what we call the result of computation.

The importance of TM lies in the fact that it is capable of performing any computation and, at the same time, it is simple enough to be considered as a convenient playground for developing and testing diverse theories of computer science. One of the most important distinguishing properties of the traditional Turing machine is that it does something useful for us, i.e., its users. We write the programs for TM, we provide it with data, we start it, we wait until it finishes and we become nervous if it does not stop. The machine itself does not care what it does, why it does it and how it does it. It is simply a tool for serving our needs and simplifying our work in fulfilling these needs – it is like a shovel, hammer or bulldozer, but nothing more.

Can we use such a machine for programming artificial intelligence? Probably yes, but only to some extent – until we can anticipate in advance all the tasks the machine may need to perform for us. We may be very smart and build very smart machines which will fulfill all our current needs. But in any case, this will not be genuine intelligence because it assumes the ability to learn new things and – which is even more important – the ability to define new problems.

We can say that traditional computers help us to answer our questions. But what we intuitively expect from intelligent systems is the ability to ask questions. We want to deal with computing devices which would help us to be on the cutting edge of progress and drive it by creating new knowledge about nature. But if all this knowledge needs to exist in advance on the tape presented to Turing Machine then what would we expect that machine to do for us?

Here we consider another version of Turing machine which, as we think, is free of the problems we mentioned above. We can call it the Autonomous Turing Machine (ATM). The idea is to let the machine itself decide what it needs. The first thing the ATM may need is energy. Having the energy, it can move, perform calculations and maybe even produce something useful for us. But how can the machine find this energy? Do we need to equip it with batteries or power supplies? Not necessarily. Here is a good place to remember that information can serve as a fuel, so theoretically our ATM may find everything it needs in the memory space where it lives and moves.

This circumstance makes ATM similar to an autonomous robot placed in an unknown fuel-bearing terrain and trying to survive in it. To survive the robot needs to consume fuel. But to consume fuel it needs to find fuel deposits. But to find fuel deposits it needs to understand the patterns of its possible distribution – i.e., to learn from experience. Then it needs to move to the places where it thinks the chances of finding fuel are high. All these actions require energy which robot can extract only from the fuel it consumes. We arrive at the infinite logical loop: the robot consumes fuel to be able to act and acts to be able to consume fuel again. And this is what we call life. If we replace the word “fuel” with the word “information” and the word "robot” with the word "ATM”, we will get an idea of how the ATM’s dynamics in memory space may look.

## What we plan to discuss here

This cycle consists of the following three lectures:

1. The life of information
2. Intelligence: an internal point of view
3. Intelligence: an external point of view

A good prerequisite for this text would be some high-level knowledge of the existing facts linking the purely mathematical aspects of information to its physical, biological and business/economical aspects. The classical set of review articles and books summarizing these results in connection to physics includes (Bennett, 1982), (Landauer, 1991), (Zurek, 1991), (Plenio & Vitelli, 2001), (Leff & Rex, 2003), (Bennett, 2003), (Bais & Farmer, 2007), (Maroney, 2009), and (Feynman & Hey, 2000). An excellent overview of biological and sociological aspects of information can be found in (Adami, 1998), (Adami, 2012), (Adami, 2016 ), (Avery, 2003 ). The business and economic aspects of information form a huge domain. Despite the fact that we will be interested here only in its relatively small subdomain related to the predictive analytics and its practical role in solving diverse resource optimization problems, it is still too large and too diverse to be represented by a few references. I would recommend considering Wikipedia's article "Predictive Analytics" as an entry point of your search and use the references cited therein.

Lecture one is mostly based on the results obtained in the middle of the 20th century in attempts to resolve the famous Maxwell's daemon paradox and actually stating the equivalence between information and fuel. Freely rephrasing these results I will show how to use this equivalence for building information-driven engines capable of converting any predictable information into useful work. The main concluding messages of this lecture are: To create new information we need to spend energy. To extract energy from the existing information we need to know it in advance. The energy extraction process kills information – it becomes a waste. One can build information-driven engines that are tolerant to errors. The effectiveness of these engines is describable by Kullback-Leibler formula (Kullback & Leibler, 1951) for information gain. Different information patterns require different engines and vice versa: simply speaking, patterns and engines should match. In other words, there is no such thing as an absolute value of information – this value can only be defined and measured relatively to a certain engine, and this is probably the central point of this lecture. The rest of it is devoted to energy aspects of computation. The main message of this part is that computation, if properly organized, does not require any energy. It does not directly lead to any energy gains either. This raises the question why we may need computation if it does nothing. The answer is given in lectures 2 and 3.

In lecture two we provide an internal view on computation by discussing it from the standpoint of the computing device. We demonstrate that by using properly organized computations one can increase the initial energy value of practically any information reaching an energy extraction device specified a priori. In other words, computation can be viewed as a process of making information ready for the consumption by a given engine. In a sense, this process is similar to the process of cooking food before eating it. We call this procedure the *information refining process* and show that it lies in the basis of any statistical learning algorithm. Computation can also be considered as a unitary evolution operator applied to the input vector of registry bits. Being itself energy-neutral, it transforms this input vector into an output vector having maximal energy value, i.e., matching a given engine the best possible. The energy extraction procedure is, however, non-unitary and leads to a registry vector collapse – it forgets everything. This makes this process similar to what happens with the wavefunction in quantum systems after the measurement . The rest of this lecture is devoted to autonomous Turing machines considered from the standpoint of their internal organization. We describe the two main building blocks of these machines and outline their functionality. These two blocks are responsible for (i) understanding and (ii) making decisions. From the energy perspective, the "understanding block" maximizes energy accumulation while the "decision block" minimizes its spending. The optimal dynamics of ATM can be derived from the problem of statistical optimization of the accumulation and spending difference.

In lecture three we consider the problem of computation from the external point of view. As an example, we discuss autonomous Turing machines considering them as point-wise objects and show that the problem of finding the optimal trajectory of such ATMs in memory space reveals striking similarities with the famous Least Action Principle in the context of finding the optimal trajectory of a mechanical particle moving in an external potential. Basing ourselves on this equivalence, we will present the exact solution of this problem and discuss it in the context of diverse physical systems. The most interesting aspect of this discussion is the similarity between mathematical structures appearing in the description of the open and closed, living and non-living, and even physical and socio-economic systems. Then we will discuss some general questions related to the relationship between models of artificial intelligence, theoretical physics and business economics. We will show that all these systems can be treated and examined using the resource maximization principle which seems to be a natural generalization of the least action principle from closed conservative to open autonomous systems. It is interesting that this purely theoretical discussion may have practical implications, too. For example, the almost literal analogy between businesses and simple mechanical systems like swings allows one to conjecture that one of the ways of facilitating business growth could be based on the effects of parametric resonance which seem to be typical for both mechanical and economic processes.

As seen from this plan the subject I am trying to cover here is very cross-disciplinary and assumes a rather broad target audience including people working in diverse areas of physics, computer science, and business economics. For this reason, I tried to keep the style of this discussion at the highest possible level. If you think you will find some ready-for-implementation solutions, new algorithms or hardware architectures, you are mistaken. However, if you are ready to look at the existing problems from a somewhat different angle, these lectures are is for you.

Most of the facts mentioned in this text in connection to physics of information should not surprise people with background in theoretical physics or evolutionary biology working in the area of computer science, especially in business environment. There are many excellent reviews, books and research papers discussing this subject from different angles and partially overlapping with the factual material presented here. I have already mentioned some of them earlier.

Nevertheless, the goal of this text is not a simple exposition of known facts. This is rather an attempt of their unification into some new piece of knowledge which, as I believe, may have its own spectrum of practical applications. This is actually an attempt to approach the set of very specific, extremely complex and seemingly unrelated problems of artificial intelligence, theoretical physics and business economics from the positions of a certainly much broader but substantially simpler problem. I believe that such unification not only would allow one to simplify developing new methods in machine learning and applied business economics but also would give us another fresh view on the problems of theoretical physics. I could not resist a strong desire to find a systematic way of exposing these ideas as a whole, which gradually has crystallized in my decision to write this text and essentially determined its form as a cycle of lectures. These lectures have never been delivered to any real audiences – the chosen format simply reflects my attempts to organize my thoughts in a better way and share them with people from diverse industries and areas of research who may find this subject useful for any reason, be it theoretical or practical.

# Lecture 1. The life of information

## 1.1. Information as a fuel

Information can be defined as something that draws one's attention. To have such a capability, the state of this "something" must be distinguishable from that of its surroundings. According to the Second Law of Thermodynamics, this distinguishability makes the state potentially unstable. If we do nothing, then, sooner or later, the natural forces will remove the barriers keeping this state different from others and it will change to a state similar to other states around. The information will disappear. In nature, this process may take many different forms: for example, it could appear as erosion (if slow) or as explosion (if fast). What is important here is that the process of decay is always accompanied with the release of some amount of energy: this is the hidden potential energy of the unstable state transformed into the kinetic energy of the decay processes. If such decay happens in an uncontrollable way, then all this energy will be lost together with the information. But if we control the process, we can use the released energy for our needs, and this is what physicists usually call the "useful work". The device that achieves this in a systematic way is what we call the "engine". And the stuff intentionally converted in the engine into a useful work is called "fuel". Therefore, information and fuel are the same.

Of course, there are many other features shared by both information and fuel which can convince us in many different ways that these two are very similar. It is sufficient to remember, for example, such features as storability, controllability and transportability. Storability means that all the barriers preventing unstable states from decay must be strong enough to keep the latter unchanged for an indefinitely long time. Controllability means that there should be a simple way of removing these barriers at a negligibly small energy cost at any time as the need arises. And transportability means that the energy cost of moving the information or fuel from one place to another should be negligibly small compared to the energy value of what is being transported.

There is another important layer of similarities between information and fuel: both require a perfect match with the engine that uses them. If there is no match, or if it is insufficiently strong, we will not be able to extract any useful work at all. The reason for which sugar is a good food for dogs but bad food for cats has the same origin as why one and the same message may contain valuable information for one recipient and be completely meaningless for another one.

But all that is still an abstract theory. If information and fuel are really the same, there should be a simple, practical way of demonstrating the convertibility of information into useful work. Now we will try to show how to do so.

## 1.2. How to extract energy from a bit

Consider a cylinder of the length $L$ divided by a diaphragm into two equal parts and having a single atom inside. This system has two clearly distinguishable states: state one, when the atom is in the left compartment and state two, when the atom is in the right compartment. It represents the simplest memory unit with the capacity of 1 bit. Why 1 bit? Because to correctly identify one of these two states we need to answer exactly one question:

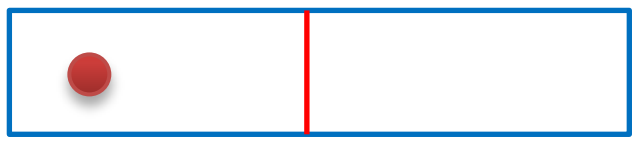
or
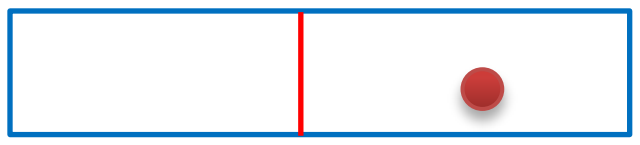
?

Now consider this memory unit from the standpoint of physics and demonstrate that we can use it as a simple source of energy. Consider one of these two states, say the state when the atom is in the left compartment and insert a piston into the cylinder from the right, as shown below:

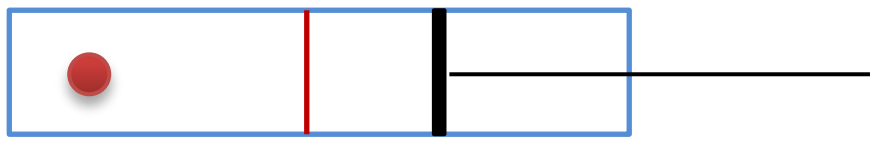

Since the right compartment is empty, we can freely move the piston to the left until it reaches the diaphragm.  Then we open it:

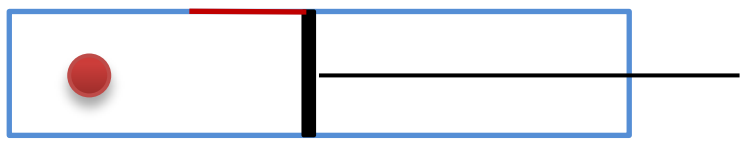

At this moment the piston will start experiencing a right-directed force derivable from the formula

$$pV = kT$$

in which $p$ is pressure, $V$ is the volume in which the atom moves, $T$ is the temperature of the thermal bath the cylinder is assumed to be in contact with and $k$ is the so-called Boltzmann constant. Representing the volume as $V = Sx$, where $S$ is the area of the cylinder's cross-section and$x$is the piston's position measured from the left, noting that $F = pS$ is the force acting on the cylinder, we can deduce the simple formula for it: $F = kT/x$. Remember that initially, we had the piston at the position$x = L/2$. Now, suppose that the piston slowly moves to the right. We need to move it slowly because in that case the temperature $T$ will not change. Each time the piston will move at a tiny distance $dx$ it will produce a tiny amount of useful work equal to $dE = Fdx = (kT/x)dx$. This work extraction process finishes when the piston reaches the position $x = L$. Using the formula for the force we just derived we can write the expression for the total work extracted:

$$E = kT \int_{L/2}^{L} \frac{dx}{x} = kT \ln 2$$

Note that the length of the cylinder has amazingly disappeared from the result! This circumstance which may look somewhat trivial in this particular case is a manifestation of a very deep and universal fact: It does not matter how we realize the 1 bit of information -- it is always capable of generating the useful work equal to $kT\ln 2$. Different memory realizations may, however, require different energy extraction mechanisms.  Indeed, in our case, the length of the piston (characterizing the energy extraction mechanism) should be correlated with the length of the cylinder (characterizing the memory cell).

After completing the cycle, we can close the diaphragm. However, we now cannot say where the atom resides: we have completely lost all the information about the atom's location by converting it into useful  work.

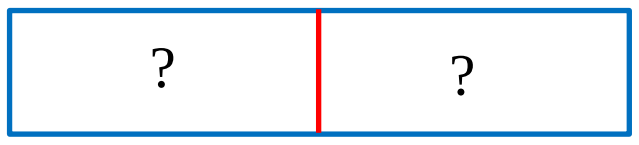

## 1.3. How to build an information-driven engine

We have already learned how to extract information from a single bit – now it is time to learn how to deal with large collections of such bits.

Assume that there are $I$ identical cells evenly distributed along a line. One way of extracting the energy from this line is to use $I$ identical pistons and insert them in $I$ cylinders simultaneously (i.e., in parallel). However what to do if the number $I$ is indefinitely large and we cannot get immediate access to all of the cells at the same time?  In this case the parallel solution will not work and we will need to use a more universal and less hardware intense serial solution. The idea is to use only one piston combined with a mechanism which would move along the line and step by step extract the energy from each of its memory cells. We want to make such a mechanism self-propelled. This means that the energy extractable from the sequence of memory cells should fully compensate all the losses associated with the motion of this mechanism along this sequence.

To achieve this, we need to convert the back-and-forth motion of the piston into some progressive motion. The simplest mechanism that allows one to do that is to use the inertial wheel whose inertia can play the role of an energy accumulator. One can use the accumulated energy in many different ways: to help the piston to pass its turning points, i.e., the points when it occupies the utmost right and left positions, to open the diaphragm when the piston reaches it, and to move the entire mechanism to the next available cell after finishing the whole cycle:

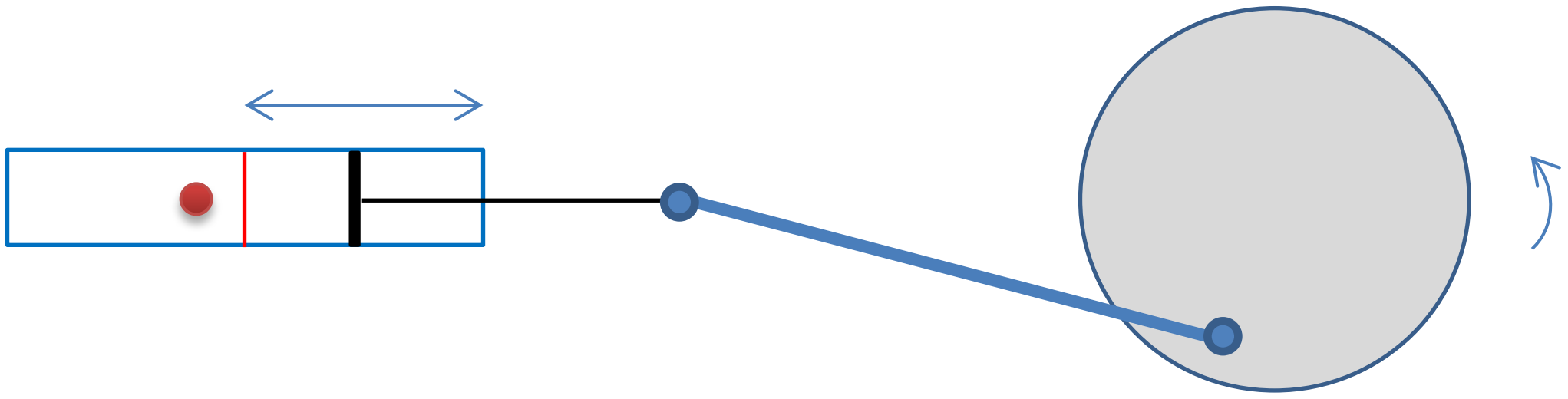

Summarizing, we can say that having some sequence of $I$ identically organized 1-bit cells (i.e., for example, having atoms always in the left compartment) we can extract from it the energy equal $E = IkT \ln 2$ by using the engine we just described above. The sketch of this extraction process is shown in the picture below:

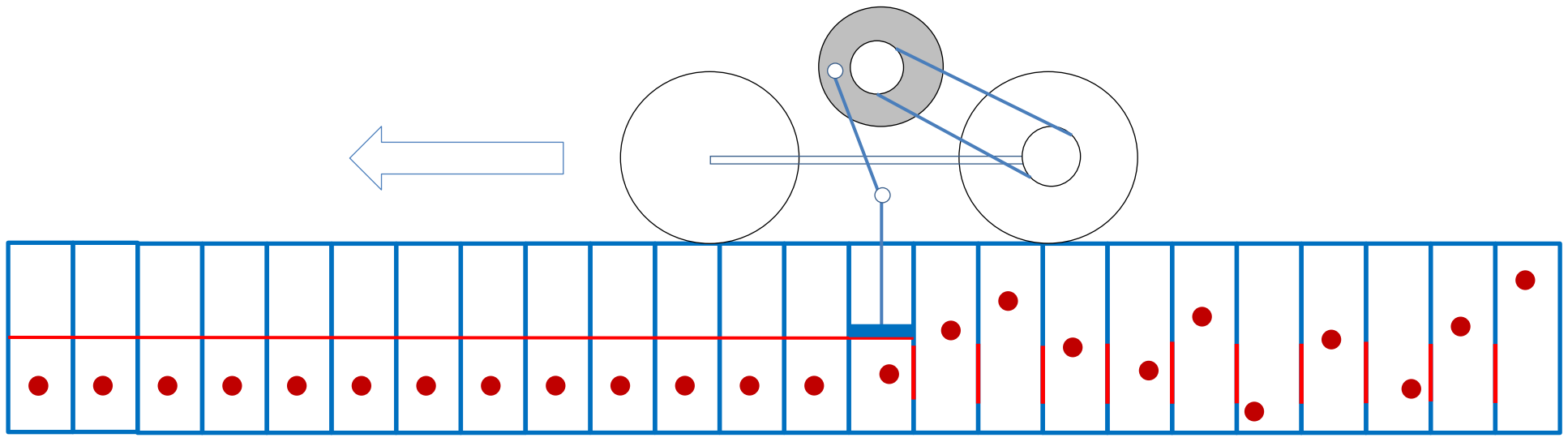

## 1.4. How much should we pay for our errors?

We have just demonstrated the usefulness of information by showing that it can be a source of useful work. However, this is only true if the information is *a priori* known. In our case, for example, we know that all the atoms are in the left compartment and therefore use the appropriate (right-handed) energy extraction device. But at the same time, we know very well that perfect knowledge (i.e., knowledge excluding any errors) is impossible. When predicting something we always make errors. What will these errors change?

Consider our engine again and imagine that we were wrong when we assumed that all atoms are in the left compartment and there is one cylinder having the atom on the right side. In this case, we cannot move the piston freely through the right compartment anymore because now this compartment is occupied by the atom. If we continue pushing the piston to the left, it will experience increased resistance from the atom. To overcome it, we will need to spend infinite energy.

This sad fact will prevent our engine to work well and will cause its stop even before the piston reaches the point ½. To demonstrate this we will consider our mechanism again but now with the atom residing in the right compartment:

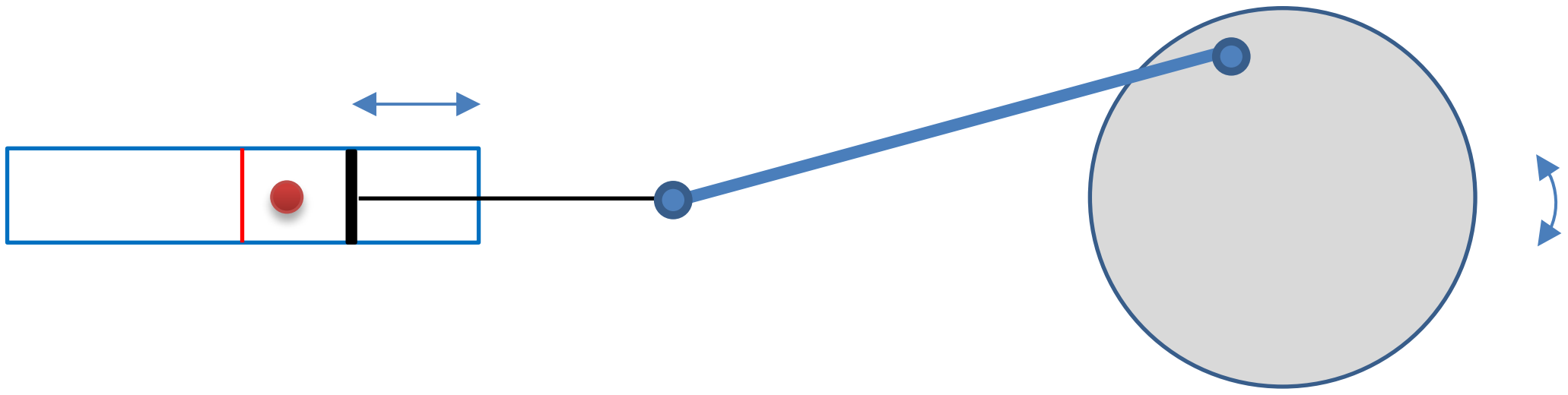

In the beginning, everything will look as before: the piston will start moving to the left. However, at some point the forces acting on the piston from the atom will exceed the inertial forces of the wheel and the latter will simply stop. Even worse: after stopping the wheel will start rotating in the opposite direction which will completely ruin all the process we have so carefully designed.

This leads us to a sad conclusion: the engine we just designed will not work because even one single error will be fatal for its normal functioning.

## 1.5. How to protect our engines from errors?

It is easy to see that the source of this problem is the singularity at $x = L/2$. In order to overcome this problem we can try to redesign our engine in such a way that the piston will not be able to reach the point $x = L/2$ with the diaphragm closed. So we will open the diaphragm earlier when the piston is located halfway towards the diaphragm at some point $x = M$, where $L/2 < M < L$. After simple calculations similar to those we did before we will find that if the atom is initially in the left compartment then the work performed by the system is

$$E_{\text{left}} = kT \ln \frac{L}{M}$$

Analogously, we can show that if the atom is in the right compartment, the work performed by the system is described by

$$E_{\text{right}} = kT \ln \frac{L}{M} - kT \ln \frac{L/2}{M - L/2}$$

Looking at the numbers $E_{\text{left}}$ and $E_{\text{right}}$, we can see that $E_{\text{left}}$ is always positive (energy gain) and $E_{\text{right}}$ is always negative (energy loss). However, since $M > L/2$, these losses are always finite. This gives us some hope that the vehicle we have just designed will continue moving forward even in the case of some minor errors it may encounter. So we have at least conceptually solved our problem with infinite energy losses.

If we look at the formulas we have obtained, we will see that at first sight they explicitly depend on the cylinder length. However, a closer look shows that they depend on the ratio of two lengths $L$ and $M$. If we introduce the quantity

$$Q = \frac{L}{2M}$$

then we arrive at the mathematical expressions for work balance in case we are correct and in case we are wrong:

$$E_{\text{left}}(Q) = kT \ln \frac{Q}{1/2}$$

$$E_{\text{right}}(Q) = kT \ln \frac{1 - Q}{1/2}$$

Note that although these two formulas look symmetric (i.e., relative to the exchange of $Q$ and $1 - Q$) they are actually not symmetric because the mathematical expression for $E_{\text{left}}(Q)$ is always positive and the one for $E_{\text{right}}(Q)$ is always negative. Technically, this asymmetry comes from the fact that $M < L$ and therefore $> 1 - Q$ . However, its actual physical origin is that we have used the right-handed device based on the insertion of the piston from the right. If we would select the left-handed device we would obtain opposite inequalities.

This leads us to the following important conclusion, which, in a sense, is a central point for understanding the rest. We claim that selecting the device is equivalent to defining what is correct and what is wrong, or, in other words, what is good and what is bad. If we are correct – we gain energy, which is good, and if we are wrong – we lose it, which is bad. Or, conversely, if a given device interacting with a certain state leads to energy gain (or respectively loss) then we can *define* this state as correct, true (or respectively wrong, false). We can even name these states TRUE or FALSE or just use any other symbols like 1 or 0 for distinguishing between them. Here is where numbers first come into play. And note that they appear not simply as logical symbols but as symbols representing some values: we can even order them by saying that $0 < 1$ simply because 1 symbolizes energy receipt and 0 – energy loss. For the same reason, 1 is good (for a system) and 0 is bad (for a system).

We are probably at the right place to stress that these values have value for a system only – not for us. The values are always subjective – there is no absolute notion of what is good and what is bad: it can be defined only relative to a given system and this definition is more than pragmatic. The way the world appears to a system – i.e., what is good and what is bad for it – depends exclusively on which devices it uses for data extraction. Two different systems equipped with two different data extraction devices will perceive the world differently and may have different, even opposite, opinions about the values of the same things around.

## 1.6. Towards optimal engines

In previous sections, we explained how to build information-driven engines and how to avoid infinite energy losses in case of errors. Now we need to go further and understand how to build optimal engines. To achieve optimality we try to find those values of $Q$ at which the losses are minimal.

Assume that our engine moves along the line and the probability of visiting cells having the atom in the left compartment is $R$. In this case, the total energy balance for this engine will be described by

$$E(Q) = RE_{\text{left}}(Q) + (1 - R)E_{\text{right}}(Q)$$

or, equivalently, by

$$E(Q) = kT\left[R\ln\frac{Q}{1/2} + (1 - R)\ln\frac{1 - Q}{1/2}\right].$$

We can treat the first and second terms of the last mathematical expression as the consumption and spending terms, respectively. The first term is always positive and the second one is always negative. This is quite obvious. What we desperately need is the positivity of the entire expression

$$E(Q) > 0$$

because the engine is efficient only if its total consumption spending balance is positive.

But how to build such an engine? One thing we can easily try to do is to optimize the position of piston's turning point by properly selecting parameter $Q$. In other words, we need to maximize the expression for $E$ with respect to $Q$, which gives us the unique solution

$$Q = R.$$

Substituting the found value of $Q$ back into formula for $E(Q)$ we obtain

$$E_{max} = \max_{Q} E(Q) = kT \left[ R \ln \frac{R}{1/2} + (1-R) \ln \frac{1-R}{1/2} \right]$$

Note that up to the purely "physical" factor $kT$ this is the famous Kullback-Leibler divergence (Kullback & Leibler, 1951), also known in information theory as "information gain". The numbers 1/2 and $R$ are usually treated as "prior" and "posterior" probabilities of a certain event and reflect our knowledge about that event before and after its actual examination. In our case, the role of this event is played by finding an atom in the left compartment.

The most amazing feature of this expression is that it is always positive. This fact formally follows from the so-called Gibbs inequality. For us, the practical meaning of this fact is that for any distribution of information stored in this memory and characterized by the $R$ parameters, we always can find the right action characterized by the $Q$ parameter that will allow us to consistently extract positive energy from that system.

## 1.7. From engines to generators

In this section, we will consider another device which is totally opposite to the engine – the generator. In our case, the role of generators is to convert work back to information, i.e., to create new information. Remember that the engine's output was a memory cell in which the location of an atom was totally unknown. Since the generator as a device is opposite to the engine, this memory cell should be treated as its input. So consider the cylinder and assume that the position of the atom in it is initially totally unknown.

Our goal is to contract the atom in the left compartment regardless of its original location. Such a contraction will make the atom's position completely known and will therefore result in creating 1bit of information. The task looks pretty straightforward, but if we try to perform it practically, we will face the same difficulties as we had when we tried to build engines.

Indeed, consider two opposite scenarios:

1. The diaphragm opens when the piston reaches it. There is no problem in implementing this scenario if the atom is in the left compartment. However, if it is in the right compartment, the piston, in order to reach its final position at $L/2$, will need to overcome infinite resistance and spend an infinite amount of energy. So this scenario is absolutely unacceptable.

2. The diaphragm opens when piston just starts its motion towards it from the position $L$. In this case, we spend only $kT\ln 2$ of energy no matter where the atom is initially located. At first sight this amount may seem not too big. However, if the atom is already in the left compartment and the diaphragm does not open, the piston could reach its final position at $L/2$ without spending any energy at all. This means that this scenario is not optimal either.

From these two examples it follows that the optimum lies somewhere in the middle: the diaphragm opens when the piston reaches some intermediate position $M$ lying just in the middle between its initial position at $L$ and its final position at $L/2$. If we repeat all the reasoning of the previous sections, we will find that the optimal value of $Q$ at which the energy loss is minimal is achieved at $Q = R$, which leads to the following expression for the energy:

$$E_{max} = \min_Q E(Q) = -kT[R\ln R + (1-R)\ln(1-R)].$$

This expression is always positive which means that in either case we need to spend some energy to generate information. As we can see, there is a simple formula connecting the minimal work spent the in information generator and the maximal work extractable in the information-driven engine:

$$E_{max} + E_{min} = \min_Q E(Q) + \max_Q E(Q) = kT\ln 2$$

## 1.8. The asymmetric memory cells

Assume that we have an asymmetric cylinder of length $L$ divided into two unequal compartments 1 and 2 of lengths $K_1$ (left) and $K_2$ (right), respectively. Assume that we believe that the atom is in the left compartment and we want to benefit from that belief by converting it into useful work. This belief not necessarily reflects the reality – it could be simply wrong. Denote by $R_1$ the probability that it is correct. Then the probability that we are wrong is given by $R_2 = 1 - R_1$. If we repeat the reasoning given above for the symmetric case, we will arrive at the generalized version of formula for energy gain

$$E = kT\left[R_1\ln\frac{Q_1}{P_1} + R_2\ln\frac{Q_2}{P_2}\right]$$

describing the average energy extractable from the memory cells with non-equally sized compartments. The meaning of the parameters $P_1$,$P_2$, $Q_1$ and $Q_2$, is described below.

The parameters $P_1$ and $P_2$ are defined as

$$P_1 = \frac{K_1}{L}, \qquad P_2 = \frac{L-K_1}{L}$$

and mean the proportions in which the diaphragm divides the cylinder. For that reason, we can treat them as probabilities of finding the atom in the compartments 1 and 2 respectively. These two parameters, $P_1$ and $P_2$, are obviously external to our system; the environment fully controls them, and the system can only try to accommodate to their existing values.

As to the parameters $Q_1$ and $Q_2$, they are defined as

$$Q_1 = \frac{K_1}{M}, \qquad Q_2 = \frac{M - K_1}{M}$$

where $M$ is the position of the piston's turning point. Thus they reflect the way the energy extraction mechanism is organized, which allows one to interpret these parameters as totally internal to our system and therefore fully controllable by it. Note that the above formula also allows to treat these parameters as proportions in which the old diaphragm (which, as before, occupies the position $K_1$) divides the new length $M$ (which is shorter than the length of a cylinder) into two compartments. This, in turn, allows one to treat them as probabilities. But what is the physical meaning of these probabilities?

At first sight, there should be no big problem with answering this question because the definitions of the $Q$ and $P$ parameters are very similar. The only difference is that in the case of the $Q$- probabilities they are measured not relatively to the total length of the cylinder $L$, but rather relatively to its contracted (rescaled) version $M < L$. But there is just a problem with $M$ because there is no clear picture of what the physical meaning of this contracted length is. Yes, we know that this is the point at which the piston should stop and then turn back but this is not something what physicists would be aesthetically satisfied with – this explanation is too device-dependent.

Fortunately, there is an interesting circumstance that may help us to avoid answering this tough question about $M$ and rephrase the entire problem in an absolutely different way. The point is that the formula for energy extractable from information does not contain the length $M$ at all! The formulas for probabilities are scale-invariant: if we rescale both numerators and denominators by the same factor, nothing will change. Let's select this scaling factor as

$$\alpha = \frac{L}{M}.$$

Then, after applying it to fractions defining parameters $Q_1$ and $Q_2$, we obtain for them:

$$Q_1 = \frac{M_1}{L}, \qquad Q_2 = \frac{M_2}{L},$$

where $M_1$ and $M_2$ are defined as

$$M_1 = \frac{K_1 L}{M}, \qquad M_2 = \frac{(M - K_1)L}{M},$$

and thus satisfy the condition:

$$M_1 + M_2 = K_1 + K_2 = L.$$

Now the meaning of these two numbers becomes clear: they simply describe a different partition of the same cylinder of the length $L$. This partition defines a different position of the diaphragm and different probabilities of finding an atom in the left and right compartments. But how to combine this fact with the existing architecture of our engine? It is easy to see that this is impossible and, if we actually want to keep the similarity between the interpretations of the parameters $P$ and $Q$, we need to redesign our engine, too. In next section we show how to do that.

## 1.9. Towards simpler engine

The fact that transition from $P$ to $Q$ probabilities is equivalent to a simple change of partition is more important than it may seem at first sight. It allows us to build an energy-extraction engine that does not require any complex cylinder-opening, piston-inserting and diaphragm-opening mechanisms. Everything becomes much simpler now. The only thing what we will need to do now is to perform one and the same motion each time we get an access to a new cylinder: simply move its diaphragm from the old position$K_1$ to the new position$M_1$without opening it as shown in the figure below:

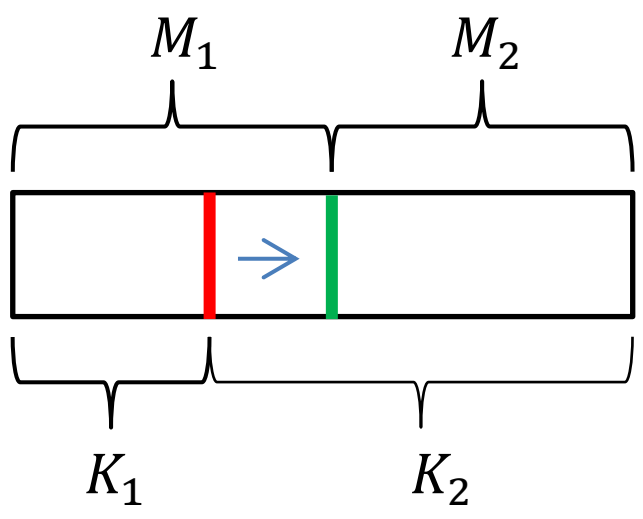

Here the colored vertical lines represent the initial (red) and final (green) positions of a diaphragm.

And we can easily see that the net result will be the same as we had before. Indeed, assume that the atom is in the left compartment. Then, if we move the diaphragm from position $L_1$ to position $M_1$ the energy change will be equal to

$$E_1 = kT \ln \frac{M_1}{K_1} = kT \ln \frac{Q_1}{P_1}$$

Similarly, if the atom is in the right compartment, then, the same move of the diaphragm will lead us to the energy change equal to

$$E_2 = kT \ln \frac{M_2}{K_2} = kT \ln \frac{Q_2}{P_2}$$

Because of the two inequalities

$$M_1 > L_1 \quad M_2 > L_2$$

the energy deltas $E_1$ and $E_2$ cannot be both positive at the same time. Now, if we introduce the actual probabilities $R_1$ and $R_2$ of finding the atom in the left compartment, then we will obtain exactly the same energy formula as we had before.

## 1.10. The minimax principle

Another reason why a picture based on identifying the piston with the diaphragm is justified is that it treats both parameters $P = (P_1, P_2)$ and $Q = (Q_1, Q_2)$ equally. However, they are not completely equal. In some sense they are even opposite because the formulas for $E$ are antisymmetric with respect to their exchange. Let us discuss this in more detail.

First of all, we can see that in the general, asymmetric case, the information-to-energy conversion formula we derived above

$$E(Q,P) = kT\left[R_1 \ln\frac{Q_1}{P_1} + R_2 \ln\frac{Q_2}{P_2}\right]$$

has a maximum with respect to $Q$ attainable when

$$Q_1 = R_1, \qquad Q_2 = R_2.$$

The corresponding maximum value

$$E_{max}(P) = \max_Q E(Q,P) = \ kTI(R|P)$$

is always positive (as before) and is related to the general Kullback-Leibler divergence

$$I(R|P) = R_1 \ln\left(\frac{R_1}{P_1}\right) + R_2 \ln\left(\frac{R_2}{P_2}\right).$$

Now, consider the information-to-energy conversion formula again and note that $E(Q,P)$ has no maximum with respect to $P$ (can be made arbitrarily large if $P \to 0$ or $P \to 1$). However it has a minimum attainable when

$$P_1 = R_1, \qquad P = R_2.$$

and also related to Kullback-Leibler divergence as

$$E_{min}(Q) = \min_P E(Q,P) = \ -kTI(R|Q)$$

This minimum is always negative. It is not difficult to see that there is a minimax (or maximin) principle stating that

$$\min_P \max_Q E(Q,P) = \max_Q \min_P E(Q,P) = 0.$$

The extremal values of this double-optimum are attained at

$$Q_1 = P_1 = R_1, \qquad Q_2 = P_2 = R_2.$$

How to interpret this principle? What is the meaning of minimizing the expression for $E(Q,P)$ by $P$? The system we are discussing here is not interested in minimizing its energy balance at all. Even more, as we already noted it simply can't do that, because it has no control over parameters $P$ belonging to the environment.

The answer can be obtained if we note that formula for $E(Q,P)$ is antisymmetric with respect to the exchange of $P$ and $Q$:

$$E(Q,P) = -E(P,Q)$$

This means that $P$ plays the same role for the environment as $Q$ for the autonomous system we are considering. To make it more clear, imagine for a moment that the environment does have the same rights as our system does. In other words, we can treat it just as another system which interacts with our system and has its own motivation to survive. Then it will become clear that the environment would be interested in increasing its own energy balance. But everything what our system treats as consumption (energy gain) the environment will regard as nothing but spending because it means energy loss to it. And, conversely, if the system loses the energy, the environment gains it. But this means that the environment would be interested in minimizing the function $E(Q,P)$ with respect to all the parameters it has control of, i.e., $P$.

So we come to the following picture: Two systems (one of which is our engine and another one – the environment) fight for the same energy resource. Each system tries to increase its total energy balance by manipulating with those variables it has control of. In the ideal case, the result of this fight is zero, and this is what the minimax principle is telling us about.

## 1.11. How many engines do we need?

Assume that we have a sequence of memory cells evenly distributed along the line and storing the same information, for example, the number 1

$$\ldots 11111111111111111111111111111111111111111111111111111111111111 \ldots$$

In previous sections we have demonstrated that if we consider some physical realization of this information, for example, realize it by means of atoms located in the left compartments of identical cylinders, then it is possible to design an engine (based on the insertion of a piston into each of these cylinders from the right) capable of extracting the energy $kT \ln 2$ from each of these bits.

This situation is fully symmetric. If instead of the number 1 the memory cells store the number 0

$$\ldots 00000000000000000000000000000000000000000000000000000000000000 \ldots$$

then the same energy effect can be achieved if we use a mirrored engine (based on the insertion of a piston from the left).

Consider now the mixed situation when we have two types of bands of memory cells of the same length $I$: one storing numbers 0 and another one storing numbers 1. Assume that these bands form a strictly intermittent sequence as shown below for the case $I = 10$:

$$\ldots 111111111100000000001111111111000000000011111111110000000000 \ldots$$

How can we extract energy from such a chain? It is obvious that we need two engines in this case – one right-handed (for numbers 1) and another one left-handed (for numbers 0). Only one of these two engines can be active each instant of time. So we need a swapping mechanism that will periodically swap the engines to ensure that the engines and their fuels (in our case the values of binary numbers) do match. Since the sequence is strictly periodic – the bands of 0s and 1s are of the same length - we can easily design such a mechanism. For example, we can connect the inertial wheel we used above with another (slower) wheel which makes 1 whole circle while the inertial wheel makes $I$ such circles. Each time the slow wheel completes a circle the swapper swaps the engines. This procedure obviously does not require any energy, which means that the amount of extractable energy in this case will be the same, namely $kT \ln 2$ per bit.

But what if the sequence is not strictly periodic? What if the only thing we know about it is that the average length of 0-bands and 1-bands is equal to $N$?

$$\ldots 111111000000000000001111000000111111111111111100011111111111 \ldots$$

In that case, our engine-swapping mechanism will not be coherent with band distribution and we will get no energy gain. How can we improve the situation? We can do the following: attach to the slow wheel we just designed an additional measuring device which will check the number stored before swapping the engines. The problem with such an approach is that it will cost us some energy. We can understand that by analogy with bombs. To make sure we are not dealing with a fake bomb we will need to detonate it, but if we detonate it, we cannot use it anymore! The same here: to figure out which number in the memory cell is stored we will need to perform a measurement. It can be demonstrated that to perform such a measurement, we do not need to open the cell and destroy the information it contains. But we need some place to store the result of our measurement. If the results of our measurements are completely unpredictable then, as we know from the previous section, updating the information content of this additional memory cell will cost us $kT \ln 2$ of energy per bit. In previous case the process was periodic and thus totally predictable. Therefore it did not cost us anything. But now, having random distribution of band lengths, we are forced to pay for it. But how much?

To make a simple estimate note that if the measurement occurs every $I$ step, this means that every $I$ steps we loose $kT \ln 2$ of energy. If the average length of a band is $N$, then the number of such measurements per band will be $N/I$. At the same time, the average length of each band with wrong cells will be $I/2$. So the total energy loss in that case will be

$$E_{\mathrm{loss}} = \left(\frac{N}{I} + \frac{I}{2}\right) kT \ln 2.$$

The value of $I$ minimizing this expression is

$$I = \sqrt{2N}.$$

Thus, the minimal energy loss per band is

$$E_{\text{loss}} = kT \ln 2 \sqrt{2N}.$$

This loss is negligibly small in relative units (i.e. measured in losses per bit):

$$\frac{E_{\text{loss}}}{E_{\text{gain}}} = \frac{kT \ln 2 \sqrt{2}}{\sqrt{N}} \to 0, \qquad \text{if} \quad N \to \infty.$$

In the next sections, we show that this result can be improved. But what we need to know now is the fact that if we have a random macroscopic sequence of bands of 0s and 1s of arbitrary but sufficiently large length, we can extract from such a sequence asymptotically the same energy as from the sequence consisting exclusively of 0s or exclusively of 1s.

In conclusion, note that any presence of errors worsens this result. So if the bands of 0s and 1s are imperfect and contain some small portions of wrong numbers (just 1s and 0s) then, if we denote their amount by$\varepsilon$, from the Kullback-Leibler formula we get the order of these losses per bit:

$$\frac{E_{\text{loss}}}{E_{\text{gain}}} = kT \, \varepsilon \ln 1/\varepsilon$$

which is also negligible small as $\varepsilon$ tends to zero.

## 1.12. Between birth and death

In previous sections, we considered two basic life events associated with information – its birth and its death. The birth of information is the process of converting disorder into order and its death is the opposite process of converting order into disorder. The birth and death of information are two extreme cases of the processes which we can represent graphically as:

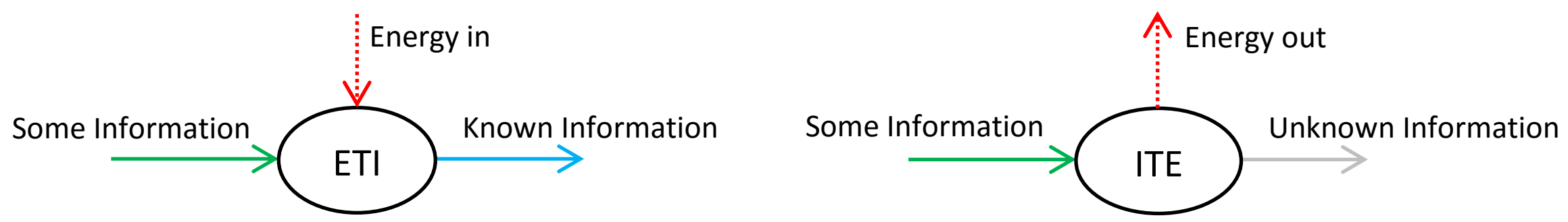

Here the blue and gray output lines represent two extreme cases of information flow: the flow of perfectly known information is represented by a blue line (the corresponding probabilities are either $R = 0$ or $R = 1$) and the flow of a completely unknown information is shown as a gray line (in this case both probabilities are $R = 1/2$). The green input line represents a certain intermediate case – i.e., the case when the incoming information is partially known. And finally, the red line depicts the energy (useful work) which is either spent in the ETI block or extracted from the ITE block. Here the acronyms

ITE and ETI stand for the "energy-to-information" and "information-to-energy" conversion processes respectively.

When talking about the two extreme cases representing the birth of information and its death, we mean those cases when the input green line representing incoming information becomes either blue (which corresponds to perfectly known information) or gray (which corresponds to completely unknown information). In the first case the ITE block becomes an information destroyer producing maximal energy output, while in the second case the ETI block becomes the information creator requiring maximal energy input. This relation of a system to information birth and death creates its somewhat counterintuitive behavioral strategy: the system tries to minimize the birth rate of information and maximize its death rate.

Here we are going to discuss some other things that may happen with information between its birth and death and which comprise what we call the computation process. To better understand the energy aspects of computation it is convenient to start with a very simple example that, technically, has nothing to do with computation but may help us to understand its main problem.

Assume we have 1 bit of information realized as an atom residing in the left compartment. We know that inserting the piston from the right and performing all manipulations described in the previous sections we will be able to extract the $kT\ln 2$ of useful work at the cost of complete loss of information. But what if we simply open the diaphragm and then close it again after some time? The information will be lost as before but in this case with no benefits for us – along with losing the information we will lose the opportunity of using it for our needs. Graphically this can be represented as a circle – the "lost opportunity device" – which has one blue input but no blue outputs.

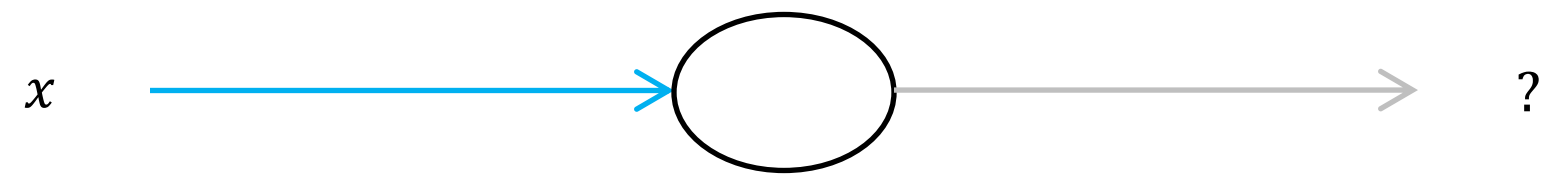

The same situation may occur in real computation. Indeed, we can treat computation as a process of updating the collection of system registers. Each of these registers we can model as a memory cell which we discussed above and which may be in one of two states: 0 or 1. The computation starts with some initial state (which encodes our question to computer) and ends with some final state (which encodes the computer's answer). The algorithm breaks this big question-answer pair down into a sequence of smaller and simpler question-answer pairs. Since answers are usually shorter than questions, one often deals with the situation when the number of registers needed to formulate the question (the input for a certain step of calculation) is larger than the number of registers needed to store the answer (the output from the same step). The best example of such a situation is the XOR operation $x \oplus y$. We initially need two registers that store the numbers $x$ and $y$ while at the end we need only one register to store the result. We may have the second register but, since it is not needed, we do not pay attention to it. As a result, all the information contained in it is lost. We can depict this process by using a diagram similar to the previous one but having two blue inputs and only one blue output. The second output will be gray:

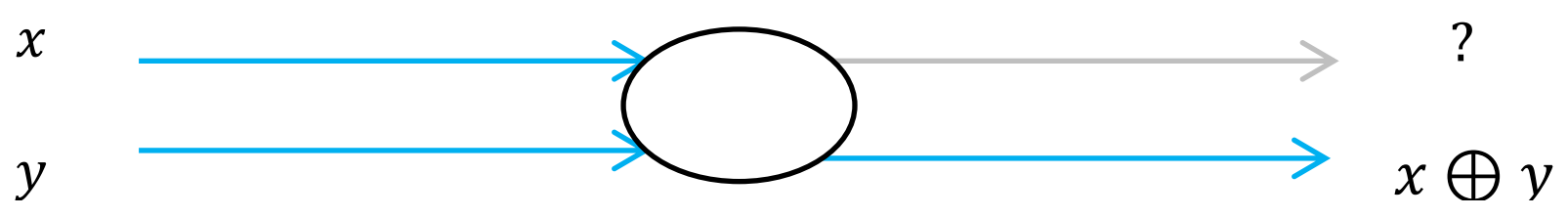

At first sight, it is nothing wrong with having gray lines with unspecified (random) information in the output. Like in the previous case (one blue input – one gray output), we have not lost any energy, so why should we care about such cases? The point is that the problem may (and will!) arise later, when we will decide to use this randomized register for storing something else – some other information. But this is nothing but what we just have called the information birth, which, as we know very well from the above discussion, always requires spending energy! So we can conclude that information loss at any step of computation is highly undesirable – it will lead to energy loss, which we should avoid whenever possible.

The practical criterion of information loss is very simple: if the input of some computation is reconstructable from the output, then the information is conserved. Otherwise, it is lost. In other words, logical irreversibility leads to energy dissipation, while logically reversible operations allow physical implementations not requiring any energy costs. So to avoid information and energy dissipation we should make all computations logically reversible. We can achieve this by using the so-called reversible logical gates. In the next sections, we consider some examples of such gates.

## 1.13. Saying "no" is easy!

The simplest reversible logical gate is the NOT gate. It has one input and one output and performs the negation operation

$$x \rightarrow \sim x$$

Since the operation is invertible: $\sim\sim x = x$, it prevents us from information loss. The energy is also conserved. To demonstrate this we will need two wheels (let us mark them by letters X and Y) connected with each other as shown in the picture below:

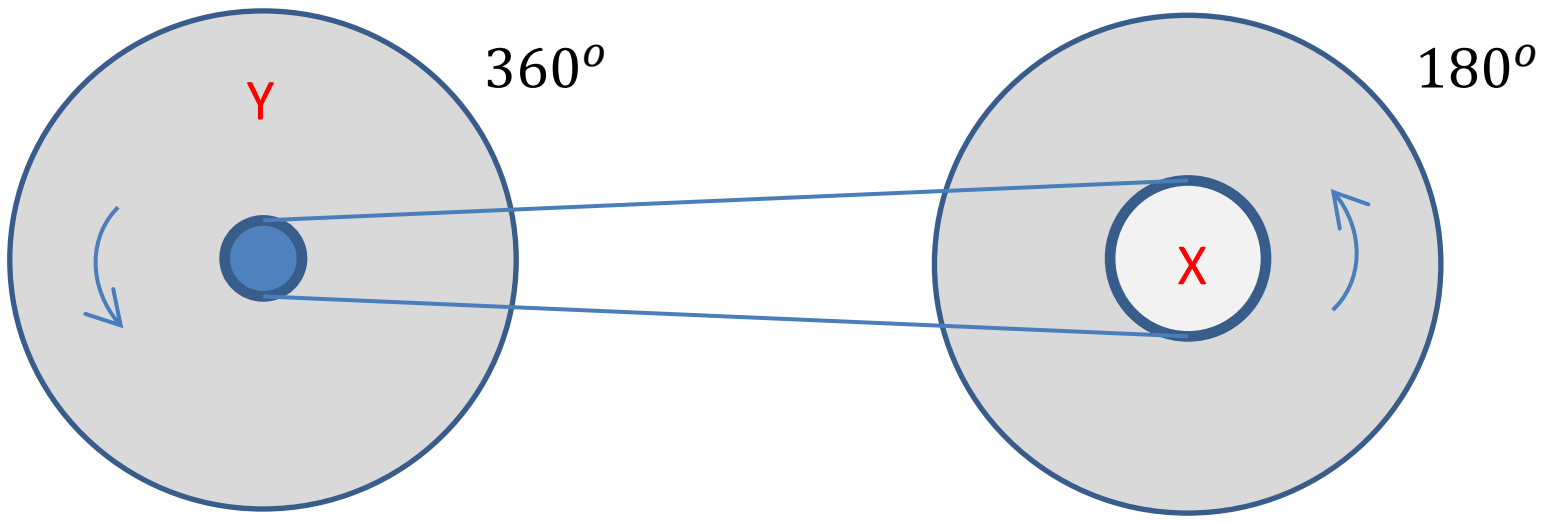

We calibrate the wheel Y in such a way that to make its initial orientation (shown in the picture with the correctly oriented red letter 'Y' on the top) slightly preferable than others. For that, we introduce an infinitesimally small asymmetry in the system, by adding to it some tiny attractor and allowing some tiny friction. The goal is to ensure that after slightly pushing the wheel in the counterclockwise direction it will complete exactly one cycle and stop in the same position as started. The second wheel X is assumed to be symmetric and not experiencing any friction forces which allows it to rotate freely infinitely long time.

Now we will connect the wheels Y and X by requiring that when wheel Y rotates for 360 degrees, the wheel X rotate only for 180 degrees. It is easy to see that this device will work as a negation operator. Indeed, placing the 1-bit memory cell we discussed above on the wheel X and slightly pushing the wheel Y in the counterclockwise direction we will change the atom's position to opposite irrespective of its initial position in the cell. This transformation will neither be accompanied by any energy loss, nor will it require the opening of the cell.

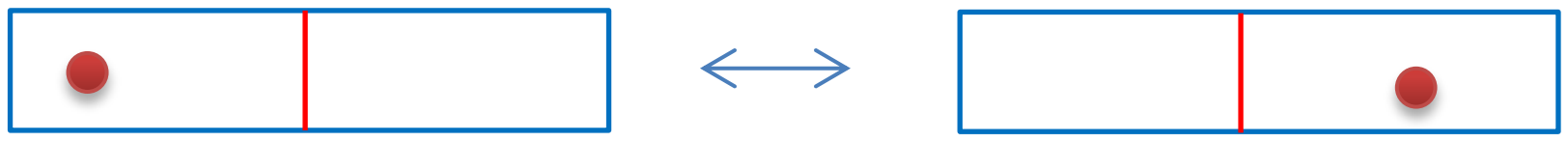

## 1.14. But… how to control our "no"s?

The negation operation we just considered was a unary reversible operation having exactly one input and exactly one output. What about binary reversible operations having two inputs and two outputs? The simplest operation of this type is the so-called CNOT (controlled-NOT) gate which is defined by the following relation:

$$\binom{x}{y} \to \binom{x}{x \oplus y}.$$

We see that this operation is an invertible version of the XOR operation we considered above. The gray line in the diagram (the second register) is now occupied with number $x$ and is not gray anymore. As it is easy to see this operation is totally invertible, because knowing $x$ and $x \oplus y$ one can uniquely reconstruct both $x$ and $y$. Theoretically such an operation should not lead to energy loss. How to demonstrate this? First of all note that this relation breaks down into two sub-relations

$$\binom{0}{y} \to \binom{0}{y}, \quad \binom{1}{y} \to \binom{1}{\sim y}.$$

The first one is the identity. We should be OK with implementing the second one because we have already done that by using the system of two wheels. The only challenge is how to incorporate the second variable $x$ in this picture. How to tell the mechanical system that it should negate $y$ only if $x = 1$, and if $x = 0$ it should leave $y$ unchanged. Let us demonstrate that this can easily be done if we add to the system the piston-based mechanism similar to the one we already considered before in connection with information-driven engines.

Let's attach the piston to the wheel and ensure that all proportions of the device are such that two turning points of a piston are at positions $L/2$ and $L$ as shown in the picture below:

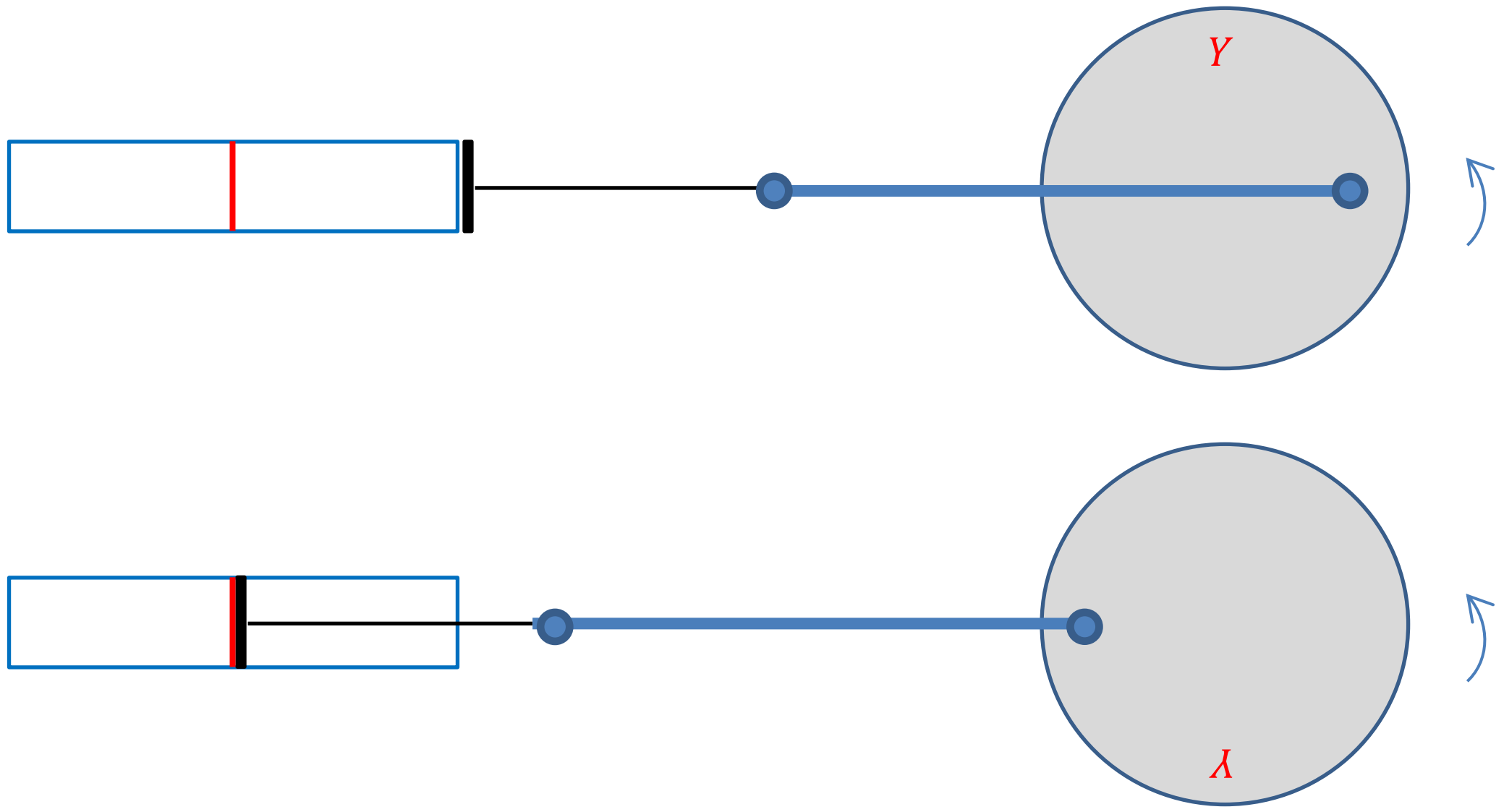

Assume that the atom is in the left compartment. Let us slightly push the wheel counterclockwise and wait until it passes its turning point at $L/2$ and returns to the initial position. What is important here is that the right compartment is empty and nothing prevents the piston from reaching it, turning back and completing the cycle. Now assume that the atom is in the right compartment.  Again, as before, we start by slightly pushing the wheel counterclockwise from its starting position. The piston will move to the left but will experience increasing resistance (because the right compartment is occupied by an atom). As a result, the piston will not be able to reach the diaphragm and will turn back somewhere in the middle. The wheel will also change the direction of its rotation and will return to its initial position without completing the whole circle. In both cases the final result of this second operation will be indistinguishable from the result of the first one. However, in the first case the wheel will complete a rotation for 360 degrees and in the second case its net rotation will be 0 degrees only. What is important here is that in either of these two cases we did not open the diaphragm and no mechanical work was spent or obtained. The state of the memory cell has not changed either. Now, if we connect the wheel  A to the wheel  X  as we already did before, we will obtain the desired embodiment of the CNOT operator that will be able to negate $y$ or leave it unchanged depending on the value of $x$. Remember we agreed to take $x = 1$ if the atom is in the left compartment (negation is performed), and $x = 0$ if the atom is in the right compartment (no negation is performed).  And let us stress once more that no energy is required to perform all these manipulations. In other words, the CNOT operation can be performed in an energy-neutral way.

## 1.15. Computation as permutation

Unfortunately, the logical operations considered above are not sufficient to perform *any* computation. For that purpose we can use another reversible logical gate invented by Edward Fredkin in 1981 (Fredkin & Toffoli, 1982). This gate works as follows:

$$\begin{pmatrix}0\\x\\y\end{pmatrix}\to\begin{pmatrix}0\\x\\y\end{pmatrix},\quad \begin{pmatrix}1\\x\\y\end{pmatrix}\to\begin{pmatrix}1\\y\\x\end{pmatrix}$$

The universality of Fredkin gate can be demonstrated by explicitly constructing all standard logical operations like CPY, AND, OR, XOR, AND, and NOT (Kerntopf, n.d.), (Perumalla, 2014), which are known to form a sufficient toolset for performing any computation. This means that any imaginable computation can be performed by using the Fredkin gates only. From the fact that Fredkin operator is an elementary permutation and any combination of elementary permutations is a permutation again it follows that any computation can be considered as a particular case of a general permutation. The number of zeros and ones in such permutation is a conserved quantity.

To physically implement Fredkin gate we need to implement the swap operator acting on two binary objects $x$ and $y$ as $(x, y) \to (y, x)$. This swap should be controlled: it should be performed only if the third variable $z$ is 1 and shouldn't when it is zero. But since we already had such situation when we tried to implement the controlled NOT gate, the implementation of Fredkin gate should not encounter any difficulties. The only difference with the CNOT case is that now the wheel X will be used for blind swapping in which the nature of the swapped objects is inessential. These could be any objects including the memory cells storing the numbers $x$ and $y$.

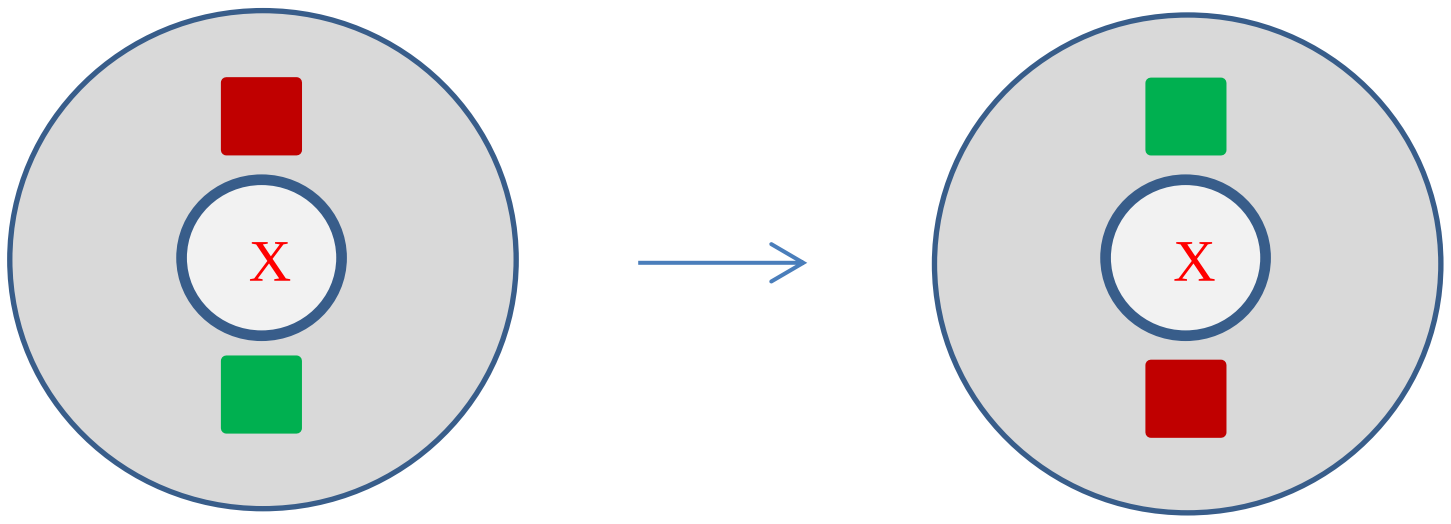

Let us summarize everything that we have learned so far. We can formulate this in three sentences:

1. The birth of information is always associated with energy loss.
2. The death of information always creates an opportunity to gain some energy.
3. Any computation could be performed in energy neutral way – with no losses and gains.

By analogy with the two blocks graphically describing the birth and death of information we can introduce the third block symbolizing energy-neutral computations. We call it the ITI block where the acronym ITI stands for "information-to-information". In graphical representation of this block the energy arrow is missing:

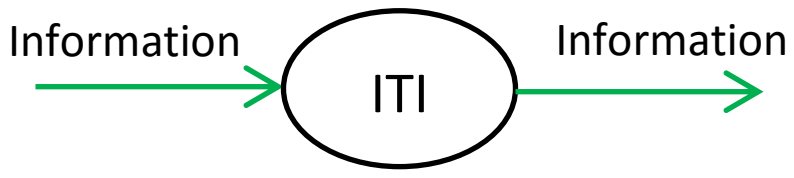

We have already discussed the ETI and ITE blocks. Let's now focus on the ITI block.

# Lecture 2. Intelligence: an internal point of view

## 2.1. Why do we need to compute?

In Lecture 1 we have demonstrated that computation can be viewed as a sequence of permutations of binary numbers. In this Lecture we will consider this fact from a practical standpoint. The main question we plan to address here is what is the main point of performing such permutations. Why should we consider this seemingly useless activity as something deserving our close attention? After all, what is computation for? Why is it so important for us?

To answer this set of questions let us assume that a system's memory is linear, so all its registers are physically arranged along a line. The distribution of 0s and 1s between these registers will give us some linear pattern which can be represented as a string of binary numbers. In this language, when talking about the initial state of a system, we will actually mean a certain binary string with some initial distribution of 0s and 1s. Each computation step will consist in taking a certain pair of these numbers and either permuting them or leaving unchanged depending on the value of a third (control) binary number from the same string. This, obviously, will not affect the total amounts of 0s and 1s. After some number of such steps the distribution of binary numbers in the string will change and, most likely, will have nothing to do with their initial distribution. The question is: Can we find such a chain of permutations that will bring all 0s and 1s from the initial string into a certain order specified a priori? Based on what is known about permutations, the answer to this question is yes. But this means that for any initial string (no matter how irregularly its 0s and 1s are distributed) it is always possible to find such a computation that would transform it into another string consisting of two separate and homogeneous substrings:  one consisting exclusively of 0s and another one consisting exclusively of 1s. The states represented by strings of such a sort will further be referred to as the *refined states*.

Why would having such a refined state be important to us? Because, by applying to its two homogeneous substrings the left- and right-handed engines we discussed in previous lecture, we would be able to extract the maximal energy per bit! Well, almost the maximal: of course, there will be some losses associated with switching between these two engines at the transition point where the 0s end and the 1s begin, but, as we noted earlier, in the previous lecture, all these losses can be made negligibly small, especially in the cases of long sequences of 0s and 1s.

This essentially answers the question of why we may need computation. Because computation helps us to put unstructured input into a highly structured form which, as we already know, can be used as a source of energy supporting any of our potential needs.

The only question is how to find the right computation corresponding to the desired permutation. In a defined, specific case this is not a difficult problem. However, when trying to find such a computation for a general case, we will immediately see that its form depends on the initial distribution of 0s and 1s in the registers and that different distribution may lead to different permutation algorithms. This fact could be highly discouraging unless there is some regularity in the arriving data. Indeed, if we saw a more or less similar distribution of 0s and 1s each time when loading data into registers, this would give us a certain hope that these data can be refined (or almost refined) by means of a single permutation algorithm. This explains us the role of computation and sets up the right expectation of how the ideal computation algorithm should look like.

## 2.2. Computation as a motion in memory space

Let us somewhat formalize the qualitative picture we have just described in the previous section. Denote by $\psi$ the vector representing the content of a system's registers and consisting of $N_0$ 0s and $N_1$ 1s. The total dimension of this vector is

$$N = N_1 + N_0$$

We formally define computation as the process of updating computer registers each time when the computer's clock ticks. Denote by $\psi(t)$ the vector of the registers obtained in this way by the time $t$. Assume that all computations are strictly sequential (i.e., no parallel computing is allowed) and each time $t$ the update is defined by a single Fredkin's operator $U\big(x(t), y(t), z(t)\big)$. Here by $z(t)$ we denote the address of a control bit selected at time $t$, and $x(t)$ and $y(t)$ are the addresses of other two bits – potential candidates for permutation at time $t$. Remember that the decision about whether to perform or not to perform permutation depends on the value of the bit stored under address $z(t)$. The result of such an update can be represented as

$$\psi(t+1) = U\big(x(t), y(t), z(t)\big)\psi(t)$$

Note that the $U$-operator can be represented as a $N \times N$ matrix which is either a unit matrix $I$ or a "near-unit" matrix $P$ with only two misplaced 1s:

$$U_{z=0} = I = \begin{pmatrix} I_K & 0 & 0 & 0 & 0 \\ 0 & 1 & 0 & 0 & 0 \\ 0 & 0 & I_L & 0 & 0 \\ 0 & 0 & 0 & 1 & 0 \\ 0 & 0 & 0 & 0 & I_M \end{pmatrix}, \quad U_{z=1} = P = \begin{pmatrix} I_K & 0 & 0 & 0 & 0 \\ 0 & 0 & 0 & 1 & 0 \\ 0 & 0 & I_L & 0 & 0 \\ 0 & 1 & 0 & 0 & 0 \\ 0 & 0 & 0 & 0 & I_M \end{pmatrix}$$

Here all the matrices are represented in the block form. The symbols $I_K, I_L, I_M$ denote the unit matrices of dimensions $K \geq 0, L \geq 0, M \geq 0$ such that $K + L + M + 2 = N$, while 1s are ordinary numbers.

From these definitions, it follows that the $U$ operators are real, symmetric (i.e., Hermitian) and unitary. So, irrespective of the values of , $y, z$ , we have

$$U = U^T = U^{-1} = U$$

Now if we decide to start from the very beginning, we can write:

$$\psi(t) = \left(\prod_{u=0}^{t} U\big(x(u), y(u), z(u)\big)\right)\psi(0)$$

The three functions $x(t), y(t), z(t)$ represent a trajectory in a three-dimensional space which, in turn, represents an algorithm of converting the input $\psi(0)$ into the output $\psi(t)$. We can denote this trajectory (which starts at time 0 and ends at time $t$) by$\vec{q}(t)$ stressing the fact that it is a $t$-long list of three-dimensional vectors. Analogously, the unitary operator transforming $\psi(0)$ into $\psi(t)$ we will denote by $U[\vec{q}(t)]$.Then we can write:

$$\psi(t) = U[\vec{q}(t)]\psi(0).$$

Let us stress again that the operator $U[\vec{q}(t)]$ symbolizes the computation. The character of this computation is encoded in the algorithm – i.e., in the instruction what to do with the state vector at each instant of time. And the algorithm is totally described by the path $\vec{q}(t)$ in a three-dimensional address space. Note that despite the fact that the states of our system look like vectors , they do not form a linear vector space because the operation of addition of the content registers is not defined. At the same time, the operators $U$ cannot be treated as linear operators despite the fact that we represent them as matrices. This is not only because they act at objects not forming a vector space, but also because the decision which of their two forms, $I$ or $P$, should be used depends on the vector $\psi$. So we see that computation is a non-linear evolution process despite the fact that is built of linearly-looking blocks.

## 2.3. How far can we go in memory space?

The evolution described by operator $U[\vec{q}(t)]$ may continue for an indefinitely long time until the system decides to literally benefit from it by extracting energy from the obtained state. The way the system may extract energy depends on the energy extraction mechanism it has or, in other words, on the information-driven engine it owns. As to the amount of energy extractable by means of this engine, it depends on the form of vector $\psi(t)$. For example, if the engine works like we described above – i.e., moves along the line and extracts energy from each consecutive cell – then the ideal vector for energy extraction should be maximally uniform, which means that it should be a step-shaped vector in which first $N_1$ components would be all 1s and last $N_0$ components would be all 0s. Let $\psi(t)$ be such a vector. Applying to it the cyclic permutation $C$ ,we obtain another vector

$$\psi_C(t) = C\psi(t)$$

Note that the cyclic permutation is not a Hermitian operator, but it is unitary, which means that

$$C^T = C^{-1}$$

The vectors $\psi(t)$ and $\psi(t)$ are very close to each other and differ in only two positions (in the middle where 0s change by 1s and at the end where 1s change back to 0s. So if we consider the difference of these vectors (or, more accurately, their XOR)

$$\Delta\psi(t) = \psi_C(t) - \psi(t)$$

we will get the vector which will consist entirely of 0s except the two 1s. This means that the square of this vector's norm will be equal to 4:

$$\left(\Delta\psi(t)\right)^T\Delta\psi(t) = \left(\psi_C(t) - \psi(t)\right)^T\left(\psi_C(t) - \psi(t)\right) = \psi^T(t)(C^T - I)(C - I)\psi(t)$$

This number is negligibly small in comparison with $N^2$ -- the maximal possible value of vector $\psi(t)$'s normal. This allows us to conclude that the quantity

$$Z = \frac{1}{2}\left(\Delta\psi(t)\right)^T\Delta\psi(t)$$

is the desired quantity to be minimized if we want to extract maximal energy from the final state of the system. Using the unitarity condition we can rewrite it as

$$Z = \psi^T(t)\psi(t) - \psi^T(t)H\psi(t)$$

where

$$H = \frac{1}{2}(C^T + C)$$

is a Hermitian matrix. Remembering that $\psi(t)$ is obtained from $\psi$ by a unitary operator we can simplify the expression for $Z$

$$Z = \psi^T\psi - \psi^T U^T(\vec{q}_t)HU(\vec{q}_t)\psi$$

containing all the information about the trajectories of the system, i.e., the algorithm. So we can conclude that to find the optimal algorithm for some given values $\psi$ and $H$ we need to maximize the expression

$$Z = \psi^T\psi - \psi^T\left(\prod_{u=t}^{u=0} U\big(x(u), y(u), z(u)\big)\right)H\left(\prod_{u=0}^{u=t} U\big(x(u), y(u), z(u)\big)\right)\psi$$

on all the trajectories we have.

## 2.4. Choosing the best trajectory

The problem of maximizing the $Z$-functional described in the previous section seems to be extremely complex. What makes it so complex? The main reason is that even if we solve it for some specific initial vector $\psi_1$, , later, when we get another vector $\psi_2$,, we will need to solve it again. And for each new input vector $\psi_i$ we will be forced to start minimizing the $R$-functional separately, again and again. This is because of the intrinsic non-linearity of the problem associated with the fact that the values of the trajectory's $z$-components explicitly depend on $\psi$ (which, in turn, is a direct consequence of using Fredkin gate for performing all necessary computations). But why do we want to use Fredkin gate? Because someone has told us that it is capable of reproducing all the operations NOT, OR, XOR, AND needed for performing all possible computations. And here is the core of the problem and the source of all the unnecessary complexities associated with it!

Indeed, let us ask ourselves a very simple question: do we actually need all the possible operations to solve our particular problem? The answer is no. We need only the operators that will help us to achieve the desired outcome: transform the initial vector $\psi$ with irregularly distributed 0s and 1s into the final vector $\psi(t)$ with regularly distributed 0s and 1s. And to do that we do not necessarily need the non-linear Fredkin gate whose outcome depends on the values of $\psi$. In many cases it would be sufficient to use its linear version in which the value of $z$ will depend on time only. In other words, to demonstrate the actionability of the concept of computation and even start practically using it we can limit ourselves to the case with all the three components of the vector $\psi$ having equal rights. This would make the unitary $U$-operator linear which would substantially simplify the solution of the optimization problem in many practical cases.

Now we would be able to approach this problem from the standpoint of statistics. Assume that we have multiple samples $\psi_i$ of the initial state vector and for this flow of samples we want to find a trajectory optimal from the standpoint of energy extraction process. The square error for each case is given by

$$Z_i = \psi_i^T \psi_i - \psi_i^T U^T(\vec{q}_t) H U(\vec{q}_t) \psi_i$$

And the average (statistical) error will be given as

$$\overline{Z} = \frac{1}{M}\sum_{i=1}^{M} \psi_i^T \psi_i - \frac{1}{M}\sum_{i=1}^{M} \psi_i^T U^T(\vec{q}_t) H U(\vec{q}_t) \psi_i$$

We can introduce the covariance operator (or $N \times N$-matrix) by defining it in the standard way:

$$Q = \frac{1}{M}\sum_{i=1}^{M} \psi_i \otimes \psi_i^T$$

Then the formula for $\overline{Z}$ becomes

$$\overline{Z} = TrQ - Tr\big(HU(\vec{q}_t)QU^T(\vec{q}_t)\big)$$

Since the covariance matrix is given and does not change, the problem of minimizing the above functional is equivalent to the problem of maximizing the following one:

$$F = Tr\big(HU(\vec{q}_t)QU^T(\vec{q}_t)\big)$$

Rewriting it in the expanded form gives:

$$F = Tr\left[H \cdot \left(\prod_{u=0}^{u=t} U\big(x(u), y(u), z(u)\big)\right) \cdot Q \cdot \left(\prod_{u=t}^{u=0} U\big(x(u), y(u), z(u)\big)\right)\right]$$

Summarizing, we can claim that this functional allows one to find the optimal algorithm allowing a system to convert a statistical input pattern to output with maximal extractable energy content. We can specify the form of this algorithm by specifying a chain of swap or idle operators and treat this chain as a trajectory in a three-dimensional address space. To find the optimal algorithm, one should find the optimal trajectory minimizing the functional $R$. The Hermitian operator $H$ used in the definition of $R$ encodes the energy extraction mechanism characterizing the system, while the second operator $Q$ is a covariance matrix characterizing the environment.

## 2.5. The analogy with quantum dynamics

Another thing that we have learned from this approach can be summarized as follows. There are two basic operations associated with the state vector: the evolution operator we denoted by $U$ and energy extraction procedure we can denote by $R$. The evolution operator describes computations. It is unitary and does not lead to loss of information. Its only meaning (for us) is to prepare the state for energy extraction, and this is the main reason for which we need calculations. The $R$-operator responsible for energy extraction is non-unitary and leads to complete loss of information. Using the language of quantum mechanics, we can say that the state vector collapses.

This formalism reveals striking similarities with mathematical formalisms of quantum mechanics, especially with its path integral formulation. I think this circumstance makes it quite adequate for describing quantum computations too. I am not excluding the situation that it may somehow be related to the three-dimensionality of our space – at least how it appears to us. Maybe the three-dimensionality is the way our brain's algorithms are organized and therefore we perceive our world as 3-dimensional. I am also not excluding the situation when this mathematical formalism may help to find gentle ways of introducing non-linearity in quantum mechanics (remember we started with non-linear $U$-operators) which may be helpful for understanding quantum gravity. There are obviously many topics for speculation and future research.

In any case, at the formal level, this formalism here (I mean the level of classical computing) is not equivalent to QM because its main ingredient – the linearity of the state vector space – is still missing. The sum of two state spaces in the context of classical computations is nonsense.

## 2.6. Learning as statistical sorting

What is learning? In the language of serial computations, learning is the process of finding the optimal trajectory minimizing the functional $F$. But what if we would choose the parallel approach? Or consider a combination of serial and parallel approaches?

To answer these questions, let us focus on our final goal, on what we want to achieve. In an ideal situation, the goal is to find a permutation transforming the string from a seemingly random sequence of 0s and 1s (line 1 below) to a string formed by two perfectly-separated sets of 0s and 1s (line 2 below).

$$01100011010101000011110010101001101001001111001000101011000100100$$

$$11111111111111111111111111111000000000000000000000000000000000000$$

It is easy to see that this procedure is nothing but the standard sorting. In a real (indeterministic) situation the goal is to find a permutation that transforms the flow of strings statistically similar to line 1 to the flow of strings statistically similar to line 2. By analogy with the previous (deterministic) case, we can call this procedure "statistical sorting". Thus, finding the permutation achieving statistical sorting of the incoming flow of data is the ultimate goal of every learning algorithm.

There could be many ways (serial, parallel or mixed) of achieving this goal. All these ways differ in implementation details. What is, however, important is that the transition from line 1 to line 2 can be achieved in several stages, hierarchically. We will show the idea of one of such stages in the picture below:

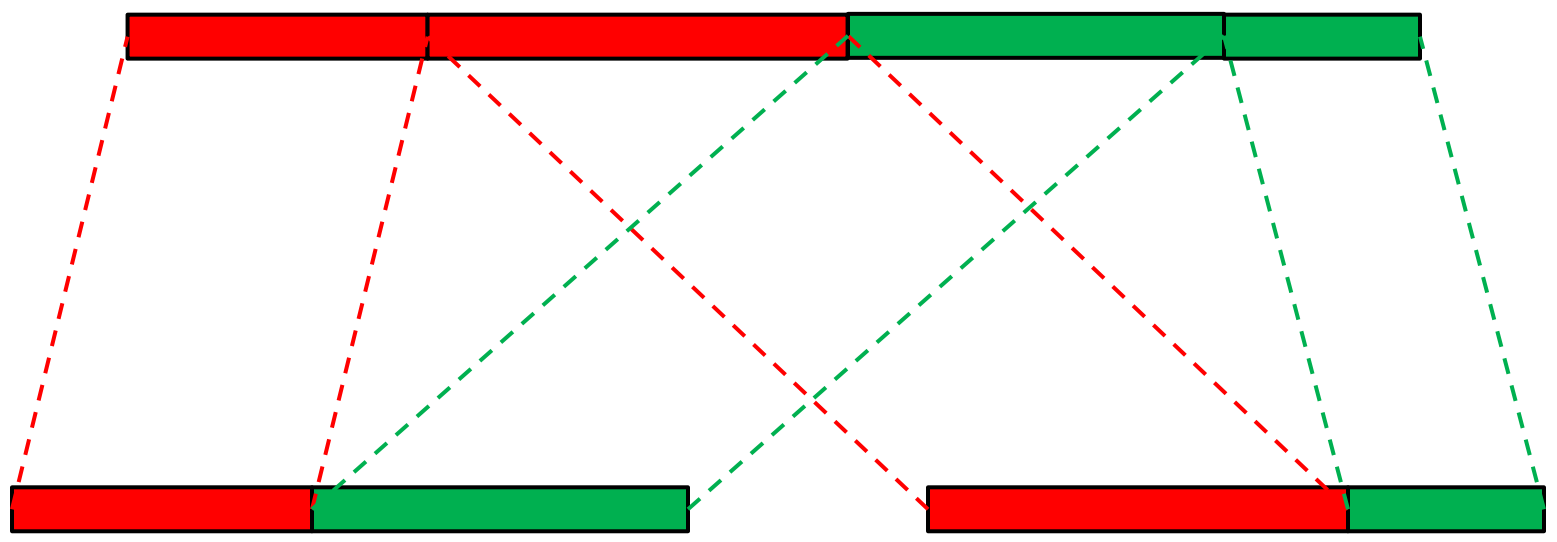

Here, the red color symbolizes the 1s and the green color – the 0s. The diagram shows two presorted sequences of 1s and 0s. The procedure combines them together and sorts the obtained combination, thus ending up with a single sorted sequence. Starting with single numbers and transitioning to their pairs

$$1+1 \to 11, 1+0 \to 10, 0+1 \to 10, 0+0 \to 00$$

and then repeating this procedure with larger and larger sequences, we will get the pyramid of similar operations. The procedure we just described is nothing but the standard "merge-sort" algorithm.

In the case of hierarchical statistical learning, we essentially use the same algorithm. Indeed, assume we have multiple stochastic variables $D_1^1 D_2^1 D_3^1 \dots D_N^1$

Let us call them the first-order variables. Consider two of them, say, $D_i^1$ and $D_k^1$. If we are able to find some permutation $P$, such that in the combination $D_{P(i)}^1 D_{P(k)}^1$ the value 1 appears more often on the left side than on the right side, then we can say that the pair of two stochastic variables $D_i^1$ and $D_k^1$ is statistically sorted by permutation $P$. But this is equivalent to saying that the variables $D_i^1$ and $D_k^1$ are correlated or, in other words that they form a statistical pattern. This pattern can be considered as a new stochastic variable – we even can give it some name. Not any pair of first-order variables can form a pattern – some of them are totally uncorrelated – such pairs are completely useless for us and we can safely consider their constituents as separate independent variables. But those variables that are correlated give rise to new patterns. We can call them the second-order variables and give them special names marking them by the upper index 2. Applying this procedure multiple times, we end up with a mixture of first- and second-order patterns. This set could be larger or smaller than the initial one. We can denote it by

$$D_1^2 D_2^2 D_3^2 \dots D_M^2$$

where $M$ is not necessarily equal to $N$. With this notation we emphasize that the output set of stochastic variables is of the same nature as the input set. But this means that the pattern creation procedure we just described above for the first-order variables can be continued further: it can be repeated for the second-order variables, then for the third-order variables, and so on.

The aim of all this reasoning was to stress the fact that the existing hierarchical learning algorithms allow formulation in the language of statistical sorting which, in turn, has a clearly energetic explanation.

## 2.7. The Autonomous Turing Machine

In the previous two sections, we have described the three basic information processing units playing the decisive role in the life of information. The first two of these units were the energy-to-information (ETI) and information-to-energy (ITE) converters while the third one was responsible for computations and was completely energy-neutral – i.e., did not lead to any energy gains or losses. In its most general form, it was represented by a circuit built of diverse reversible logical gates allowing one to manipulate with information in any imaginable way. We called it the ITI (information-to-information) converter.

In this section, we will incorporate all these three information processing units into a single mechanism whose behavior will reproduce the behavior of living and thinking organisms, of course, to some extent only. We will call this mechanism the Autonomous Turing Machine (ATM) to stress its seeming similarity with the standard Turing Machine (TM) traditionally considered in the literature as a playground of testing diverse theories of computer science. However, this similarity is only partial. There is a principal difference between the ATM and the TM: In contrast with the TM, the ATM is totally autonomous - it does not require any interaction with humans to operate. The autonomy of ATM is total in the sense that it appears on both energy and information levels. The ATM has no need for any external fuel or power supply and has no need for any data or program instructing it what to do. Everything the ATM may need for normal functioning it can extract from the information written in the memory space in which it lives and operates. As a downside of this autonomy, we should not expect the ATM to do something useful for us. We let it be on its own and do whatever it finds useful for itself.

To understand how the ATM may work, let us first describe the external memory space in which it lives. Denote by DIM the dimension of this space. Here we will limit ourselves to considering only two cases: when DIM = 1 and DIM = 2. We assume that the memory space is infinite in all directions and is divided into cells – equal-sized intervals in DIM = 1 case and equal-sized squares in DIM = 2 case. These cells thus form an infinite translational invariant grid. The second parameter is the memory capacity of each of its cells. We measure it in bits and call BPC – i.e., bits per cell. In this paper we will mostly focus on the simplest case when BPC = 1 (so each cell can store exactly one binary number) and only briefly consider the case with BPC > 1.

Now let us talk about the ATM's internals. It is critical to know that the ATM should have an energy reservoir whose capacity, due to the formula $E = kTIln2$,, can also be measured in bits. We denote this capacity by RES. In some sense, this reservoir can be treated as a certain internal long-term memory unit specialized in storing exclusively 0s (or some other a priori specified numbers playing the role of a fuel for the ATM's engine). In any case, the existence of this energy reservoir or long-term memory (whatever we call it) is a feature that makes the ATM different from the TM (remember that the latter stores the data in an external memory space). The other three blocks of ATM include the ITE and ETI converters directly connected with the RES block and the ITI converter consisting of an unspecified number of diverse reversible logical gates including the delay elements.

The final thing the ATM has is its actuator – the device that can physically move the ATM from one cell to another. The energy necessary for this actuator's work is taken from the reservoir RES. We can divide this energy into two parts. One part is the energy necessary for making decisions on where to go next. Yes, this process does require some energy because making a decision means creating new information – and we already know that it can only be done at the expense of energy. Another part is the energy necessary to perform physically this motion (or convert decision into action in business terms). It is intuitively clear that the second part is something that depends on how fast we want to move and how far we want to go. If we perform the planned motion infinitely slowly, then the total energy needed for its completion will also be infinitely small. It is worth noting that both parts of energy the ATM spends in its actuator are critically important for understanding how the entire system works. However, the different physical nature of these parts makes it more convenient to consider them separately. In the next few sections, we will mostly focus on the first (decision-making) part and the second (action) part will be discussed at the end of this section.

## 2.8. Autonomy in action

Let us finally describe how the machine works. The machine positions itself over an external memory cell and reads the information stored in it. The computing ITI block then takes this information and processes it by properly combining it with the internal information stored in the ATM's internal delay registers. The goal of this processing is to refine the incoming information to make it usable by the ITE block or, in other words, to maximally increase its potential convertibility to energy. The resulting (processed and maximally purified) information goes out of the ITI block and, after passing some special purity checker (we will talk about it in a moment), enters the ITE device in which it turns into energy. The actual amount of energy depends on the achieved level of purification (which in turn depends on how well the external information was predicted). The obtained energy is then stored in the reservoir. But what about the purity checker we just mentioned above? It acts as a special controller organized the following way. If the level of purity is acceptable, it acts as a unit operator (i.e., is completely transparent). But if the processed information has an error (i.e., there are alerts about an extent of impurity greater than acceptable), then it turns into a standard copy machine which simply creates a copy of the signal which, after some processing, is temporarily stored in one of ATM's internal registers. The role of these registers is critical: They control the actuator's behavior. The meaning of the signal stored in these registers is that it contains the detailed instruction for the actuator what to do. After the actuator has followed this instruction, the information is not needed any more and should be erased – i.e., set up to its default value. And this is where the ETI block comes into play. It does this work at the cost of energy taken from the reservoir RES.

It is important to understand the very logic of having this controlled copy machine. It works only in case of errors. If everything is fine, it does not. This is a simple manifestation of the fact that action is, in fact, the reaction of a system on the signal that something goes wrong. If everything goes well, action is not needed.

Another thing that is important to keep in mind is that not acting does not necessarily mean just staying at the same place. The ATM cannot simply stay at one place for an indefinitely long time because it needs the resource and a used memory cell cannot be used as a resource again. It is assumed that the machine performs some standard motion by default – moves from cell to cell following a certain repeated pattern stored in its internal memory. Since this pattern is already known – i.e., stored in the system's registry – following it does not cost the ATM any energy because no new information is being created. However, if the machine decides to change its pattern and start moving in a different way, this decision will be equivalent to creating new information. This is the way it is because it will result in changes that can be detected by an external observer. Creating such information results in spending some energy – its actual amount depends on how unexpected for an external observer this change was. This energy can be taken from the reservoir.

So we see that the fate of an ATM depends on the balance of the amounts of energy added to the reservoir and taken from it. If this energy will become negative or even zero, the ATM will stop functioning – or, simply speaking, it will die. So to be alive, the ATM needs to constantly make sure that the balance of energy in the reservoir is positive. The larger the balance is, the safer the ATM will be, the more leverage it will have and the longer it may live. The primary goal of the ATM is to maximize its RES, and this is what motivates the ATM to be intelligent.

The quantitative aspects and actionability of this motivation depend on the internals of the four building blocks of an ATM we discussed above, namely ITI, ITE, ETI and RES. Of course, the most important of these four blocks is the ITI block in which all the mental activity of ATM takes place. In the next few sections, we will describe step by step how it is organized.

## 2.9. The engine

Let us start with the simplest case with BPC = 1. Denote by

$$x_n \in \{0,1\}, \qquad n \in Z$$

the sequence of alphabet symbols the ATM encounters when it moves in the memory space. The order of binary numbers $x_n$ reflects the order in which the ATM visits the cells in which they are written – not the order in which they are written in memory space. For different ATM machines this sequence will look differently. So we can treat it as an ATM's internal (or subjective) representation of the external (objective) world. It is good to note here that the actual dimension of the external world has nothing to do with the perceived dimension. In our case, the memory space could be two-dimensional, but the sequence of $x_n$ will always be perceived as one-dimensional because of the chosen architecture of the ATM.

As we stressed above, we should not care about what is written in memory space. This is entirely the ATM's problem, not ours. The only general assumption we will make about the content of memory space is the character of the global distribution of 0s and 1s in it – this is what we call prior probability. We assume that 0s and 1s have equal rights, so:

$$p = p(x_n = 0) = 1/2$$

The reason why specifying these probabilities explicitly is so important to us is that it helps us to calculate the amount of energy the ATM spends in ETI or extracts from ITE. Remember that in the Kullback-Leibler formula these prior probabilities appear in the denominators of fractions staying in sub-logarithm expressions. The numerators of these fractions are the posterior probabilities describing the actual distribution of 0s and 1s in a given sequence $x_n$. If, for example, the ATM machine knows this distribution, which means that it is able to predict correctly the next symbol (for example 0) with probability

$$q = q(x_n = 0) > 1/2$$

then the amount of work extractable from such a tape would be

$$E = (kTln2)[1 + qlog_2q + (1 - q)log_2(1 - q)]$$

The need for feeding itself with energy through converting the predicted information to it is an inevitable ingredient of an ATM – earlier we called it the ITE converter. We will represent it by the following diagram:

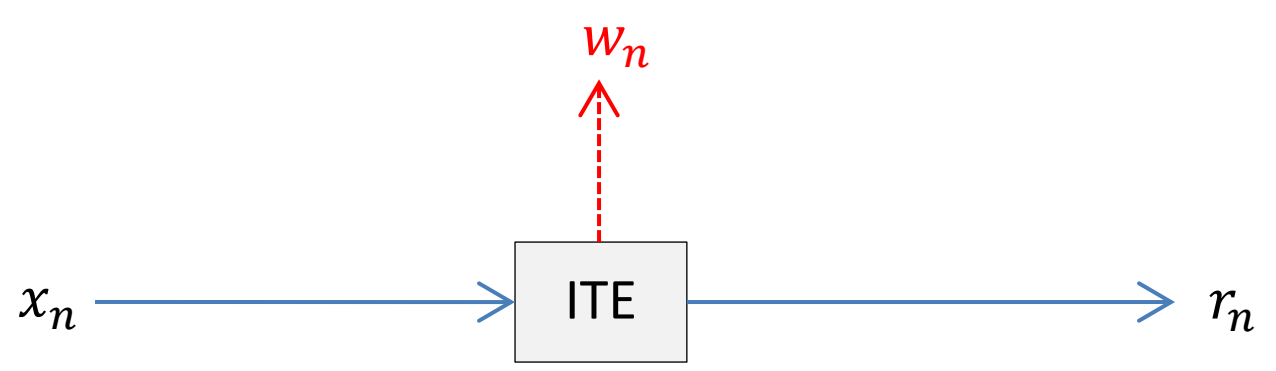

The red line shows the energy extractable from the process when the initial signal $x_n$ is converted into the completely randomized output $r_n$.

Despite the fact that this formula for average extractable energy is symmetric with respect to the exchange of $q$ and $1 - q$ , the way the ITE device is built is not (we have already discussed that earlier).

Therefore we need to explicitly select which of the two values $x = 0$ or $x = 1$ should be set as default. For the sake of definiteness we select

$$x = 0.$$

This selection, in particular, means that the optimal signal for ATM that can immediately be used as an energy source is

$$x_n = \varepsilon_n$$

where $\varepsilon_n$ denotes a binary variable statistically close to 0: $p(\varepsilon_n = 1) = \varepsilon \ll 1$. It still consists of 0s and 1s only but the 1s are assumed to be extremely rare. We call such a signal "fine" or "pure". The energy extractable from it is almost maximal. We do not hope to find absolutely fine signals consisting exclusively of 0s. This would be an ideal but practically not achievable situation. What is more realistic, however, is to try to find ways of refining the incoming signals, i.e., converting initially impure signals to one which is zero or almost zero all the time.

## 2.10. Information refiner: simplest examples

We have already seen that such a refining can be performed using an ITI block:

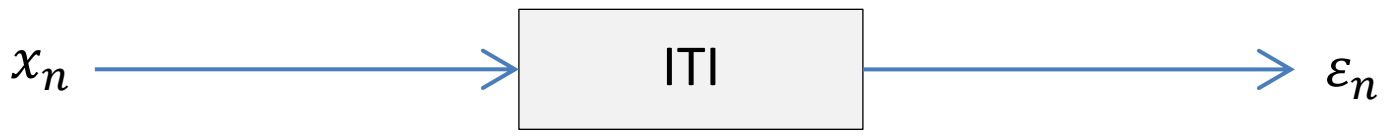

Combining this block playing, in that case, the role of the refinery with the engine will give us the general mechanism for converting information to work.

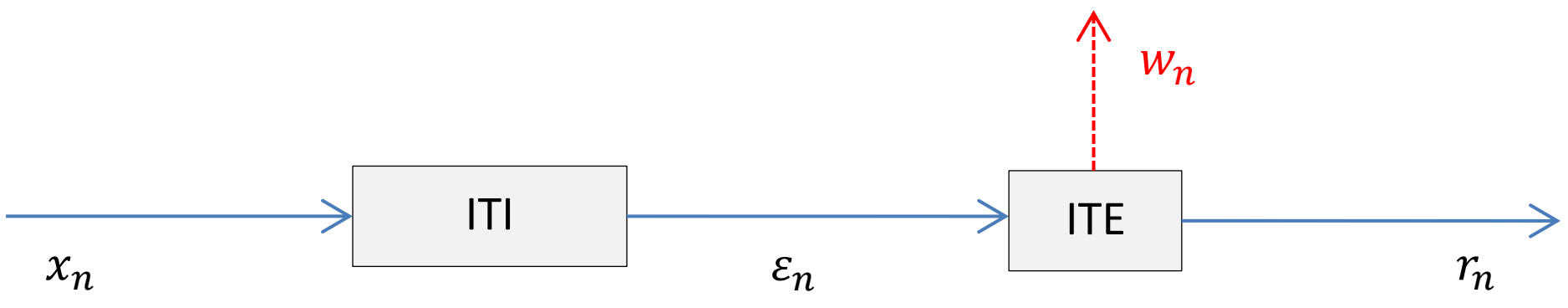

The internal organization of the ITI block depends on the type of the incoming signal.

If the incoming signal is close to 0, so it is a fine signal, then, as we saw above, the refining block is not needed, so it is simply the identity operator:

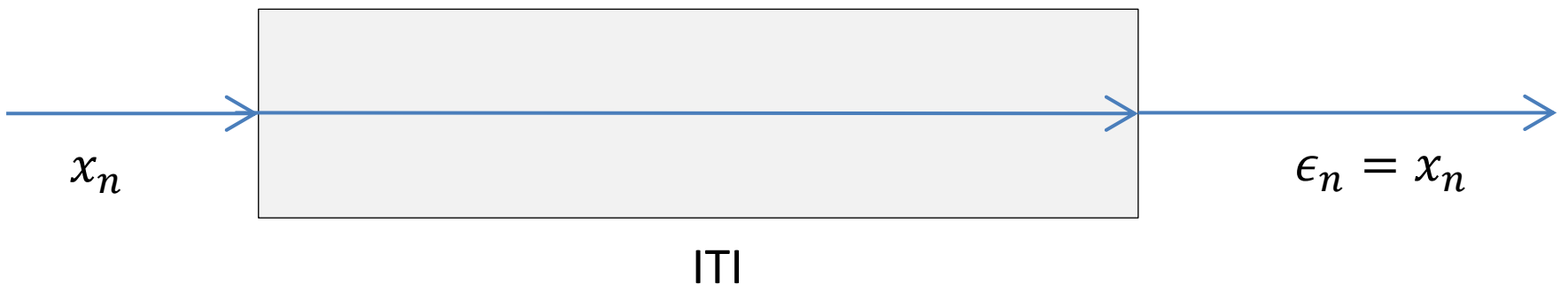

If, for example, the signal is close to 1 – we can call it an anti-fine signal:

$$x_n = 1 \oplus \varepsilon_n$$

In this case, the refining operator could have the form of a simple negation operator converting 1 to 0 or vice versa. Since the negation operator is logically invertible, it does not need any extra energy to operate.

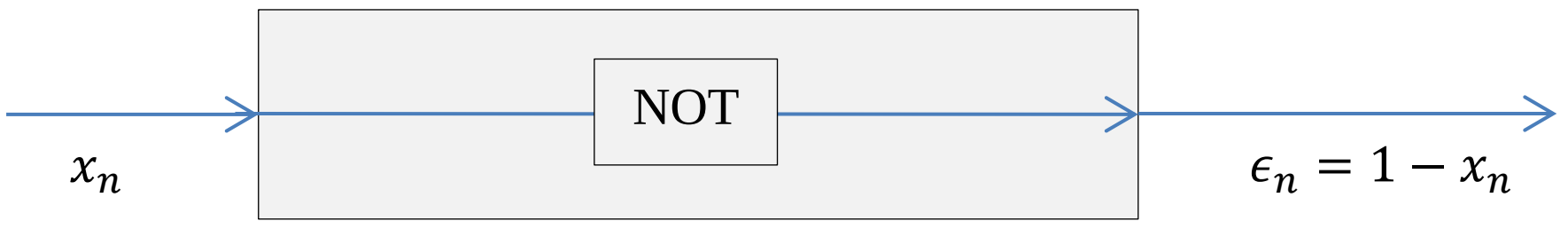

ITI

Consider a somewhat more complicated case when the signal consists of long and aperiodic sequences of fine and anti-fine signals:

$$x_n = \{\dots 0000000001111111110000000011111111111000001111111 \dots\}$$

In this case, the refining operator can be given by the following scheme:

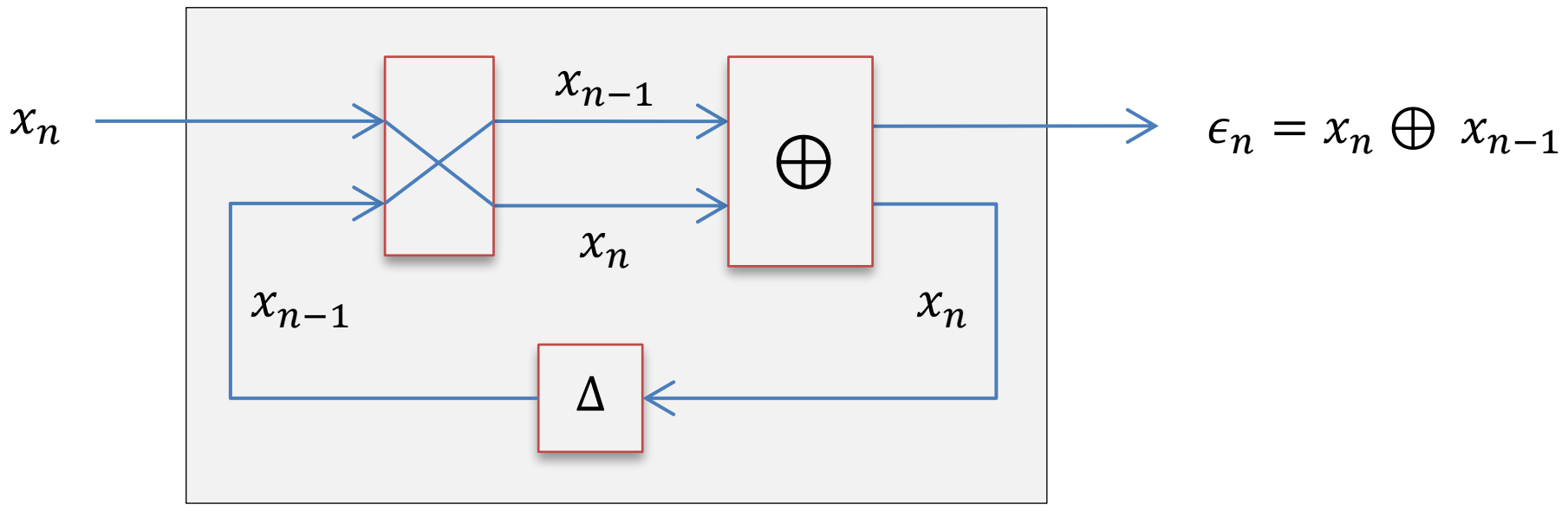

ITI

The essential element in this scheme is the delay node $\Delta$ which acts as a temporary memory storing the signal just for the period equal to the standard signal update time. Having this delay node allows one to have the previous value of a signal available along with its current value. Both the current and previous values then enter the XOR block denoted here by $\oplus$ . Its output, along with the XOR of two input values, contains the value of one of the input signals. This additional output makes the XOR-block logically reversible, which, in turn, leads to its thermodynamic reversibility. From the practical standpoint, this means that the computation of XOR does not require any external energy. The XOR of two consecutive values of the same signal forms the output which in most cases is equal to 0 except those relatively rare cases when changes from 0 to 1 or from 1 to 0 occur.

## 2.11. Information refiner: general case

To design a more general ITI-operator, we need to introduce some notations. Let

$$\vec{X}_n^k = \{x_n, \dots, x_{n-k+1}\}$$

denote a $k$-dimensional vector formed by signal $x_m$ and its $k-1$ previous values. The delay operator $\Delta$ will act on this vector as follows:

$$\Delta\vec{X}_n^k = \{x_{n-1}, \dots, x_{n-k}\} = \vec{X}_{n-1}^k$$

Note that the $(k+1)$-dimensional vector $\{x_n, \dots, x_{n-k}\}$ can be represented in two equivalent ways

$$\{x_n, \dots, x_{n-k}\} = \{x_n, \vec{X}_{n-1}^k\} = \{\vec{X}_n^k, x_{n-k}\}$$

Denote by $G$ the re- grouping operator that converts the first representation into the second one

$$G\{x_n, \vec{X}_{n-1}^k\} = \{\vec{X}_n^k, x_{n-k}\}$$

Now we can depict the construction of the ITI-operator as

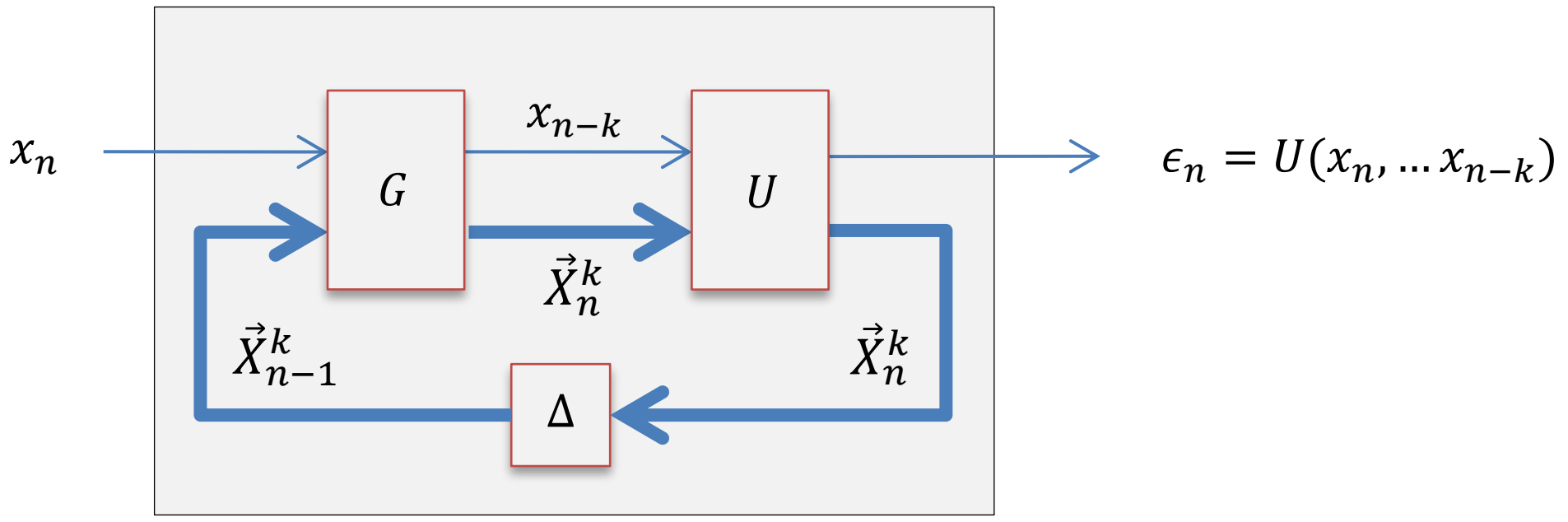

ITI

The most important object in this block is the operator $U$. Formally this $U$-operator is defined as

$$U\{x_n, \vec{X}_n^k\} = \{U(x_n, \dots x_{n-k}), \vec{X}_n^k\}$$

where $U(x_n, \dots x_{n-k})$ is a certain function of signal history. The role of this function is to refine the signal based on its limited-time history. Each step, both the signal and its history are being updated so the width of the history window is kept the same all the time. In the general case, the operator $U$ is not necessarily reversible. Here we will be interested in reversible operators only, because if the ITI block is built of reversible operators only, its operation does not require any energy cost.

## 2.12. What is understanding?

Note that the function $U$ reflects our ability to understand how the sequence of signals $x_n$ is internally organized. Indeed, if we understand all the temporal patterns formed by the sequence $x_n, \dots, x_{n-k+1}$ then we obviously can construct a function $U$ converting this sequence to zero. Conversely, if we know the explicit form of the function $U$ that converts some finite sequence of signals $x_n, \dots, x_{n-k+1}$ to zero , this means that we know the regularities in the signal flow or, in other words, we understand its patterns.

Understanding can take several particular forms. Here we will consider only two of them having the highest practical importance:  ability to explain and ability to predict. Let us consider them separately.

**Ability to explain**. It stands for the understanding of the past. Assume that we know a function $u$ such that

$$x_{n-k} = u(x_n, \dots x_{n-k+1}) \oplus \epsilon_n$$

In this case, we say that we can explain the past value of a signal $x_{n-k}$ by using its future values. The accuracy of such explanation is $\epsilon_n$. Function $U$ with needed refining property can then be defined as

$$U(x_n, \dots x_{n-k}) = x_{n-k} \oplus u(x_n, \dots x_{n-k+1})$$

**Ability to predict**. It stands for the understanding of the future. Assume that we know a function $u$ such that

$$x_n = u(x_{n-1}, \dots x_{n-k}) \oplus \epsilon_n$$

In that case, we say that we can predict with accuracy $\epsilon_n$ the future value of a signal $x_n$ by using its past values. Function $U$ with the needed refining property can then be defined as

$$U(x_n, \dots x_{n-k}) = x_n \oplus u(x_{n-1}, \dots x_{n-k})$$

We see that these two forms of function $U$ are "almost" equivalent.  However, the energy aspects of these two forms of $U$-operator are not completely the same. The problem lies in the reversibility of this operator. It is known that if an operator is logically reversible, then it is thermodynamically reversible too, and this means in turn that performing the $U$-operation will not cost the system any energy.

The question of logical reversibility can be stated in the following way: is the knowledge of the output, i.e., $x_n, \dots x_{n-k+1}$ and $(x_n, \dots x_{n-k})$ , sufficient for reconstructing its input, i.e., $x_n, \dots x_{n-k+1}$ and $x_{n-k}$? We can easily see that if the form of the $U$-operator is explanatory, then such a reconstruction is always possible, which means that this form of it is always reversible. However, in case of a predictive version of the $U$-operator this is not necessarily the case. There are cases when the $U$-operator is reversible, but there are also many cases when it is not. For example, as we saw earlier in the case of the XOR-based function

$$U(x_n, x_{n-1}) = x_n \oplus \ x_{n-1}$$

the corresponding $U$-operator is reversible because knowing $x_n$ and $x_n \wedge x_{n-1}$ ,we can uniquely reconstruct the values of $x_n$ and $x_{n-1}$. But if we would consider instead the AND-based function

$$U(x_n, x_{n-1}) = x_n \& \, x_{n-1}$$

or OR-based function

$$U(x_n, x_{n-1}) = x_n | \, x_{n-1}$$

then the logical reversibility would not be given.  Indeed, knowing that

$$x_n = 0, \quad x_n \wedge x_{n-1} = 0$$

or

$$x_n = 1, \quad x_n \vee x_{n-1} = 1$$

would not allow us to identify the value of $x_{n-1}$ uniquely. It could be either 0 or 1.

The above consideration shows us that predicting something new from purely energy-based considerations is somewhat more difficult than explaining something existing. However, from the same consideration, it also follows that there should be a form of understanding which is more general than its particular versions realized either as explanation or prediction.

## 2.13. Passive and active Turing machines

If we limit ourselves to ATMs placed in a one-dimensional memory space (on a tape) and allowed to move only one step forward or one step back – then the interaction of such ATMs with the external environment (the tape) will have a clearly passive character. Despite the fact that formally such ATMs will have the freedom of deciding where to go – one step forward or one step back – in reality this freedom will not be realized. The ATM cannot change the direction of its motion from purely energetic considerations. Indeed, all the cells the ATM leaves behind are randomized and visiting them again does not make any sense – their energy content is zero. So the only way for our ATM to survive is to continue moving forward hoping that the data it will find ahead will have enough energy content. We can call them the Passive Autonomous Turing Machines (PATMs).

So the PATMs cannot change the incoming flow of external data (which we denoted by $x_n$). They only thing they can do is to try to learn how to better accommodate to it. The $U$-operator we discussed above just incorporates the result of this learning – we called it the knowledge block. However, this knowledge has only internal importance. If the PATM decides to upgrade its knowledge block – this change will not be detected outside of the PATM. For any external observer, it will continue moving forward along the tape. Yes, the upgrade will allow the PATM to increase the quality of its life – but all these changes will be internal.

Again, such an introverted behavior of PATMs is the direct consequence of the imposed limitations allowing them to move only one cell forward or back on a 1-dimensional tape. But it is clear that if we remove at least one of these limitations, the PATM will gain something that we can call free will – the ability to decide what its next move should be.

We will call such versions of the ATM the Active Autonomous Turing Machines (AATMs). We can realize the concept in several different ways, either by allowing non-local movements (jumps) or by increasing the number of degrees of freedom giving the AATM the possibility of selecting from more local options (higher dimensionality). The jump-based version of AATM could be a model of how we read – we know very well that the process of reading is not always monotonous. The dimension-based version could be the model of how we see – our gaze point usually performs complex motions in 2D called saccades.

The question we want to address now is how these motions are scheduled and what informs their shape. To do this regarding our AATM model we need to add an important ingredient to it – the actuator. In fact, the PATM we have considered above already had such an actuator – we implicitly assumed that there is a mechanism that moves the PATM along the tape, cell by cell. However, this actuator was trivial and uncontrolled – it always performed one single operation and could not stop. What we need now is to have a controlled actuator – sort of a wheel – which would allow the AATM to change its trajectory in the state space.

## 2.14. Designing the actuator

The role of the actuator is to change something in the external world. This change may trigger some changes in the patterns of incoming data – and this is what we actually may need because this is the only way we can change the incoming data. The situation is somewhat similar to choosing the memory address and indeed, on the conceptual level, choosing a new memory address and selecting a new action is the same. If we change address, we may get access to different data. If we perform an action – we also may get access to different data.

In which case the system may need to change the way it acts? To answer this question, let us first try to understand in which cases it does not need to change the character of its action. The answer is obvious: in cases when everything is going well; if we get maximal energy; if the output of the perception block is 1: $\varepsilon_n = 1$. Then it becomes clear that the only case when we may need to change the character of the action is the case when we have an error: $\varepsilon_n = 0$. So one of the control bits controlling the action should be the error bit $\varepsilon_n$ itself. However, it is not sufficient to have only the error bit: although it tells the system that it is time to perform an action, it does not specify which action should be performed. So in order to fill this gap we need to add additional context bits to the error bit. In general, the context bits can be extracted from incoming data.

This essentially shapes up the control part of the actuator:

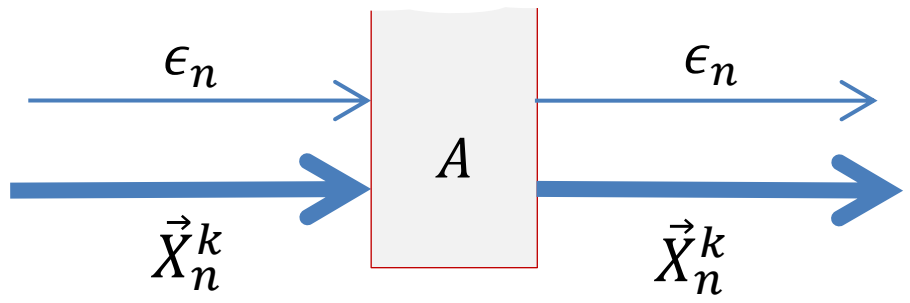

These are bits that do not change when passing through the actuator: they just control the changes that may or may not occur in the environment. Note, however, that not all the context bits contained in the vector $\vec{X}_n^k$ might actually be needed. Denote by

$$\vec{\delta}_n^k = \vec{D}(x_n, \dots x_{n-k})$$

the minimal set needed for specifying the needed action -- we can call it the decision function. Then the control part of the actuator can be represented as

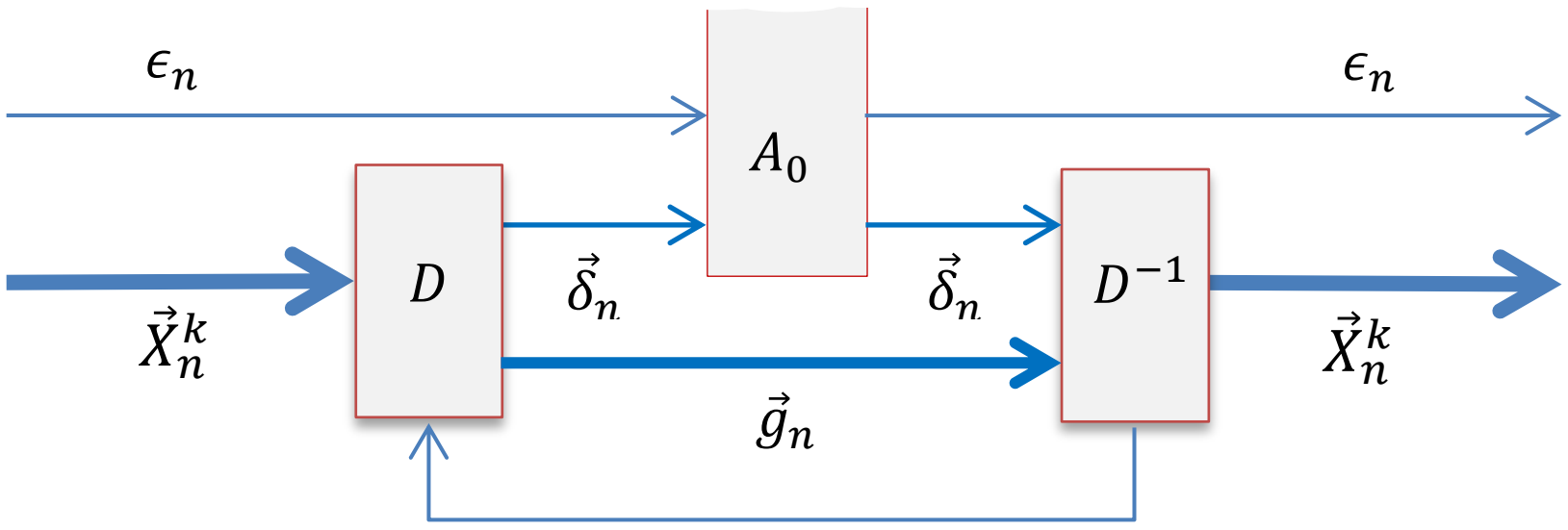

where the blocks $D$ and $D^{-1}$ denote the direct and inverse gates corresponding to the invertible versions of the decision function. The additional "garbage" bits between these blocks are also shown. Let us talk about the data one can control by this actuator – which data should be entered in the actuator's register to make it work as needed? But it is clear that this is nothing but just the decision data itself – because it by definition contains the full specification of the variety of the system's actions!

But this allows us to add the missing (data) part to our actuator:

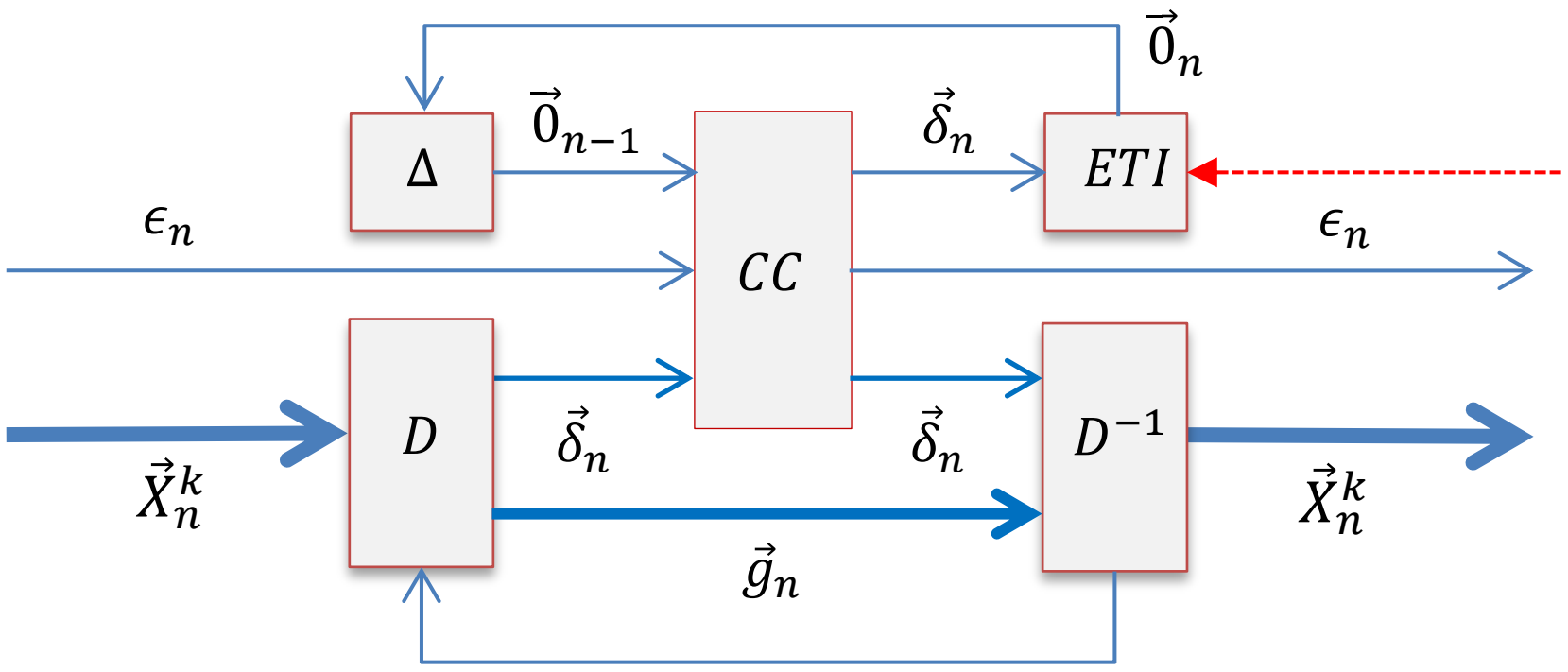

The role of the upper loop is to update the actuator's register with data coming from the decision block. The central role in this actuator is played by the controlled copy gate which obviously is reversible. However, to reset the content of the actuator's register to zero one needs to spend some energy – and this is what the G-block does if energy feeds it.

## 2.15. Combining all parts together

Note that the actuator does not consume any energy if the value of the error control bit is 0. In this case, no action is performed and the CC-operator is acting as the identity function which is reversible by definition. The zero content of the actuator's register characterizing its waiting regime does not change – it remains zero. However, in the case when the error control bit changes from 0 to 1, the CC-operator turns into a copy machine that copies the decision vector and places it into the actuator's register.  This operation is by itself reversible, but after completing the needed action the actuator's register should be reset to zero and this operation requires spending of energy. The necessary energy one can take from the energy extractor marked by 'E' in the complete block diagram of ATM shown below:

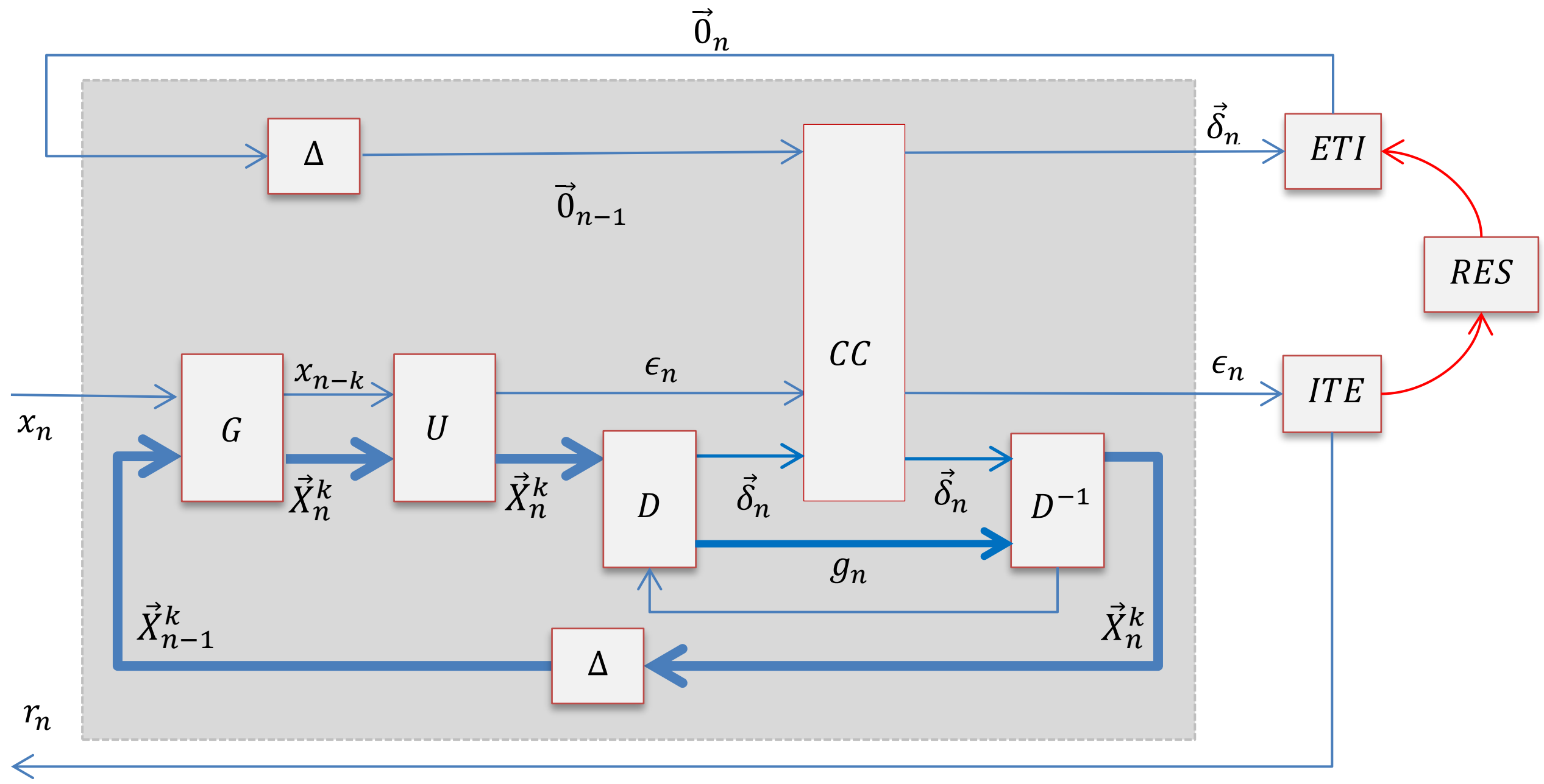

Hiding internal details of the gray box we can represent the ATM in the following way:

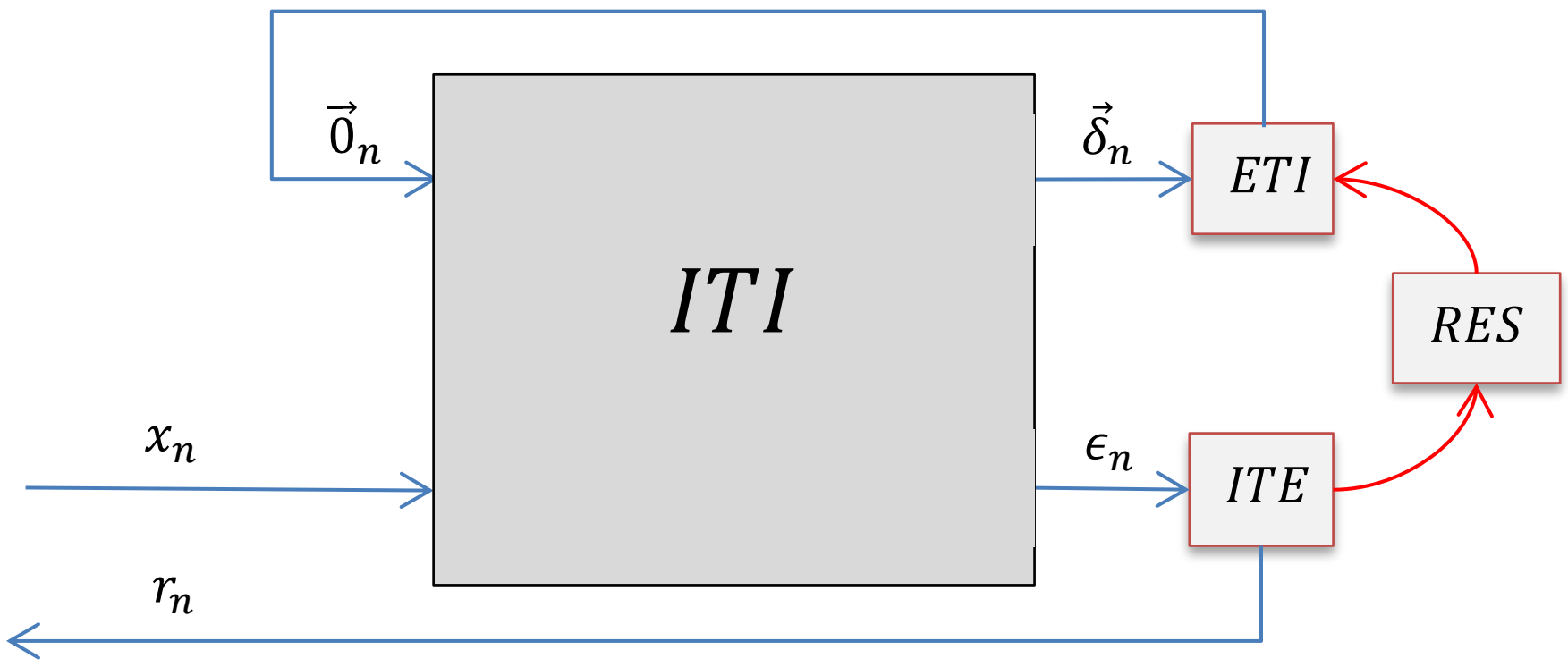

The energy production formula for this ATM looks as follows

$$E = \sum_n \left[ (\epsilon_n = 0) - (\epsilon_n = 1) \sum_{k=1}^{K} (\delta_n^k = 1) \right]$$

Or, introducing the quantities

$$U(x) = \sum_n (\epsilon_n(x) = 0)$$

and

$$A(x) = \sum_n (\epsilon_n(x) = 1) \sum_{k=1}^{K} \left(\delta_n^k(x) = 1\right)$$

we can represent the energy as

$$E = U - A$$

This analysis of energy spending in AATMs actuators reflects an idealized case when we can identify the actuator with its register, not assuming any other energy spending except the one that occurs when this register updates. This idealization is quite coherent with the level of theoretical abstraction we have chosen to discuss physical aspects of information. It corresponds to the case when all the macroscopic motions in the devices we considered so far are infinitely slow – this allows our systems always to be in thermodynamic equilibrium with the heat bath at temperature $T$.

It is intuitively clear, however, that the actual magnitude of energy spent in all these processes should be higher: it should also be somehow related to the speed of changes initiated by the actuator. Here we mean not only external but also internal changes. At this moment, it is not completely clear to me how to model the energy cost function to make it realistic on one hand and simple enough to be used in the algorithmic style discussions on the other hand. However, we can consider a rather instructive limiting case which may give us a feeling how this analysis may theoretically look. In this limiting case, we completely ignore all the internals of AATM and consider it from the standpoint of a certain hypothetic external observer looking at its behavior from outside.

## 2.16. Action-oriented machine learning

Today's machine learning algorithms have mostly a passive and data-driven nature. They deal with patterns of data only. The corresponding algorithms usually work as follows. The system initially has a device called a perception block, capable of recognizing $N$ different patterns of data forming a certain set

$$D = \{D_1, D_2, \dots, D_N\}.$$

Then the system starts observing the temporal sequences of these patterns

$$\dots, D(t_{n-1}), D(t_n), D(t_{n+1}), \dots . \text{ where } \ D(t_n) \in D$$

and tries to find regularities in their flow. If it discovers a pair of patterns $D_i$ and $D_k$ in this sequence which appear together more often than just by chance, then it forms a new pattern

$$D_{N+1} = D_i D_k$$

and extends the list

$$D = \{D_1, D_2, \dots, D_N, D_{N+1}\} \to D, \quad N + 1 \to N$$

storing its updated version in the perception block. It simultaneously checks if the probability of some of the existing patterns falls below some threshold and removes it if needed. Then the procedure is repeated with the new set of patterns. This is how hierarchical learning works.

This picture does not take the fact that the system may perform some actions which may completely change the character of the incoming data into account. To allow such a feature, we will add an actuator to the system, a block capable of initiating $M$ different actions the set of which we will denote by

$$A = \{A_1, A_2, \dots, A_M\}.$$

This addition completely changes the character of the patterns the system should deal with. Instead of purely data-based patterns, now the system should handle the mix of data + action patterns whose most general form is

$$\dots, D(t_{n-1}), A(t_{n-1}), D(t_n), A(t_n), D(t_{n+1}), A(t_{n+1}), \dots. \text{ where } \; D(t_n) \in D, \; A(t_n) \in A$$

The main difference between the data and action patterns is in the question of who initiates them. Initiating patterns and controlling them is not the same. At first sight, it looks like the system has no control over data patterns because the they are generated by the environment and, conversely, has full control over action patterns created by the system. However, the paradoxical situation lies in the fact that the actions and data are not completely independent. The actions the system performs may change the incoming data and, conversely, the character of the incoming data may suggest a system which action to perform. How to learn these relationships? The answer is simple: in the same way as before. The system starts randomly generating the actions and studies the character of composite patterns that arise as the result of these actions. But now the composite patterns may include both data and action pieces. If, for example, the triple of patterns $D_i$, $A_l$ and $D_k$ appears together more often than by chance, then the system forms a new data pattern

$$D_{N+1} = D_i A_l D_k$$

and extends the list $D$. Similarly, if the triple of patterns $A_i$, $D_l$ and $A_k$ appears together more often than by chance, then the system forms a new action pattern

$$A_{N+1} = A_i D_l A_k$$

and extends the list $A$ accordingly. The most important point in this procedure is that composition of the action patterns into a new data pattern makes them more deterministic (data dependent) and thus reduces the amount of energy spent in performing this action.

The simplest example illustrating this point is a task of learning patterns in a two-dimensional black-and-white image

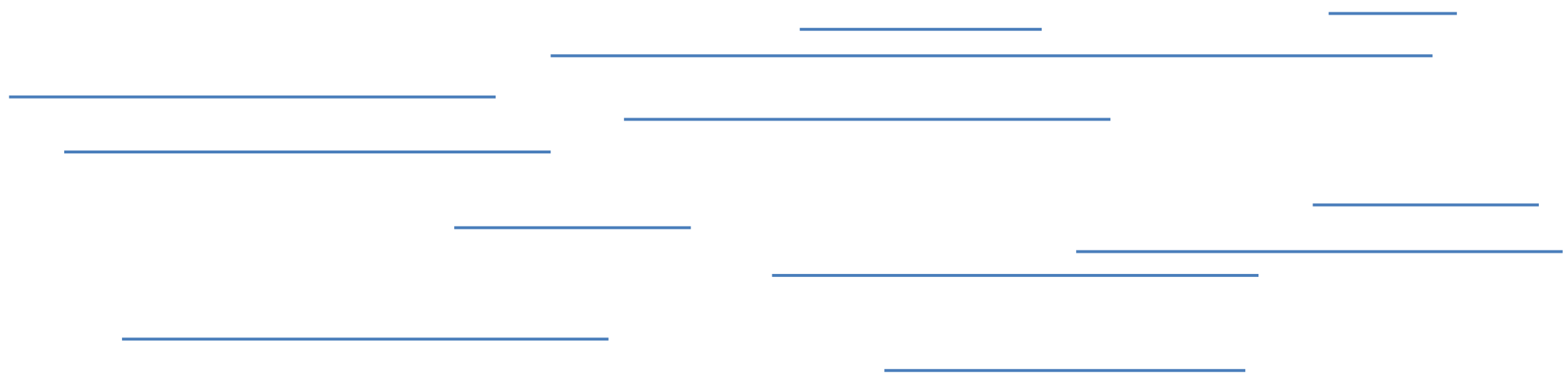

This image contains one basic pattern: the horizontal line. How to learn it? The initial set of data patterns contains two elements only: $D = \{0,1\}$, where 0 stands for black and 1 for white pixels. The set of actions will contain four elements: $A = \{0,1,2,3\}$, in which the four numbers denote four different directions of motion of our gaze point: right, below, left and up. For the sake of simplicity we forbid jumps and will consider only continuous motions. The learning starts with selecting some pixel in this image. Its color specifies the value of $D(t_1)$. Now we randomly select an action $A(t_1)$ which tells us where we should go next. This will give us the next data value $D(t_2)$. After that, we select a new random action $A(t_2)$ which will specify the next cell we will need to visit, and so on. It is easy to see that specifying action is essentially the same as specifying a memory address in which some data are stored. The only difference with the standard memory address is that in standard computers it is usually specified in absolute global coordinates while here we use the relative and local coordinates only. After multiple steps like what we described above we will end up with a random path describing random walk of our gaze point in the image space. This path will be represented by the list of performed actions and colors of visited cells. If this path is long enough, we can calculate some statistics and figure out, for example, that the black-right-black patterns are much more frequent than black-right-white patterns. We manifest this by creating a new "black-right-black" pattern. Now we add this pattern to our pattern recognition block and start our journey again. We start at a certain pixel as before and check its color. If it is white, we switch on our random number generator to select the next pixel to visit. But if it is black, we will first check the longest pattern we have, which is the "black-right-black" pattern. But reading this pattern actually means reading the instruction where to go to check if our starting point belongs to this "black-right-black" pattern: we must go right. As a result, this particular action is not random anymore! It is fully deterministic because it belongs to the pattern we have already learned. But this means that we have actually saved energy because if we follow a stored algorithm, we do not spend energy because we do not create new information.

# Lecture 3. Intelligence: an external point of view

## 3.1. The Autonomous Turing machine as a robot in unknown terrain

From the standpoint of an external observer, the AATM we have just described is nothing but a model of a robot living in a certain fuel-bearing terrain. It is totally autonomous and does not expect any help from its creators. The model of its life is extremely simple: to live, the robot needs fuel somehow distributed in the terrain. To find this fuel, it needs to move in the terrain. But how should it move? To answer that question we will need to approach this problem quantitatively and use some math.

To have some physical clarity let us agree to measure the amount of fuel as the amount of energy extractable from it. For the sake of mathematical simplicity, we will also assume that a robot is a point-like object and the fuel resource is distributed in the terrain continuously. Let $V(\boldsymbol{x})$ be a function describing this distribution. Then the quantity

$$dF_+ = eV(\boldsymbol{x})dt$$

will by definition give us the amount of fuel which the robot can suck during the time$dt$ being at point $\boldsymbol{x}$. The coefficient $e$ is introduced to characterize the robot's internal ability to suck the fuel provided that the distribution of the latter is given. The larger the coefficient $e$ is, the more fuel the robot can suck in a given interval of time.

Since all the formulas for energy in the AATM are proportional to the temperature $T$, we can write

$$e = T\varepsilon$$

where $T$ is the temperature and $\varepsilon$ is a certain constant independent on the external conditions. This relationship may have a simple illustration: Everyone will agree that drilling frozen soil and trying to suck the fuel through it is more difficult than doing the same with wet soil.

After sucking the fuel from the point-like cell $\boldsymbol{x}$ it empties, which forces the robot to leave that cell and move to other areas of the environment hoping to find a richer deposit of fuel. Assume that it covers the distance $|d\boldsymbol{x}|$ during the time $dt$. It is intuitively clear that the amount of effort to move this distance should be proportional to the distance $|d\boldsymbol{x}|$ itself. At the same time, the faster the robot covers this distance, the more resource it spends. For this reason it should also be proportional to the velocity

$$|\boldsymbol{v}| = |d\boldsymbol{x}|/dt$$

The simplest formula describing these combined losses is

$$dF_- = \frac{m|\boldsymbol{v}|}{2}|d\boldsymbol{x}|$$

where $m$ is a certain coefficient that characterizes the amount of efforts the robot needs to move. The larger this coefficient is, the larger amount of resources is needed to cover a given distance in a given interval of time. Note also that both the name of the coefficient and the number 2 in the denominator are introduced for cosmetic purposes only – just for historic reasons and further convenience. Using the previous formula for the velocity this expression can be rewritten as

$$dF_{-} = \frac{m\boldsymbol{v}^2}{2} dt$$

Before continuing, let us try to imagine what would happen with the robot if the temperature in the environment would change. Even from our limited everyday experience it is intuitively clear that moving on the surface of a frozen soil is easier than moving on the wet soil. The robot's difficulties to move, expressed in the coefficient , should be sensitive to the temperature. To be precise, they should increase if the temperature increases (motion becomes more difficult) and decrease if the temperature decreases. The simplest relationship like this is a linear function, so we can write

$$m = T\mu$$

where $T$ is the temperature and $\mu$ is a certain constant independent on the external conditions.

The robot's total resource balance (i.e., net gain minus net loss) during time $dt$ is thus given by

$$dF = dF_{+} - dF_{-} = eV(\boldsymbol{x})dt - \frac{m\boldsymbol{v}^2}{2} dt = -T\left(\frac{\mu\boldsymbol{v}^2}{2} - \varepsilon V(\boldsymbol{x})\right) dt$$

If the robot moves along some path $x(t)$, then the total amount of fuel accumulated along this path will be given by the integral:

$$F = U - T \int_{t_0}^{t_1} \left(\frac{\mu\dot{\boldsymbol{v}}^2(t)}{2} - \varepsilon V(\boldsymbol{x}(t))\right) dt$$

in which $U$ is the initial fuel resource the robot has, $t_0$ is the time when the robot starts its motion and $t_1$ is the time when it plans to end it. This equation must be complemented with the information where the robot has started its motion (the robot's initial condition) and where it wants to finish it (the robot's final goal):

$$\boldsymbol{x}(t_0) = \boldsymbol{x}_0, \qquad \boldsymbol{x}(t_1) = \boldsymbol{x}_1$$

Let us look at the integral expression for $F$. Since $F$ has the meaning of energy and $T$ is the temperature, the integral in front of $T$ should have the dimension of entropy. Indeed, remember the famous formula

$$F = U - TS$$

connecting free energy $F$ with full energy $U$, temperature $T$ and entropy $S$. Indeed, in this case the similarity of our expression with this formula is not a simple coincidence. Remember that the general energetic characteristics of the resources consumed by all living organisms are described by the thermodynamic “free energy” defined by the formula above. This is the quantity that should be maximized. This is equivalent to saying that the entropy content of good food should be as low as possible. But in our case, the information distributed in memory space whose consumption the AATM is trying to maximize is nothing but food for the AATM.

Summarizing we will come to the conclusion that the quantity the robot is trying to minimize

$$S[\boldsymbol{x}(t)] = \int_{t_0}^{t_1} \left( \frac{\mu \dot{\boldsymbol{x}}^2(t)}{2} - \varepsilon V\big(\boldsymbol{x}(t)\big) \right) dt\,, \quad \boldsymbol{x}(t_0) = \boldsymbol{x}_0, \quad \boldsymbol{x}(t_1) = \boldsymbol{x}_1$$

has the meaning of the total entropy balance along the trajectory. And the goal of a robot is to minimize this entropy balance as much as it can.

## 3.2. Surviving in an unknown memory space

Now the problem becomes mathematically tractable: a robot is successful if it minimizes its total losses and maximizes its total gain. Such robots can live longer. So, speaking about optimization of a robot's life in the mathematical context, we mean finding the path $\boldsymbol{x}(t)$ along which the functional $S[\boldsymbol{x}(t)]$ is minimal provided that the boundary conditions are satisfied.

So how to find that path? I think that the best way is to open the Landau and Lifshitz's book *Theoretical Physics. Vol. 1: Mechanics* (Landau & Lifshitz, 1969) and read a chapter devoted to the principle of least action. Why? Because a complete solution to this problem is described there. The point is that mathematically, the robot's problem of finding the optimal path through a fuel-bearing terrain is exactly equivalent to the problem of describing the motion of a classical mechanical particle with the mass $m$ and the (say, "electric") charge $e$ placed in an external (say, "electric") potential $V(\boldsymbol{x})$. The quantity $R[\boldsymbol{x}(t)]$ we are trying to maximize is nothing but the so-called classical action for the trajectory $\boldsymbol{x}(t)$ (taken with the negative sign), which in the case of classical mechanics is a subject of minimization! And the terms $m\,\boldsymbol{v}^2/2$ and $(\boldsymbol{x})$ , the difference of which stands in a sub-integral expression, have in classical mechanics the meaning of the kinetic and potential energy of a particle moving along the trajectory $\boldsymbol{x}(t)$.

In Landau and Lifshitz's book this problem is solved by using the variation method:

$$\delta \int_{t_0}^{t_1} \left( \frac{\mu \dot{\boldsymbol{x}}^2(t)}{2} - \varepsilon V(\boldsymbol{x}(t)) \right) dt = 0$$

This condition is equivalent to

$$\mu \boldsymbol{x}(T)\delta\boldsymbol{x}(T) - \mu\boldsymbol{x}(0)\delta\boldsymbol{x}(0) - \int_{t_0}^{t_1} (\mu\ddot{\boldsymbol{x}}(t) + \varepsilon \boldsymbol{\nabla} V(\boldsymbol{x}(t)))\delta\boldsymbol{x}(t)dt = 0$$

Taking into account that the variation $\delta \boldsymbol{x}(t)$ is arbitrary except the times $t_0$ and $t_1$ when, according to the starting and finishing conditions (see above), it is zero

$$\delta \boldsymbol{x}(t_0) = 0, \quad \delta \boldsymbol{x}(t_1) = 0$$

we can conclude that if the robot starts at time $t_0$ at position $\boldsymbol{x}_0$ and wants to successfully reach its a priori specified goal $\boldsymbol{x}_1$ at an a priori given time $t_1$, it should move along the trajectory

$$\boldsymbol{x} = \boldsymbol{x}(t)$$

in such a way that it would satisfy the equation

$$\mu \ddot{\boldsymbol{x}}(t) + \varepsilon \boldsymbol{\nabla} V(\boldsymbol{x}(t)) = 0$$

As a side result of this derivation there is the fact that

$$E = \frac{\mu \dot{\boldsymbol{x}}^2(t)}{2} + \varepsilon V(\boldsymbol{x}(t))$$

is a conserved quantity – i.e., it does not depend on time. So we have obtained certain characteristics of a robot – a quantity that does not change if it follows the optimal path. Measuring this quantity and checking it for time-independence would allow us to see how optimal the behavior of a robot is.

What is especially amazing about this solution is that it is local, so the only thing the robot is required to know is what happens near the point where it currently resides. To make this especially clear, we can rewrite the equation of motion in the discrete form assuming that the time interval Δt between robot steps is finite. Replacing the second time-derivative with its discrete analog

$$\ddot{\boldsymbol{x}}(t) \ \rightarrow \frac{\boldsymbol{x}(t+\Delta t) - 2\boldsymbol{x}(t) + \boldsymbol{x}(t-\Delta t)}{(\Delta t)^2}$$

we obtain the relation

$$\boldsymbol{x}(t+\Delta t) = -\frac{(\Delta t)^2}{\mu} \varepsilon \boldsymbol{\nabla} V\big(\boldsymbol{x}(t)\big) + 2\boldsymbol{x}(t) \ - \boldsymbol{x}(t-\Delta t)$$

which instructs the robot where to go next if its current state together with the current and previous positions are known. Having this indeed very limited information would allow the robot to use these "next step rules" to move step by step on a routine base.

## 3.3. Best survival practices

The solution we presented in the previous section is very simple, but it is suspiciously simple. A more careful look might make us ask some difficult questions about the relationship between the globality of the initial problem and the locality of its solution. The point is that when we formulated the robot's optimization problem, we formulated it as a global problem by specifying the robot's starting position $(\boldsymbol{x}_0, t_0)$ and its goal $(\boldsymbol{x}_1, t_1)$ and allowing these positions to be strongly separated in space and time. But how is it possible that these global parameters did not fit the solution we just found? We do not see any trace of them in our "next step rule" prescriptions.

The answer lies in the fact that before following the "next step rule" on a routine base, the robot needs to start somehow, i.e., make its very first step from $\boldsymbol{x}(t_0)$ to $\boldsymbol{x}(t_0 + \Delta t)$. But according to this rule, in order to do that it needs not only to know $\boldsymbol{x}(t_0)$ but also $\boldsymbol{x}(t_0 - \Delta t)$. And here is the core of the problem because whereas the initial position $\boldsymbol{x}(t_0)$is obviously known to the robot, its position one step before $\boldsymbol{x}(t_0 - \Delta t)$ is completely undefined – because, by condition, no steps existed before $t_0$ . This makes the robot's first step $\boldsymbol{x}(t_0 + \Delta t)$ undefined too.

So we see that all the globality of the problem has amazingly concentrated in one single question the robot should ask before starting doing anything: How to start? If there were some hypothetic advisor who would tell the robot what its first step should be, then, knowing both $\boldsymbol{x}(t_0)$ and $\boldsymbol{x}(t_0 + \Delta t)$ ,it would be able to easily calculate all the subsequent steps $\boldsymbol{x}(t_0 + 2\Delta t)$, $\boldsymbol{x}(t_0 + 3\Delta t)$, etc. and continue moving forward without any further help. But who can play the role of such an advisor? There is no one around except the robot itself.

And this is a very important point which teaches us that to be successful the robot should have both global and local vision. The knowledge of the local solution only is not enough because only through the analysis of the global solution the robot has a chance to reconstruct the magnitude and direction of its very first step. Formally, this follows from the fact the since the equation of motion is a second-order vector differential equation. For this reason, its total solution is parametrized by two arbitrary vector constants

$$\boldsymbol{x}(t) = \boldsymbol{x}(t, \boldsymbol{C}_1, \boldsymbol{C}_2).$$

Theoretically, these constants can be found after adding information of points where the trajectory is expected to start and end

$$\boldsymbol{x}(t_0, \boldsymbol{C}_1, \boldsymbol{C}_2) = \boldsymbol{x}_0, \qquad \boldsymbol{x}(t_1, \boldsymbol{C}_1, \boldsymbol{C}_2) = \boldsymbol{x}_1$$

Solving this system of two 2d vector equations on two 2d vector $\boldsymbol{C}_1, \boldsymbol{C}_2$, we can reconstruct the explicit form of the function $\boldsymbol{x}(t)$. But this way is too sophisticated even for advanced mathematicians and may require the knowledge of the function $V(\boldsymbol{x})$ in the entire environment or, at least, in a substantial part of it containing both the start and end points $\boldsymbol{x}_0$ and $\boldsymbol{x}_1$.

Of course, the robot may not have the capacity of knowing all this information or solving complex mathematical problems. However, it may state a simpler problem: just try to guess how this environment may look like in a certain medium-size neighborhood of the point where the robot initially resides. Assume that the robot initially resides at point $x_0$ and is able to predict the shape of function $V(\boldsymbol{x})$ in a certain neighborhood $N_0$ of that point with some reasonable accuracy. This would allow him to set up an intermediate goal, i.e., select some temporary destination point $\boldsymbol{x}_1$ belonging to the same neighborhood $N_0$. Knowing the approximate distribution of fuel in $N_0$ plus the initial and final conditions the robot can construct the approximate global solution – the trajectory connecting the points $\boldsymbol{x}_0$ and $\boldsymbol{x}_1$ and entirely lying in $N_0$. After reaching the endpoint $\boldsymbol{x}_1$ of this trajectory and thus completing the cycle, the robot can treat it as a new starting point and repeat the same cycle again and again.

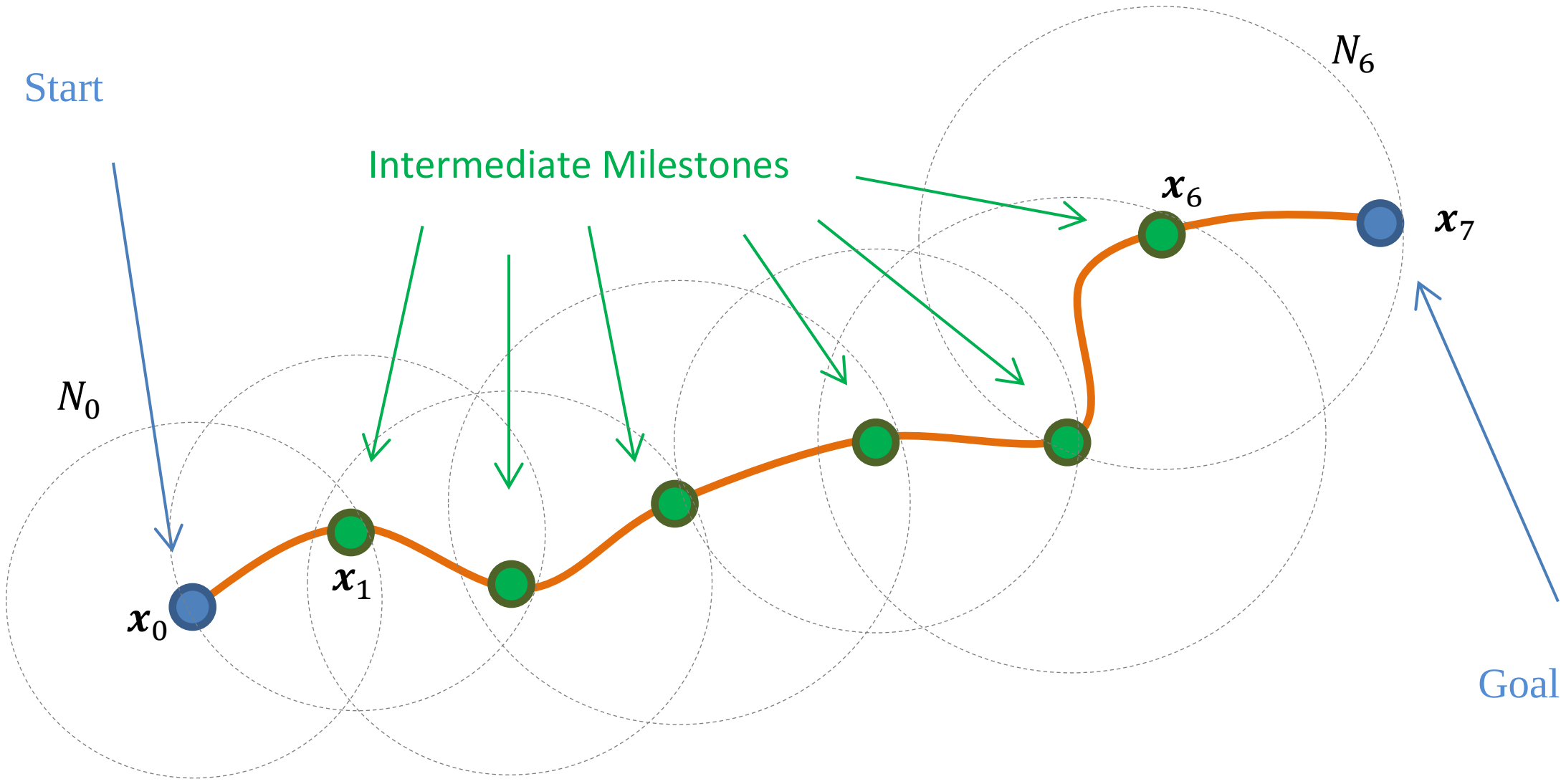

This picture allows one to formulate the general strategy for a robot. It consists of repeating interlaced steps of global and local visioning indefinitely many times. We can call them the “strategic” and “operational” planning steps. Both steps are necessary – the strategic planning helps one to meaningfully set up relatively distant goals and the operational planning instructs the robot how to practically achieve these goals – i.e., what to do on a regular base.  Whereas the strategic planning assumes intense observation and may be technologically intense, the operational planning is a fully automatic process – just a monotonous repetition of simple steps blindly following the “next step rules” derived at the theoretical stage.  There is, however, a great wisdom in this monotonicity: maximally avoiding unplanned actions allows one to save the resources in the most optimal way.  The longer the fully operational paths are, the more resources the robot can save.

If we want to have longer paths, the neighborhoods should be larger and this, in turn, assumes the ability of a robot to predict the content of larger neighborhoods accurately. So we can conclude that a robot's predictive powers are directly responsible for its overall success.

## 3.4. Two kinds of physics – one kind of math

The analogy between the problem of a robot in an unknown terrain and the problem of classical mechanics of a particle is amazing. But it becomes even more striking if we notice that the physical meaning of these two formal structures is different, despite their full mathematical equivalence.

Indeed, in the case of AI the quantity $S[x(t)]$ we have considered here had the meaning of total entropy along the path and was obviously measured in units of *energy/temperature*. In classical mechanics the same quantity (let us invert the sign to make sure that they are formally identical) has the meaning of classical action and is measured in different units of *energy · time*. Similarly, the things having the meaning of energy in classical mechanics have "slightly" different meanings in our robot-environment problem: the terms $\mu \boldsymbol{v}^2/2$ and $\epsilon V(\boldsymbol{x})$ we just talked about are not the energies in the robot's case. Although they seem to be related to the energy, they actually are not. Instead they describe the amounts of energy that could be spent or accumulated in a given interval of time at given temperature. This means they have different dimensions and are measured in different units: in *energy/(time*temperature)* instead of *energy*.

Having different dimensions is not a minor difference – it is a huge difference in physics. One of the consequences of this fact is, in our case, that the energy (in the standard sense of this word) is not a conserved quantity anymore. The quantity that *is* conserved instead has nothing to do with energy. It is the energy flow. This quantity is a very important feature of a robot: it tells us how successful the robot is. It shows the amount of external resources available to a robot if it does nothing ($\acute{x} = 0$) and, at the same time, tells us how fast the robot can move if it is in out of external resource ($V(x) = 0$). However, it is not energy.

But is it not strange that energy is not conserved? Does it not violate the most unshakable law of physics? The answer is no because the unshakable law of conservation of energy is only unshakable for isolated stationary systems and the robot/environment system is not an isolated stationary system at all. The physical meaning of this system is different: Yes, the robot takes energy from the environment in the form of fuel, but it does not return it. It fans it out when it runs its engines and burns out the fuel.

How to explain this fact: On one hand, we have exact mathematical correspondence between the two systems and on the other hand, the physical meanings of these systems are so different? On the formal level, the origin of all these differences comes from the fact that the time and the inverse temperature are not the same. They have different physical meanings and are measurable in different units. If these units were the same, then the difference between the classical action and thermodynamic entropy would disappear, at least on the formal level. A similar situation happens in relativistic quantum theory where both the speed of light and Planck constant are considered dimensionless and are taken to be equal to 1. In this system, both time and inverse temperature get the dimension of length [time] = [1/temperature] = [length]. The latter becomes the only basic dimension in that system. All other dimensions can be expressed via it: for example, [energy] = [mass] = [1/length]. As a bonus from using this system of units, we arrive at the dimensionless action and dimensionless entropy integrals. This fact essentially kills the formal difference between the Least Action Principle used in Classical Mechanics and the Least Entropy Principle used in Artificial Intelligence.

Of course, having no formal difference between these two principles does not allow one to say that these theories describe the same physics. Indeed, if [time] = [1/temperature], it does not necessarily mean that time = 1/temperature. Nevertheless, the problem of determining the motion of a closed classical system through minimizing its action functional along its path in the coordinate space is formally equivalent to the problem of determining the motion of an open, autonomous system through minimizing its entropy function along its path in the state space. The usefulness of this correspondence is that it allows us to convert some facts about open systems into facts about closed systems and vice versa.

## 3.5. The generating functional

There is a common belief that mathematics and physics are linked on a very deep level in a certain simple and fundamental way. Despite substantial difficulties in understanding this linkage in each particular case (we know that doing physics is hard), the understanding of its most general aspects (at least on the conceptual level) does not require any special knowledge of either physics or mathematics.

Indeed, consider the following obvious facts:

- A physical system may be in multiple states forming its state space.
- The motion of a system is a trajectory in this state space.
- Different trajectories may have different chances to occur.
- By specifying these chances numerically we specify the entire physical system.

At first sight, these four assertions seem pretty trivial, but they tell us about something really important. The main message encoded in them is that a single mathematical object mapping the set of all trajectories of a system onto a set of numbers may contain the complete description of a physical theory! In mathematics, such a mapping is called *functional*. Here we will call it the *generating functional*. So what we are trying to say is that knowing the generating functional for a system is equivalent to knowing everything about that system's dynamics. This almost trivial assertion essentially establishes the link between mathematics and physics. It is simple and pretty general.

The reason why, despite their simplicity, some areas of physics are not finished yet is that it is not always clear how to write down the generating functional for a concrete physical system. This task may require a lot of knowledge, creativity and hard work. But its results could be rewarding: In all known cases, specifying a physical system via its generating functional is the most economical way of describing all the variety of physical phenomena associated with it.

There is a very important class of physical systems – the so-called closed conservative systems – for which the prescriptions of creating the generating functional are more or less established and relatively well understood. In closed conservative systems, their energy exchange with the environment is so small that one can safely ignore it. For such systems, the generating functional is defined as the integral over time $t$ along the system's trajectory in its state space and the corresponding sub-integral expression is simply the difference between the kinetic and potential energies of the system. Such a generating functional is traditionally called “action“ and denoted by $S$. For purely historical reasons the action functional is defined in such a way that more probable trajectories correspond to the smaller values of it. In classical physics, for example, the correct trajectory (the number of which is always finite) always corresponds to the minimum of the action functional. The principle establishing this amazing fact is called the Least Action Principle (LAP). It is interesting that the same action functional describes quantum systems too. The only difference with the classical case is that the number of allowed trajectories in quantum case is much larger. In fact, all trajectories are allowed, but those that are far from the minimum cancel each other during the procedure called functional integration and their contribution to the final expressions becomes negligibly small.

Despite the fact that the closed conservative systems are very important in fundamental science, they form only a small part of the systems we deal with on an everyday base. Our cars, computers, animals, businesses, societies etc. are not describable (at least not fully describable) in the language of LAP. The distinguishing property of these systems is that they are open autonomous systems in which the energy is not conserved, but instead is actively exchanged with the environment. And because of this exchange we cannot apply the traditional LAP formalism to them.

But why not try to apply other optimization principles to them, principles based on more appropriate generating functionals? From our general reasoning, it follows that it does not matter how the system behaves: If it is a physical system, some of its trajectories should be more favorable than others. The only thing we need to do is to write down the right functional (analog of classical action) correctly assigning the "favorability" weight to each trajectory. The only problem is how to find such a functional. Fortunately, this problem is not as difficult as it may seem at first glance.

It turns out that we can easily explain many aspects of the behavior of open autonomous systems if we use an optimization principle structurally similar to LAP, but with different quantities in the roles of kinetic and potential energy terms. We call these quantities the consumption and action terms.

Consumption stands for the process of accumulating some resource (it could be energy, of course, but it also could be something else). The action is a reverse process when a system spends the accumulated resource for doing something useful. Here, the word usefulness refers to the usefulness for the system itself – anything that helps it to persist and to preserve its individuality.

Although the notions of consumption and spending describe opposite processes, in living systems they coexist and cannot be separated from each other, thus forming a dialectical pair: all living beings consume to spend (they energy in action), and they spend (they energy in action) to consume.

The systems of this type are everywhere. Many biological, intelligent and socio-economical systems allow this sort of description – they differ only in the types of resources they consume and in the ways they spend them. We can add to this list a wide class of open but non-living systems like diverse engines, hurricanes or stars. So the area of applicability of this consumption-action-based standpoint is broad, as we can see in the following short table containing several most typical examples:

| **System** | **Key Resource** | **Consumption** | **Key action** |
|---|---|---|---|
| Animal | Food | Eating | Hunting |
| Brain | Information | Perceiving | Looking for new info |
| Business | Money | Accumulating revenues | Delivering product |
| … | … | … | … |
| Engine | Gas | Injecting into a cylinder | Pushing a piston |
| … | … | … | … |
| Pendulum | Energy (potential) | Deviating from equilibrium | Moving to equilibrium |

What is especially amazing is that even closed conservative systems we discussed above like, for example, diverse oscillators allow this sort of description.

Note that the systems listed in this table can spend the accumulated resources in many different ways: The spectrum of possible actions may be huge. But among all the possible actions the system can perform there are always some that are critical for the system's functioning: we call them “key actions” and list them in the last column of the table. If a system stops performing any of such actions, it dies (or simply ceases to exist). Otherwise, it enters a logical loop which we can observe everywhere: for example, in the case of business, it is called business cycle, in the case of engines – Carnot cycle, and in the case of a pendulum – oscillation. And the repetition of this cycle is what in an animal's case is life and in a brain's case – the thinking process.

The most common feature of all such systems is that they try to behave in a way that maximizes the consumption of resources and minimizes their spending via reducing the number of actions to some acceptable minimum.

## 3.6. The resource maximization principle

To continue our discussion in a little bit more formal way, we’ll need to use some math. Let $X$ be the space of states $\boldsymbol{x}$ the system can be in. System’s behavior can then be described by function $\boldsymbol{x}(t)$ – representing its trajectory in the state space. Let $C(\boldsymbol{x}(t))dt$ denote the amount of resource the system consumes during an infinitesimally small time interval $dt$ when it is in the state $\boldsymbol{x}(t)$. And let $A(\dot{\boldsymbol{x}}(t), \ddot{\boldsymbol{x}}(t), \dddot{\boldsymbol{x}}(t), \dots)dt$ be the amount of resource the system spends during the same interval $dt$ when it acts. In this language, action means a change of state, therefore, instead of the function $\boldsymbol{x}(t)$ we have used its derivatives. The system moves in the state space in a way that will maximize the integral

$$R = \int_{t_0}^{t_1} \left(U(\boldsymbol{x}(t)) - A(\dot{\boldsymbol{x}}(t), \ddot{\boldsymbol{x}}(t), \dddot{\boldsymbol{x}}(t), \dots)\right)dt$$

along the trajectory. If we know both that the initial state system starts with at time $t_0$ and its future state system wants to be at some time $t_1$

$$\boldsymbol{x}(t_0) = \boldsymbol{x}_0, \quad \boldsymbol{x}(t_1) = \boldsymbol{x}_1$$

then this optimization problem for a path $x(t)$ becomes mathematically tractable. The form of this problem resembles the form of the least action principle (LAP), but we will call it here the (internal) Resource Maximization Principle, or simply, RMP.

There are at least two reasons why we prefer to use for this principle the name different than LAP.

The first reason is that we expect RMP to have broader applicability than LAP. Indeed, RMP can automatically be applied to all closed conservative systems to which LAP is applicable. To show this it is sufficient to interpret potential energy as consumption, kinetic energy as action and change the sign of the functional we want to optimize by replacing minimization by maximization. From the standpoint of fundamental science, the analogy between open and closed systems may sound rather metaphorical, but for all practical purposes, it is not. In what follows, we will try to demonstrate that it is deep enough to be taken seriously and to have many intriguing theoretical and practical implications.

The second reason is related to the fact that LAP is a mathematical problem in the traditional sense of this word and RMP is not. The first difference becomes apparent if we look at the ways LAP and RMP answer the question who is solving the problem. For LAP, this question is irrelevant. Indeed, who cares who is going to solve a given mathematical problem? This question lies beyond the scope of math. But in the case of RMP, the situation is different because the designated problem solver for a given system is the system itself. And both the character and the quality of the problem solution may depend on the amount of resource the system has. The second difference is related to the question what is given. In traditional mathematics, when formulating a problem, we usually specify this in advance. In LAP, for example, it is implicitly assumed that the forms of the functions describing both kinetic and potential energy *are given*. In RMP, this is not always the case. The systems may have a very limited initial knowledge about the functions $U$ and $A$ and even may need to learn $U$ and build $A$ from scratch. The character of this learning process and thus the quality of the resulting solution again highly depends on the internal characteristics of a system itself. So, summarizing what we said above, we can conclude that RMP is not only broader than LAP but also very different.

The number of candidate systems to which the RMP could be applied at least in theory is huge and includes all systems for which the paradigm of resource/consumption/spending/action makes sense. For all these systems (which initially look very different from each other) the RMP could play the same unifying (cementing) role as the LAP has played in physics of closed conservative systems.

## 3.7. What can help us to understand AI better?

In the previous two sections, we have discussed some quasi-philosophical questions related to the resource optimization problem and its relationship to the problems of physics. However, we should not forget the main reason why we have started this discussion. Our primary goal was to understand on a quantitative level the motivational dynamics of AI systems. Therefore, the answer to the question of what kind of systems describable by RMP we should focus on in the first place is more or less clear: AI systems. But what about other systems? The best candidates for the second echelon are those systems that reveal some similarities with AI and thus could potentially help us in studying the latter by analogy. We already considered one such system – the model of classical mechanics (CM). But there is also another system similar to AI which is no less interesting and helpful. This is the system of business economics (BE). These three systems are related to each other as shown in figure below:

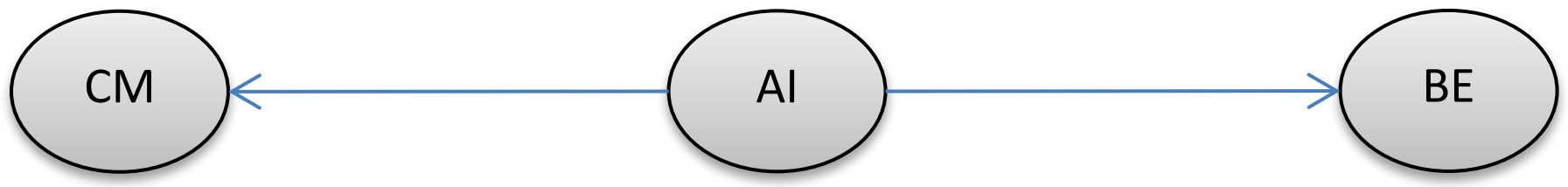

The Artificial Intelligence (AI) lies in the middle between two diametrically opposite systems of Classical Mechanics (CM) describing the non-living world and Business Economics (BE) describing social phenomena.

One reason for bringing these systems together is purely theoretical because they realize at least three dichotomies: closed vs. open (CM vs. AI & BE), living vs. non-living (BE vs. CM & AI), intelligent vs. non-intelligent (AI & BE vs. CM). Studying these dichotomies from the standpoint of their possible unification makes perfect theoretical sense.

Another reason is practical. Applying the rich optimization formalism of CM to AI and BE may help us to reveal some hidden structural or conceptual facts about AI and BE with which we would be able to design better AI or BE models or interact with them better. Finding interesting connections between BE and AI may have lots of practical applications. It is probably a good place to note that these two intelligent systems have a lot in common. The only difference between them is that one of them is built up of humans and others – of some electronic devices. But structurally they are very similar, and this is the most important fact for us because it is just the structure that makes these systems stable and sets their dynamical properties. We hope that we could learn lots of useful things about AI if we understood BE and vice versa.

We will end this section with the following three questions:

1. Is there any knowledge about CM that one could reuse in AI and BE?
2. Is there any knowledge about AI that one could reuse in BE and CM?
3. Is there any knowledge about BE that one could reuse in CM and AI?

## 3.8. A few words about physics: is time a resource?

When exploiting the physical laws of information dynamics of open autonomous systems, we naturally ended up with formulas similar to those that are traditionally used for describing the dynamics of closed conservative systems. In that particular case, this was a model of a non-relativistic classical particle moving in an external potential. Now we want to show you the existence of a no less interesting analogy with the relativistic case.

It is not surprising that the point $(Q, P) = (R, R)$ at which the minimax is attained is a saddle point. This means that if we perturb it as

$$P_1 = R_1 + p_1, \quad Q_1 = R_1 + q_1, \quad P_2 = R_2 + p_2, \quad Q_2 = R_2 + q_2$$

with

$$p_1 + p_2 = 0, \quad q_1 + q_2 = 0$$

and expand the energy $W$ near it, retaining the terms of orders not higher than 2 in perturbation, we will see that all first-order perturbation terms disappear and we obtain a quadratic expression

$$E = kT\left(\frac{p_1^2 - q_1^2}{2R_1} + \frac{p_2^2 - q_2^2}{2R_2}\right)$$

This formula tells us what happens on the border of chaos. By chaos we now mean a state which is unstructured and cannot be used as an energy source. We see that if our state is not strictly chaotic but is simply close to chaos, we can still extract from it some energy. This energy will consist of two parts: the positive part (corresponding to passive consumption and proportional to $p^2$) and the negative part (corresponding to active spending and proportional to $q^2$).

For the sake of simplicity consider the particular case when

$$R_1 = R_2 = \frac{1}{2}, \quad p_1 = -p_2 = p, \quad q_1 = -q_2 = q$$

In this case, the formula for energy gain near chaos becomes especially simple:

$$W = 2kT(p^2 - q^2)$$

Note that this is the average energy extractable from a single cell. If we move along an infinite line consisting of many such cells we can ask ourselves about our productivity. How much energy can we accumulate during a small interval of time? This amount should obviously be proportional to the length of this interval. So we can write:

$$dW = \frac{kT}{e}(p^2 - q^2)d\tau$$

where $d\tau$ denotes such an interval and $e$ is a certain coefficient characterizing the productivity of our engine. Now we are approaching a very interesting moment because nobody has told us that the environment parameter $p$ and our action characteristics $q$ should never change. In fact they can (and should) change and we can reflect that by considering them as functions of time $\tau$:

$$p = p(\tau), \qquad q = q(\tau)$$

Without loss of generality, we can treat these functions as derivatives of some other functions $t(\tau), x(\tau)$ representing them as

$$p = \frac{dt(\tau)}{d\tau}, \qquad q = \frac{dx(\tau)}{d\tau}$$

Now the formula for work can be rewritten as

$$dE = kT\left[\left(\frac{dt}{d\tau}\right)^2 - \left(\frac{dx}{d\tau}\right)^2\right]\frac{d\tau}{e}$$

To make this formula more realistic, we need to take into account the inertia of the wheel which is used in our engine. The larger the inertia, the smaller the risk of a touchdown and the larger amount of a resource the system can collect in the long run. So the net effect must be proportional to time. Without loss of generality, the coefficient of proportionality can be represented as $m^2 e$ where $m$ is a certain wheel-specific constant and other two parameters have been defined earlier. This will give us:

$$dE = kT\left\{\left[\left(\frac{dt}{d\tau}\right)^2 - \left(\frac{dx}{d\tau}\right)^2\right]\frac{1}{e} + m^2 e\right\}d\tau$$

But how to measure the time? The most natural way is to relate it to the wheel's rotations. But this can be done in many different ways. To reflect this freedom we can consider a time-dependent coefficient

$$e = e(\tau)$$

The big reason for using this coefficient twice in the expression for $dW$ is that the integral of this expression over $d\tau$ should be invariant under re-parametrization of $\tau$. This will give us the final functional to be maximized along the path $C(\tau)$: $t = t(\tau)$, $x = x(\tau)$:

$$E = kT\int\left\{\left[\left(\frac{dt}{d\tau}\right)^2 - \left(\frac{dx}{d\tau}\right)^2\right]\frac{1}{e(\tau)} + m^2 e(\tau)\right\}d\tau$$

Before going further, let us consider the function $e(\tau)$ which is responsible for re-parametrization of time and figure out how can this re-parametrization be achieved. It is easy to understand that the need for it may occur only if the length of cylinders changes. Indeed, we saw that the energy we can extract from the cylinder does not depend on its length, but the speed with which this extraction can be performed does. If, for example, the cylinder will be twice larger – the time needed to perform the same cycle will increase by a factor of two. But this means that the function $e(\tau)$ belongs to the environment and can be controlled by the environment only. This means that if we want to exclude this function from the above expression, we will need to minimize it with respect to $e(\tau)$. This statement is in full accordance with the form of expression for $W$ which has no maximum with respect to $e(\tau)$ but has a well-defined minimum.

Let us find it. After calculating the first variation by $e(\tau)$ and equating it to zero we get

$$e(\tau) = \frac{1}{m}\sqrt{\left(\frac{dt}{d\tau}\right)^2 - \left(\frac{dx}{d\tau}\right)^2}$$

Substituting the found expression into the previous formula, we will get the quantity that should be maximized:

$$E = kTm\int\sqrt{\left(\frac{dt}{d\tau}\right)^2 - \left(\frac{dx}{d\tau}\right)^2}\,d\tau$$

Excluding the parametrization from it, we obtain

$$E = kTm\int ds$$

Where

$$ds = \sqrt{dt^2 - dx^2}$$

But this means that the corresponding expression for entropy that should be minimized is

$$S = -m \int ds$$

But this is nothing but the action for a relativistic particle (Landau & Lifshitz, 1971) in a 1+1-dimensional spacetime! It is interesting that this same aspect of the above derivation shows striking similarity to the so-called world-line formulation of relativistic particle physics (Schubert, 2012) in which the original action for a relativistic particle is quadratic (not linear) in $ds$. Note that adding more spatial coordinates is not a problem here because spatial coordinates correspond to diverse actions the system can perform.

If this derivation is not a coincidence, it leads us to two interesting questions:

- Is it possible to treat time as something associated with the process of the consumption of a resource and coordinate – as something related to the opposite process of spending a resource?
- Is it possible to explain the asymmetry between time and coordinate in the expression for the relativistic interval in this language?

We will leave this interesting fact and its possible interpretation for the future.

## 3.9. A few words about business: how to facilitate its growth

Having some sketch of the mathematical formalism for describing autonomous systems, we can try to apply it to one of the most typical examples of open autonomous systems: to businesses. Businesses are, in a certain sense, an ideal playground for testing diverse AI hypotheses – because they are the only living organisms allowing unlimited experimentation with their "mental" phenomena. Indeed, in the case of more traditional living organisms like cells, animals and brains we have no way of controlling what happens inside them. So studying these systems will limit our ability of experimenting and testing our findings in practice. As to businesses, they are free of such limitations because, being their insiders, we have full control over their internal processes.

So let us start. Let us start with the assertion that Resource = Money in the case of business. The current resource balance is the difference between the revenues (this is what businesses consume) and expenses (this is what businesses spend when they act). The current resource balance for a business is its current profit. The role of the space $X$ is played by the metric characteristics of the corresponding economic sector the business belongs to. Diverse metrics characterizing the position of a business in that sector play the role of the coordinates $\boldsymbol{x} \in X$. In business language, they are often called the KPIs – i.e., key performance characteristics. The dynamics of a business is represented by its trajectory in KPI space $\boldsymbol{x}(t) \in X$. The main goal of every business is to optimize this dynamics through maximizing its cumulative profit $R[\boldsymbol{x}(t)]$ along this trajectory. This is how we should understand the business version of the resource maximization principle.

Solving RMP exactly is an unreal task for any business because to do it in a consistent way, one should know in advance at least two key ingredients: the full description of the economic sector (space $X$), and the distribution of resources in that space (function $V$). In reality, these are only partially known. However, businesses can simplify their lives and solve RMP approximately and this is what they usually do. To get some insights about $X$ and $V$ (and thus make the optimization problem mathematically tractable) they usually conduct market research which assumes a more or less accurate analysis of the possible distribution of resources $V$ in a certain limited sector of the space $X$. This sector is usually defined as a limited but still reasonably large neighborhood $N_0$ surrounding the point $\boldsymbol{x}_0 = \boldsymbol{x}(t_0) \in N_0$ describing the current economic situation of a business and essentially representing its current KPI.

After completing the market research, the business needs to set up its goals – i.e., choose another point $\boldsymbol{x}_1 = \boldsymbol{x}(t_1) \in N_0$ representing the future (desired) position in the KPI space. In addition to that it needs to specify the list of concrete operations that are expected to be performed on a regular base and estimate their cost (this step is equivalent to specifying the form of a kinetic term – the analog of $m\dot{\boldsymbol{x}}^2/2$ in classical action). Once this is done, the business is fully equipped to start solving the optimization problem in its neighborhood $N_0$ or, in business language, doing strategic and operational planning.

Under strategic planning, we mean presenting the sketch of its future trajectory $\boldsymbol{x}(t) \in N_0$ in KPI space. Having such a solution would, in particular, estimate the resource balances along the trajectory, which will help the business to make sure that it starts with sufficient starting capital. This is necessary to avoid accidental touchdown and getting out of business. In terms of our model this means that the constant $U$ must be large enough to make sure that $R > 0$ along the entire trajectory. Another reason for having such a solution is to decide how aggressively to start, which is equivalent to selecting the initial velocity $\dot{\boldsymbol{x}}(t_0)$ . And, as we know from our general discussion, this is a really important point that links strategic and operational planning.

Under operational planning we here understand specifying the set of everyday rules that would help the business to determine what to do today if one knows what happened yesterday and what its current situation with the KPIs is. In the language of our toy model this corresponds to the routine use of equations of motion $m\ddot{\mathbf{x}} = -e\boldsymbol{\nabla}V(\mathbf{x})$ for calculating next steps. Operational planning provides us with a very economic approach because it is entirely based on the current values of the resource acquisition and action cost functions. However, as we already noted above, it should be used in a proper combination with strategic planning.

Now let us add a few words about the operationalization stage. After completing the planning, the business can start its operations. From the strategic planning, it knows how to start and from the operational planning it knows how to move forward towards its goals in an optimal way. The key point here is that once the motion has started, the business should follow the routine rules blindly, maximally resisting any temptation of correcting them on the fly (unless one plans these corrections in advance as a part of the routine process). The reason for that is very simple: the rules have been derived to maximize revenues and minimize expenses. Any violation of them will lead to unnecessary losses.

Summarizing, we can conclude that the formalism we derived for a robot applies to businesses as well. It has several features that make it interesting. First of all, it is mathematically justified. Second, it is rather general – it should work for any business under minimal assumptions. Third, it is so well aligned to our common sense that it is not surprising that practically every business follows (consciously or unconsciously) all its recommendations. But does this approach have any predictive power? Can we go beyond the things we already know? Is there anything in this formalism that does not seem so obvious at first glance but gives something new and interesting from the practical point of view? I think there are at least two interesting things to think about in the context of the optimal business organization.

One is the principle of resource flow conservation we deduced earlier. In business terminology, this principle states that if the business follows the chosen optimal operational routine and does not deviate from it, then the sum of profits and expenses calculated over fixed time interval should be constant. For this reason, this sum could be a good indicator of business healthiness. If it is high and does not change over time, this tells us that business is doing well and its dynamics is close to optimal. In the opposite case, this could be a sign of something wrong happening with the business and may be used as a trigger for some preventive actions. The principle of resource flow conservation looks very meaningful from another point of view too. Indeed, it tells us that if the business approaches the deposit of resources (i.e., the place where the values of $V(\boldsymbol{x})$ are large), it needs to slow down to have more time for simply consuming these resources. However, in the case when a business crosses a desert (a place with a limited amount of resources where the values of $V(\boldsymbol{x})$ are small), it needs to speed up to cross this unsafe area faster and spend less time in it.

This raises an interesting and quite general problem of developing systematic ways of discovering conservation laws in diverse business-type environments (but actually – in any type of spatio-temporal data). The practical importance of such methods is obvious because they actually tell us how to create one or more riskless portfolios of company's KPIs having high insensitivity to insignificant fluctuations of business performance parameters and thus having high reliability and predictive power. The approach is in a certain sense similar to the famous Black and Scholes method for creating riskless portfolios by using such combinations of stocks and their derivatives in which the rapid fluctuations are mutually compensated (cancelled) so that the only slowly changing components survive. See for example the original paper (Black & Scholes, 1973) and also a classic book on this subject (Hull, 2008). I plan to discuss this interesting questions in a separate and more technical publication.

Another interesting possibility to explore is related to the intrinsic periodicity of businesses. Along with the monotonous trends in KPI space, businesses often demonstrate periodic behavior in some of its subspaces. This fact should not be too surprising for us because we already know that there is equivalence between the businesses and the systems of classical mechanics for which the periodic behavior is rather typical (remember the pendulum, for example). By using this equivalence, it would be quite natural to speak of some business "eigenfrequencies" – the natural periods after which diverse business cycles repeat. But if this analogy with eigenfrequencies is correct, we can benefit from it using it for our practical needs. Indeed, the simplest way of doing so is to use the effect of resonance. The meaning of this effect in classical mechanics is very simple: if one perturbs the mechanical oscillator with a certain periodic force coherent with its eigenfrequency, then the intensity of its oscillations will increase and will be accompanied by an increase of its total energy. The force does not necessarily need to be external. The same amplifying effect is achievable via a coherent change of some internal parameters of a mechanical system. In this case, the phenomenon is called *parametric resonance*. The simplest example is the swing. It would be very intriguing to try to find similar swing-effects in businesses. Theoretically, this could be achieved via changing some external or internal parameters of a business coherently to its eigenfrequencies which may lead to the resonant amplification of some of its KPIs! If such an effect does exist, it could be a new effective way of intensifying business growth.

## 3.10. A few words about math: what is its role?

The role of math in physics is extraordinary. There are at least two reasons for that. First of all, math creates abstractions allowing one to describe complex physical phenomena in a compact and elegant way. Second, it plays a unifying role in physics revealing striking similarities between its seemingly unrelated branches. All the development of physics in the 20th century was driven just by these two abstraction and unification trends. The former resulted in a huge number of models – the idealized mathematical structures allowing one to isolate the most important features of complex systems and study them in relatively simple ways. The latter essentially allowed one to find similarities between these systems and merge them, thus unifying our knowledge about nature. The main distinguishing feature of all these models – both intermediate and final – was that they were unbelievably simple and unbelievably rich at the same time. The most impressive achievement of this combined modeling and unification effort was the creation of the Grand Unification Theory (GUT) manifesting that all fundamental forces in nature (except gravity) allow a description in the framework of a single and simple mathematical model (the so-called “Standard Model”). See for example the book (Ellis, 2002). Gravity is considered as the next natural candidate for the inclusion into this modeling and unification process, and there is no doubt that sooner or later the “final” quantum theory unifying all known fundamental forces will be created.

However, all this modeling and unification boom was related to closed conservative systems only, forming a very specific – albeit important - class of systems we deal with on an everyday base. Attempts to extend the ideas of modeling and unification to open autonomous systems describing emerging phenomena like life, intelligence and self-organization have essentially failed. However, in this paper we have seen that many mathematical structures appearing in the problem of AI show similarities with mathematical structures appearing in diverse areas of physics. How to interpret this fact? I think there are two possible answers to this question, and each answer opens a separate road of further intriguing research.

**The first possible answer**. It cannot be ruled out that this similarity is a manifestation of the fact that nature is more unified that we currently think. Maybe there is a certain common approach to both closed conservative systems and open autonomous systems? If this conjecture is correct, it may open a road to a new round of unifications in physics involving both closed and open, living and non-living, intelligent and non-intelligent systems. Maybe this is how the hypothetical TOE (theory of everything) should look like?

**The second possible answer**. It cannot be ruled out either that the origin of this similarity lies not in the external world, but rather in our brain. Maybe we perceive these areas of physics as mathematically similar because we use one and the same instrument – our brain – for their description? If this point of view is correct (and I think that it is correct), then this opens a fantastic opportunity for the reverse engineering of our brain's architecture by studying the structure of the mathematical theories it creates. Then we will end up with a TOB (theory of brain).

In any case, there are many open questions which make our world a really interesting place to live.

# Afterword: The power of equality sign

## Is the phenomenon of intelligence complex or simple?

There is a strong and common belief that complex phenomena such as life and especially intelligence are *too* complex to be studied in terms of simple mathematical models. In this text I tried to defend an opposite point of view – that although all realizations of life and intelligence we know are indeed very complex (how can we disagree with that?), nevertheless, the mathematical structure underlying these phenomena is simple. This point of view is definitely not new. Originally it was expressed in connection with the so-called cellular automata. See for example (von Neumann & Burks, 1966), (Wolfram, 2002), (Adamatzky, 2010) and (Berto & Tagliabue, 2012). It was demonstrated that even very simple rules formulated at the microscopic (cellular) level may in some cases lead to complex and non-trivial behavior on the larger scales. Although the initial results obtained in this direction were very impressive and despite the fact that cellular automata clearly demonstrated that complexity is not necessary to imitate lifelike behavior, the enthusiasm for these models has gradually faded out. The reason was the absence of connections with the real physical world or, in other words, the inability of computer scientists to explain why nature might prefer rules leading to such behavior. Some simple physical principle that would be able to explain the entire picture was needed.

To be precise, on a global level such a physical principle was already known: this was the Second Law of Thermodynamics. We know very well that both life and intelligence support the processes of decay or, in other words, support the overall increase of entropy in the world. Therefore, their existence not only does not contradict physical laws but even supports them. This statement, being correct, is nevertheless too global and too fuzzy to be constructive. It tells us "why it could be possible" in general but does not tell us "how to do it" in each particular case.

All this means that if we want to move forward, we need a mesoscopic model describing living and intelligent behavior which would be supported by math at the microscopic level and by thermodynamics at the macroscopic level. But what could be a candidate for such a model? Which mathematical notion may help us to describe all the aspects of math, physics and intelligence in a unified way?

I think the answer could be found in the equality sign – the mysterious point at which math and physics merge. Indeed, the only mission of math in the world is to encrypt the results of physical measurements. And math does that in the form of equations. If we get any knowledge about the external world, this knowledge can always be expressed in the form $A = B$, where $A$ and $B$ are the results of our observations. The meaning of the word observation may vary depending on its context. It could denote simultaneously performed measurements of quantities $A$ and $B$, or measurements performed at different times. The formal mathematical character of the equality sign is not important either: instead of the "='" symbol we can use its less formal forms like "is", "was", "will be" and so on.

We know that the power of knowledge dramatically increases if it is reproducible. We call knowledge reproducible if, when saying that $A = B$ ,we actually mean something like

$$A_n = B_n, \quad n = 1, 2, \dots.$$

which is equivalent to saying that $A$ and $B$ are always equal, no matter how many times we measure them. But what makes reproducible knowledge so important for us? We can answer this question in many different ways (we know that it is good for prediction, planning, understanding, grouping etc), but in physical terms the answer is astonishingly simple. We can say that if the results of two different measurements (performed on either two different objects or the same object but at two different times) are always equal, this is equivalent to saying that the difference of these results

$$x_n = A_n - B_n, \quad n = 1, 2, \dots.$$

is always zero:

$$x_n = 0, \quad n = 1, 2, \dots.$$

But, according to what we have learned so far, the very fact of having a steady zero signal arriving from an external environment means that we have a reliable source of energy! Roughly speaking, the more equations we know, the more energy we can extract from such knowledge!

For the same reason, any regularity in space or time is something that can be used as a direct source of energy. And this is so simple because the regularly repeating patterns can be combined to give a zero signal which is nothing but a resource. And this resource is food for any living or thinking organisms, as we discussed earlier in the lectures.

## The zero-signal principle

The above allows one to conjecture the following principle explaining the behavior of autonomous intelligent systems in a purely mathematical language:

**Any autonomous intelligent system receiving the flow of an external data $x_n$ whose content can, at least partially, be controlled by systems actions $a \in A$ should be motivated to act in a way that would bring the incoming data maximally close to zero**

$$x_n(a) \to 0_E.$$

**Remark 1:** The zero which we denoted by $0_S$ in the above formula is different from the "mathematical" zero in the respect that it is system-dependent – it is defined as a signal allowing a system's engine to work for indefinitely long time with the maximal energy production rate. This means that different systems equipped with different engines may have different zeros. Actually, one can treat these zeros as system-specific constants.

**Remark 2:** The actions a system can perform could be divided into two groups: external and internal. Whereas the external actions are focused on getting data from external sources, the internal actions are all about rearranging the existing data it in a way that makes them maximally close to the system's zero (this is what we called the refinement procedure and what in computer language is called computation).

We will call this statement the zero-signal principle. It is one of the many possible mathematical forms of the resource maximization principle we described earlier. This particular form is especially suitable for computer science applications because of its intuitive character and the easiness of incorporating it in diverse machine learning algorithms.

The zero-signal principle allows one to answer many concrete questions about the emergence of diverse features of intelligent behavior in a rather non-standard way. For example:

- Why do autonomous systems need to make accurate predictions? Because by minimizing the prediction error, they generate a close-to-zero signal having the maximal energy content.
- Why do autonomous systems need to develop a memory? Because it allows them to store old patterns which, being combined with the similarly looking new ones, may produce a zero signal.
- Why should periodic processes dominate in autonomous systems? Because periodicity is something that is easily convertible to a zero signal and thus can be used as an energy source.
- Why may autonomous systems need to cluster data or, in other words, create concepts? Because the clusters give us large areas of uniformity easily convertible to system zeros.
- Why may autonomous systems need to conduct principal component analysis? Because it would allow them to isolate signals close to zero and use them as a reliable energy source! Note that traditional computer scientists prefer to pay attention to the remaining (large) components!

It is interesting that this list the practical usefulness of which, at first glance, is limited to the hypothetical future computing systems that may function on the microscopic (molecular) scales turns out to be extendable even to macroscopic systems lying far from the areas of computer science.

- Why are we looking for symmetries?
- Why do conservation laws determine the motion of physical systems?
- Why is the dynamics of physical systems described by minimization principles?
- Why do we love art: ornaments, poetry, and music?
- Why do we have stable habits?
- Why are businesses based on repetitive routines?
- Why are we looking for unifications of physical theories?
- Why do we ask questions?
- Why nature loves making multiple copies of the same thing?

All these questions can be answered in a very similar way: because any pattern, i.e., any understood regularity (uniformity, periodicity, repetition, and equality) allows one to save or generate a resource. This could be an emotional resource (we relax when we listen to music), a financial resource (we save/make money when we reduce business operations to repetitive tasks, a mental resource (we simplify physical theories formulating them in terms of symmetries, conservation laws and minimization principles), knowledge resources (we ask questions to get answers which are in form $A = B$ and thus save resources). As we see, a resource is not necessarily a physical energy (unless we consider it at the microscopic scales). However, whatever resource we may use, it is always directly or indirectly linked to physical energy responsible for our physical survival on an individual or social level.

# Acknowledgements

I would be very grateful for any comments and feedback.

# References

The list of references is organized in two different ways: by subject name listed in the order as it appears in the text and by first author's name. This list is far from being exhaustive but is quite representative because it contain some key research papers, reviews and books with very well organized references therein. An additional and, actually, excellent source of high-level review articles with carefully selected references is Wikipedia – it is highly recommended for those who want to just save time and get quickly to the point by doing simple keyword search. The items in the subject list can be used as keywords.

## References by subject:

- Artificial Intelligence:
  - (Russell, & Norvig, 2003)
  - (Wikipedia, 2016)
  - (Müller & Bostrom, 2014) URL,
- Turing Test
  - (Oppy & Dowe, 2016) URL,
  - (Saygin, et al., 2000) URL
- Deep Learning
  - (Bengio, 2009) URL ,
  - (Deng & Yu, 2014) URL ,
  - (Masters, 2016) URL ,
  - (Ng, 2016a)
  - (Ng, 2016b)
- Master Algorithm
  - (Domingos, 2015)
- Intelligence
  - (Hawkins & Blakeslee, 2005)
  - (Legg & Hutter, 2007) URL
  - (Zoldi, 2015) URL ,
  - (Lipson, 2007) URL
- Information and Energy
  - (Landauer, 1961) URL,
  - (Bennett, 1982) URL,
  - (Plenio & Vitelli, 2001) URL,
  - (Maroney, 2009) URL

- Physics of Information
    - (Landauer, 1991),
    - (Bais & Farmer, 2007) URL
    - (Zurek, 1991),
    - (Leff & Rex, 2003)
    - (Wheeler, 1989)
- Information in Biology, Evolution, Robotics
    - (Avery, 2003 )
    - (Schneider, 1994)
    - (Adami, 1998)
    - (Adami, 2012)
    - (Adami, 2016 )
    - (Palmer, et al., 2013) URL
    - (Der, et al., 2008) URL
- Turing Machines
    - (Barker-Plummer, 2016) URL ,
    - (Penrose, 1994)
- Physics of Computation
    - (Bennett, 2003) URL ,
    - (Feynman & Hey, 2000)
- Information and Statistics
    - (Kullback & Leibler, 1951)
- Reversible Gates
    - (Fredkin & Toffoli, 1982) URL
    - (Perumalla, 2014)
    - (Kerntopf, n.d.) URL
- Theoretical Physics
    - (Feynman, et al., 1963)
    - (Landau & Lifshitz, 1969)
    - (Landau & Lifshitz, 1971)
    - (Schubert, 2012)
    - (Ellis, 2002)
- Cellular Automata
    - (von Neumann & Burks, 1966)
    - (Wolfram, 2002)
    - (Adamatzky, 2010)
- Financial Mathematics
    - (Black & Scholes, 1973)
    - (Hull, 2008)

## References by name